\documentclass[10pt,twocolumn,letterpaper]{article}

\usepackage{cvpr}
\usepackage{times}
\usepackage{epsfig}
\usepackage{graphicx}
\usepackage{amsmath}
\usepackage{amssymb}

\usepackage[table]{xcolor}
\usepackage{color}
\usepackage{rotating}
\usepackage{array}

\usepackage{tikz}
\usetikzlibrary{spy, shapes, arrows, matrix, pgfplots.groupplots}
\usetikzlibrary{decorations.text}

\usepackage{graphicx}
\usepackage{caption}
\usepackage{subcaption}
\usepackage{pgfplots}
\usetikzlibrary{spy,calc}

\newif\ifblackandwhitecycle
\gdef\patternnumber{0}

\pgfkeys{/tikz/.cd,
    zoombox paths/.style={
        draw=orange,
        very thick
    },
    black and white/.is choice,
    black and white/.default=static,
    black and white/static/.style={ 
        draw=white,   
        zoombox paths/.append style={
            draw=white,
            postaction={
                draw=black,
                loosely dashed
            }
        }
    },
    black and white/static/.code={
        \gdef\patternnumber{1}
    },
    black and white/cycle/.code={
        \blackandwhitecycletrue
        \gdef\patternnumber{1}
    },
    black and white pattern/.is choice,
    black and white pattern/0/.style={},
    black and white pattern/1/.style={    
            draw=white,
            postaction={
                draw=black,
                dash pattern=on 2pt off 2pt
            }
    },
    black and white pattern/2/.style={    
            draw=white,
            postaction={
                draw=black,
                dash pattern=on 4pt off 4pt
            }
    },
    black and white pattern/3/.style={    
            draw=white,
            postaction={
                draw=black,
                dash pattern=on 4pt off 4pt on 1pt off 4pt
            }
    },
    black and white pattern/4/.style={    
            draw=white,
            postaction={
                draw=black,
                dash pattern=on 4pt off 2pt on 2 pt off 2pt on 2 pt off 2pt
            }
    },
    zoomboxarray inner gap/.initial=5pt,
    zoomboxarray columns/.initial=2,
    zoomboxarray rows/.initial=2,
    subfigurename/.initial={},
    figurename/.initial={zoombox},
    zoomboxarray/.style={
        execute at begin picture={
            \begin{scope}[
                spy using outlines={%
                    zoombox paths,
                    width=\imagewidth / \pgfkeysvalueof{/tikz/zoomboxarray columns} - (\pgfkeysvalueof{/tikz/zoomboxarray columns} - 1) / \pgfkeysvalueof{/tikz/zoomboxarray columns} * \pgfkeysvalueof{/tikz/zoomboxarray inner gap} -\pgflinewidth,
                    height=\imageheight / \pgfkeysvalueof{/tikz/zoomboxarray rows} - (\pgfkeysvalueof{/tikz/zoomboxarray rows} - 1) / \pgfkeysvalueof{/tikz/zoomboxarray rows} * \pgfkeysvalueof{/tikz/zoomboxarray inner gap}-\pgflinewidth,
                    magnification=3,
                    every spy on node/.style={
                        zoombox paths
                    },
                    every spy in node/.style={
                        zoombox paths
                    }
                }
            ]
        },
        execute at end picture={
            \end{scope}
     \gdef\patternnumber{0}
        },
        spymargin/.initial=0.5em,
        zoomboxes xshift/.initial=1,
        zoomboxes right/.code=\pgfkeys{/tikz/zoomboxes xshift=1},
        zoomboxes left/.code=\pgfkeys{/tikz/zoomboxes xshift=-1},
        zoomboxes yshift/.initial=0,
        zoomboxes above/.code={
            \pgfkeys{/tikz/zoomboxes yshift=1},
            \pgfkeys{/tikz/zoomboxes xshift=0}
        },
        zoomboxes below/.code={
            \pgfkeys{/tikz/zoomboxes yshift=-1},
            \pgfkeys{/tikz/zoomboxes xshift=0}
        },
        caption margin/.initial=.5ex,
    },
    adjust caption spacing/.code={},
    image container/.style={
        inner sep=0pt,
        at=(image.north),
        anchor=north,
        adjust caption spacing
    },
    zoomboxes container/.style={
        inner sep=0pt,
        at=(image.north),
        anchor=north,
        name=zoomboxes container,
        xshift=\pgfkeysvalueof{/tikz/zoomboxes xshift}*(\imagewidth+\pgfkeysvalueof{/tikz/spymargin}),
        yshift=\pgfkeysvalueof{/tikz/zoomboxes yshift}*(\imageheight+\pgfkeysvalueof{/tikz/spymargin}+\pgfkeysvalueof{/tikz/caption margin}),
        adjust caption spacing
    },
    calculate dimensions/.code={
        \pgfpointdiff{\pgfpointanchor{image}{south west} }{\pgfpointanchor{image}{north east} }
        \pgfgetlastxy{\imagewidth}{\imageheight}
        \global\let\imagewidth=\imagewidth
        \global\let\imageheight=\imageheight
        \gdef\columncount{1}
        \gdef\rowcount{1}
        
    },
    image node/.style={
        inner sep=0pt,
        name=image,
        anchor=south west,
        append after command={
            [calculate dimensions]
            node [image container,subfigurename=\pgfkeysvalueof{/tikz/figurename}-image] {\phantomimage}
            node [zoomboxes container,subfigurename=\pgfkeysvalueof{/tikz/figurename}-zoom] {\phantomimage}
        }
    },
    color code/.style={
        zoombox paths/.append style={draw=#1}
    },
    connect zoomboxes/.style={
    spy connection path={\draw[draw=none,zoombox paths] (tikzspyonnode) -- (tikzspyinnode);}
    },
    help grid code/.code={
        \begin{scope}[
                x={(image.south east)},
                y={(image.north west)},
                font=\footnotesize,
                help lines,
                overlay
            ]
            \foreach \x in {0,1,...,9} { 
                \draw(\x/10,0) -- (\x/10,1);
                \node [anchor=north] at (\x/10,0) {0.\x};
            }
            \foreach \y in {0,1,...,9} {
                \draw(0,\y/10) -- (1,\y/10);                        \node [anchor=east] at (0,\y/10) {0.\y};
            }
        \end{scope}    
    },
    help grid/.style={
        append after command={
            [help grid code]
        }
    },
}

\newcommand\phantomimage{%
    \phantom{%
        \rule{\imagewidth}{\imageheight}%
    }%
}
\newcommand\zoombox[2][]{
    \begin{scope}[zoombox paths]
        \pgfmathsetmacro\xpos{
            (\columncount-1)*(\imagewidth / \pgfkeysvalueof{/tikz/zoomboxarray columns} + \pgfkeysvalueof{/tikz/zoomboxarray inner gap} / \pgfkeysvalueof{/tikz/zoomboxarray columns} ) + \pgflinewidth
        }
        \pgfmathsetmacro\ypos{
            (\rowcount-1)*( \imageheight / \pgfkeysvalueof{/tikz/zoomboxarray rows} + \pgfkeysvalueof{/tikz/zoomboxarray inner gap} / \pgfkeysvalueof{/tikz/zoomboxarray rows} ) + 0.5*\pgflinewidth
        }
        \edef\dospy{\noexpand\spy [
            #1,
            zoombox paths/.append style={
                black and white pattern=\patternnumber
            },
            every spy on node/.append style={#1},
            x=\imagewidth,
            y=\imageheight
        ] on (#2) in node [anchor=north west] at ($(zoomboxes container.north west)+(\xpos pt,-\ypos pt)$);}
        \dospy
        \pgfmathtruncatemacro\pgfmathresult{ifthenelse(\columncount==\pgfkeysvalueof{/tikz/zoomboxarray columns},\rowcount+1,\rowcount)}
        \global\let\rowcount=\pgfmathresult
        \pgfmathtruncatemacro\pgfmathresult{ifthenelse(\columncount==\pgfkeysvalueof{/tikz/zoomboxarray columns},1,\columncount+1)}
        \global\let\columncount=\pgfmathresult
        \ifblackandwhitecycle
            \pgfmathtruncatemacro{\newpatternnumber}{\patternnumber+1}
            \global\edef\patternnumber{\newpatternnumber}
        \fi
    \end{scope}
}
\usepackage{booktabs}
\usepackage{xfrac}
\usepackage{tablefootnote}
\usepackage{colortbl}
\usepackage{tabularx}
\definecolor{rowblue}{RGB}{220,230,240}

\usepackage{gensymb}

\renewcommand{\paragraph}[1]{\vspace{8px} \noindent \textbf{#1} \ \ }

\usepackage[pagebackref=true,breaklinks=true,letterpaper=true,colorlinks,bookmarks=false]{hyperref}

\cvprfinalcopy % *** Uncomment this line for the final submission

\def\cvprPaperID{***} % *** Enter the CVPR Paper ID here

\ifcvprfinal\fi
\begin{document}

%%%%%%%%% TITLE
\title{Deep Video Deblurring}

\author{Shuochen Su\\
The Univerity of British Columbia
\and
Mauricio Delbracio\\
Universidad de la Rep\'ublica
\and
Jue Wang\\
Adobe Research
\and
Guillermo Sapiro\\
Duke University
\and
Wolfgang Heidrich\\
KAUST
\and
Oliver Wang\\
Adobe Research
}

\maketitle
%\thispagestyle{empty}

%%%%%%%%% ABSTRACT
\begin{abstract}
Motion blur from camera shake is a major problem in videos captured by hand-held 
devices. Unlike single-image deblurring, video-based approaches can take
advantage of the abundant information that exists across neighboring frames.
As a result the best performing methods rely on aligning nearby frames.
However, aligning images is a computationally expensive and fragile procedure, and methods that aggregate
information must therefore be able to identify which regions have been
accurately aligned and which have not, a task which requires high level scene understanding. 
In this work, we introduce a deep learning solution to video deblurring,
where a CNN is trained end-to-end to learn how to accumulate information
across frames. To train this network, we collected a dataset of real videos
recorded with a high framerate camera, which we use to generate synthetic
motion blur for supervision. We show that the features learned from this
dataset extend to deblurring motion blur that arises due to camera shake in a
wide range of videos, and compare the quality of results to a number of other
baselines.
\end{abstract}

%%%%%%%%% BODY TEXT
\section{Introduction}
\begin{figure}[t]
\centering	
\scriptsize
\newcommand{\bike}[1]{\includegraphics[width=.18\linewidth,clip,trim=210 590 960 30]{#1}}
\newcommand{\people}[1]{\includegraphics[width=.18\linewidth,clip,trim=150 180 1550 700]{#1}}
\def\arraystretch{0.5}%
\begin{tabular}{*{6}{c@{\hspace{1.5px}}}}
\begin{sideways}\hspace{1.5em}input\end{sideways}& 
	\bike{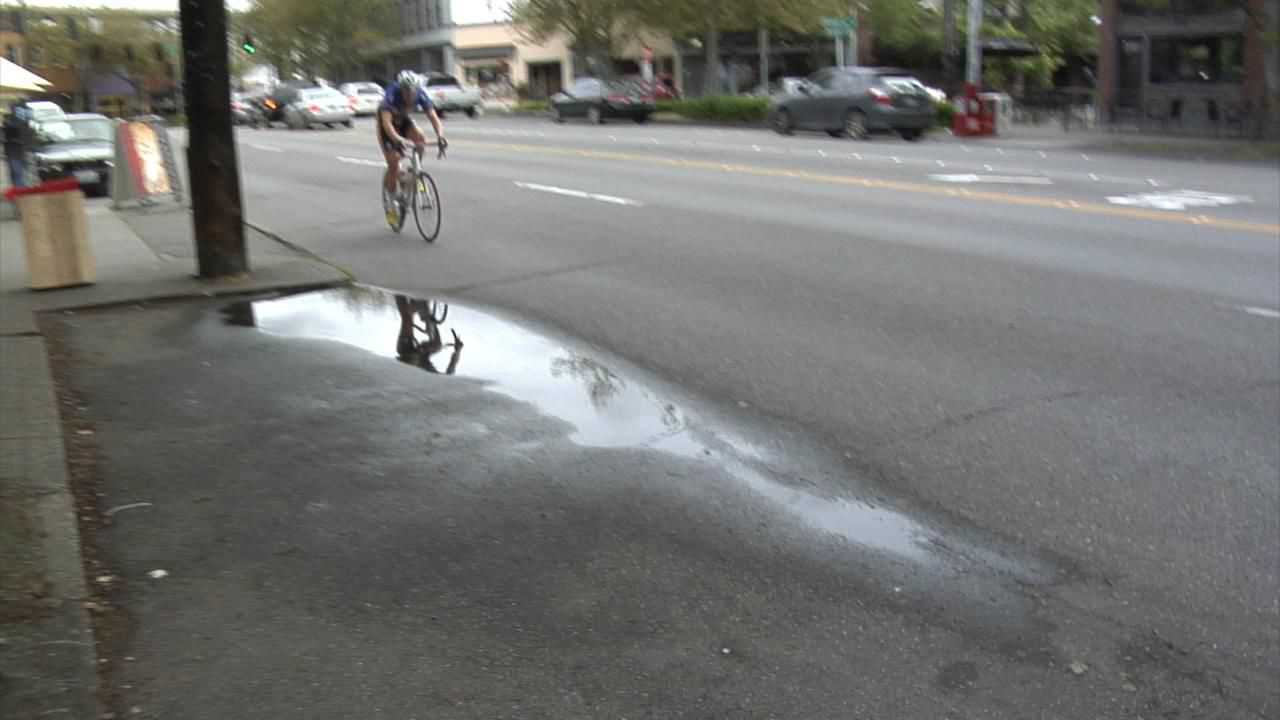} &
	\bike{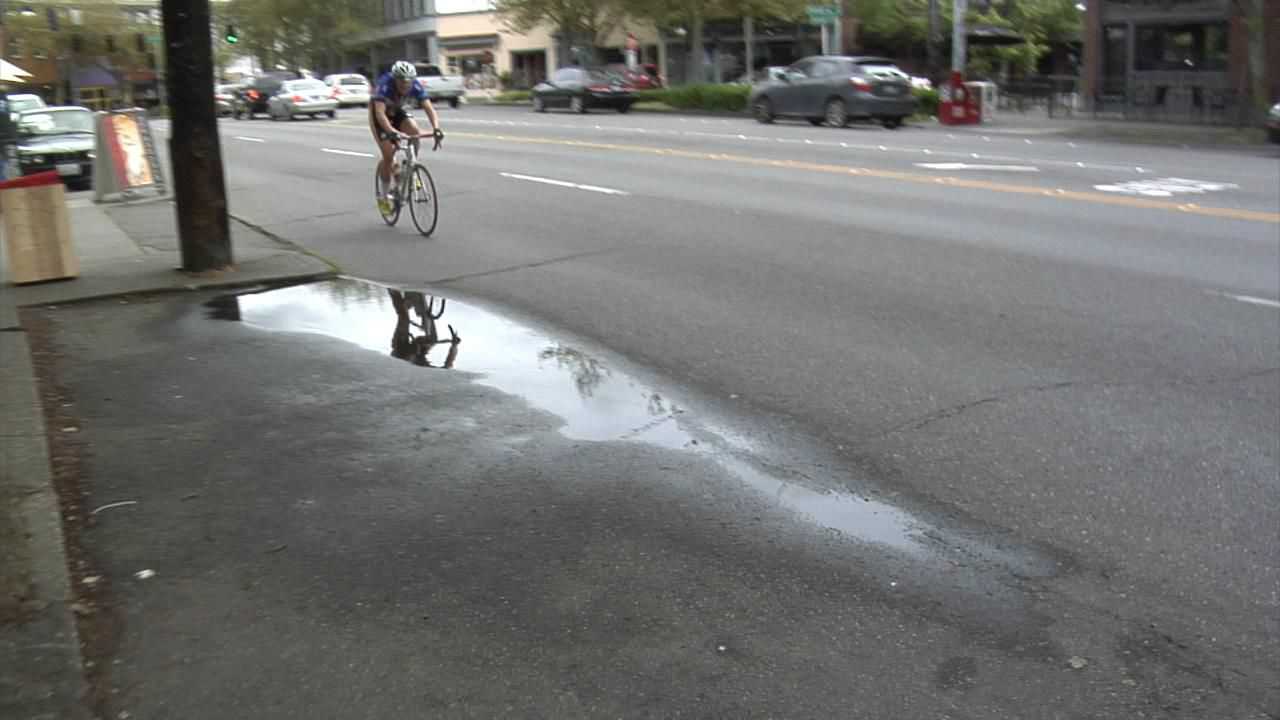} &
	\bike{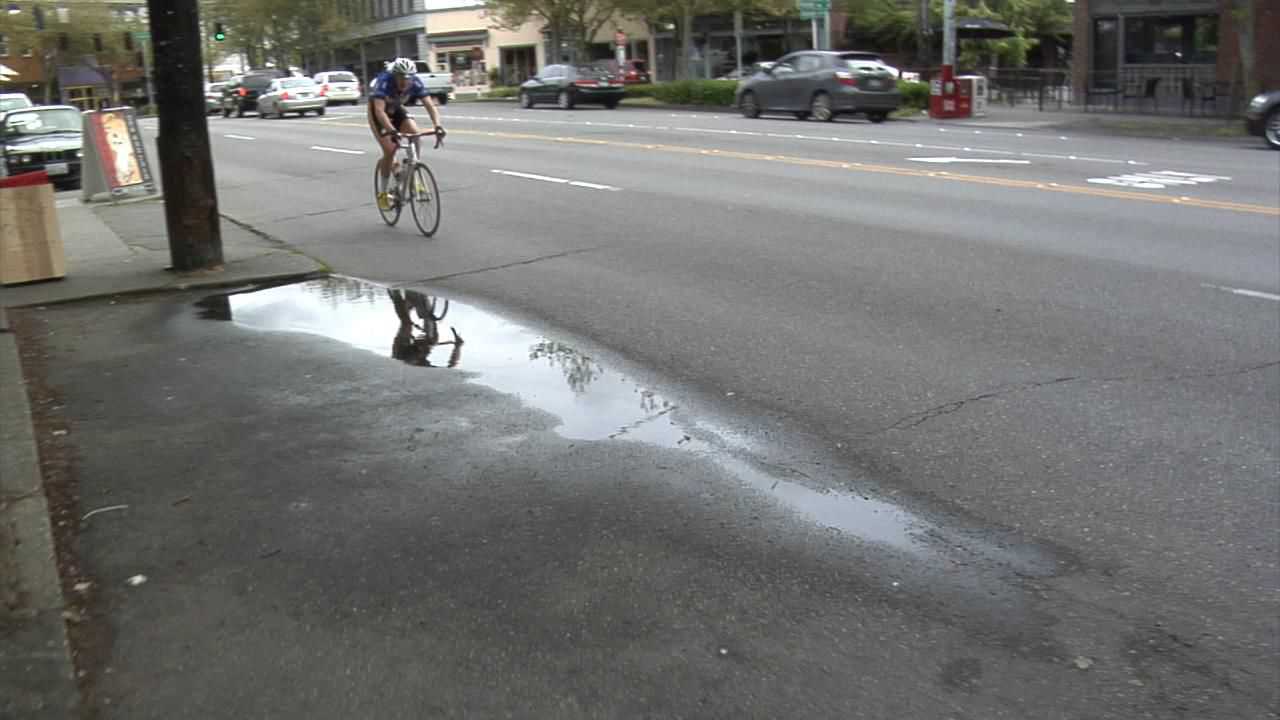} &
	\bike{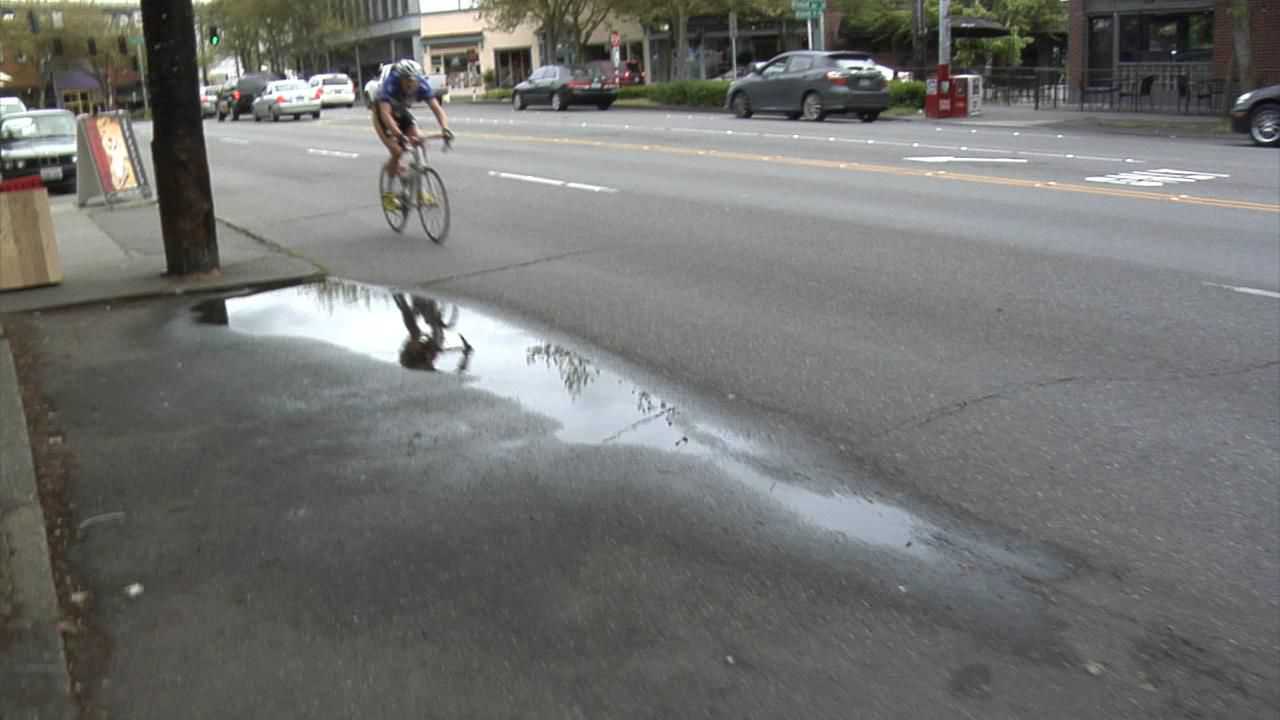} &
	\bike{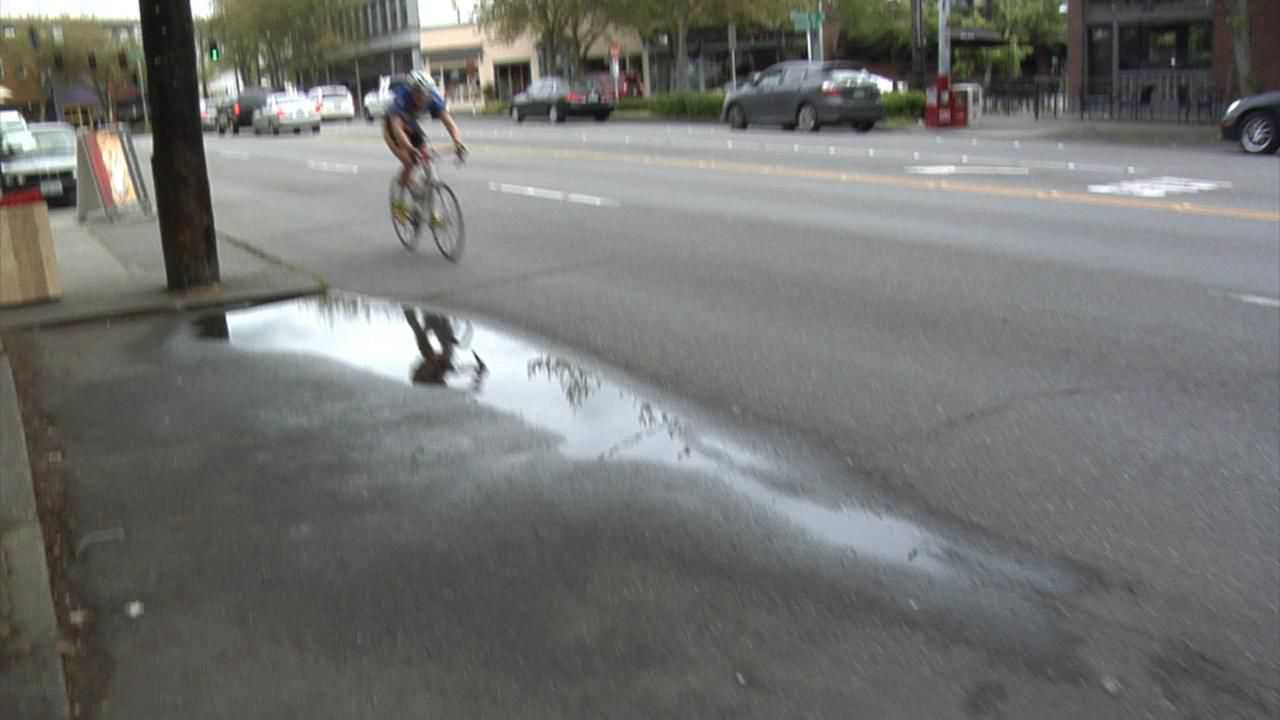} \\
\begin{sideways}\hspace{1.0em}\textbf{proposed}\end{sideways}& 
	\bike{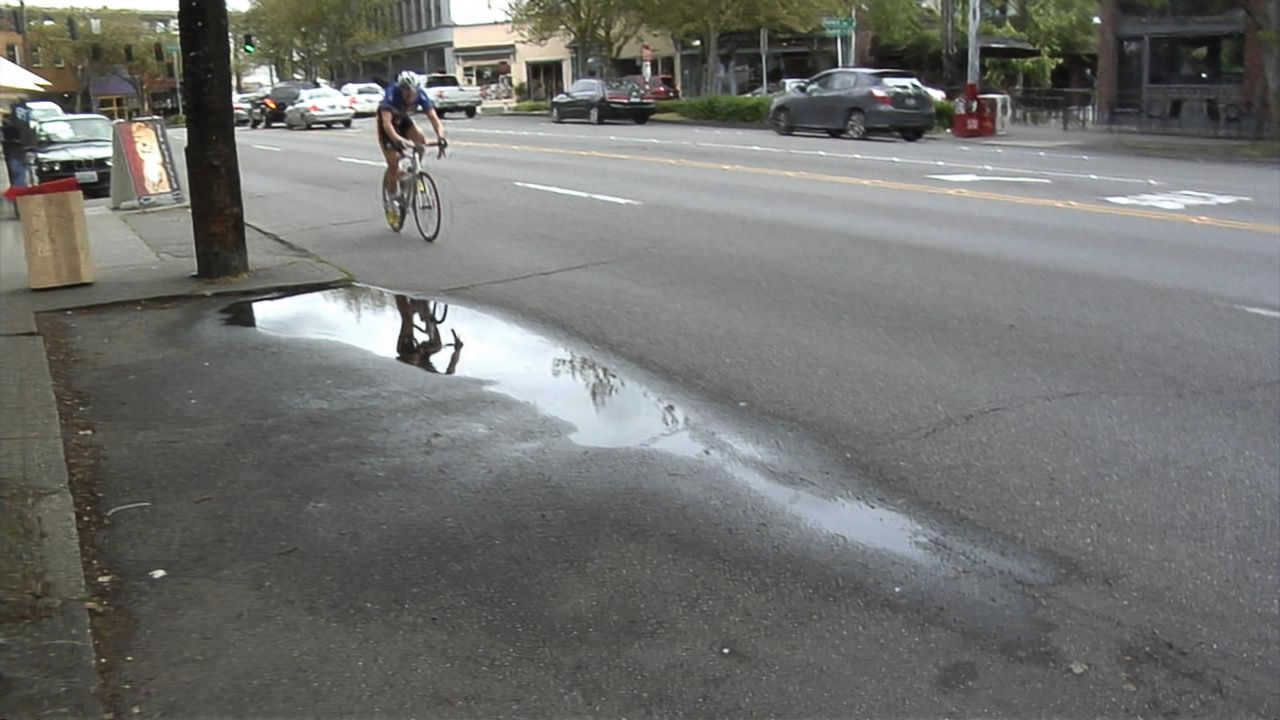} &
	\bike{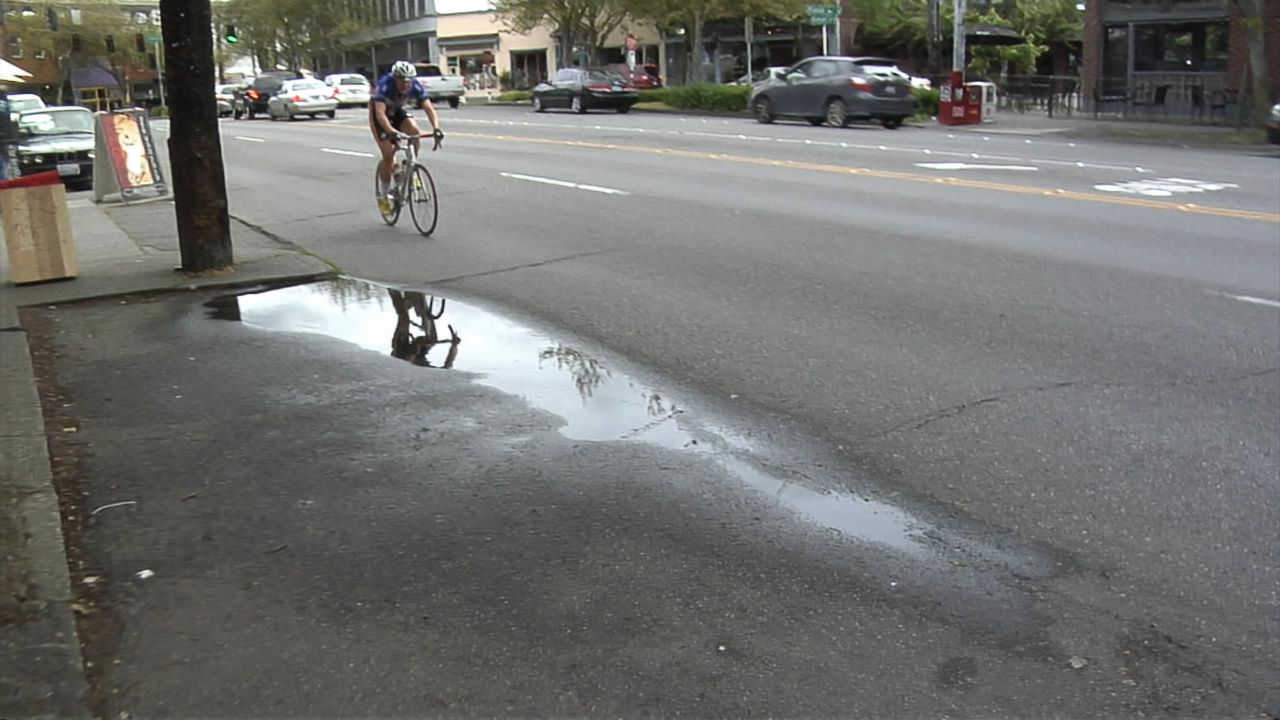} &
	\bike{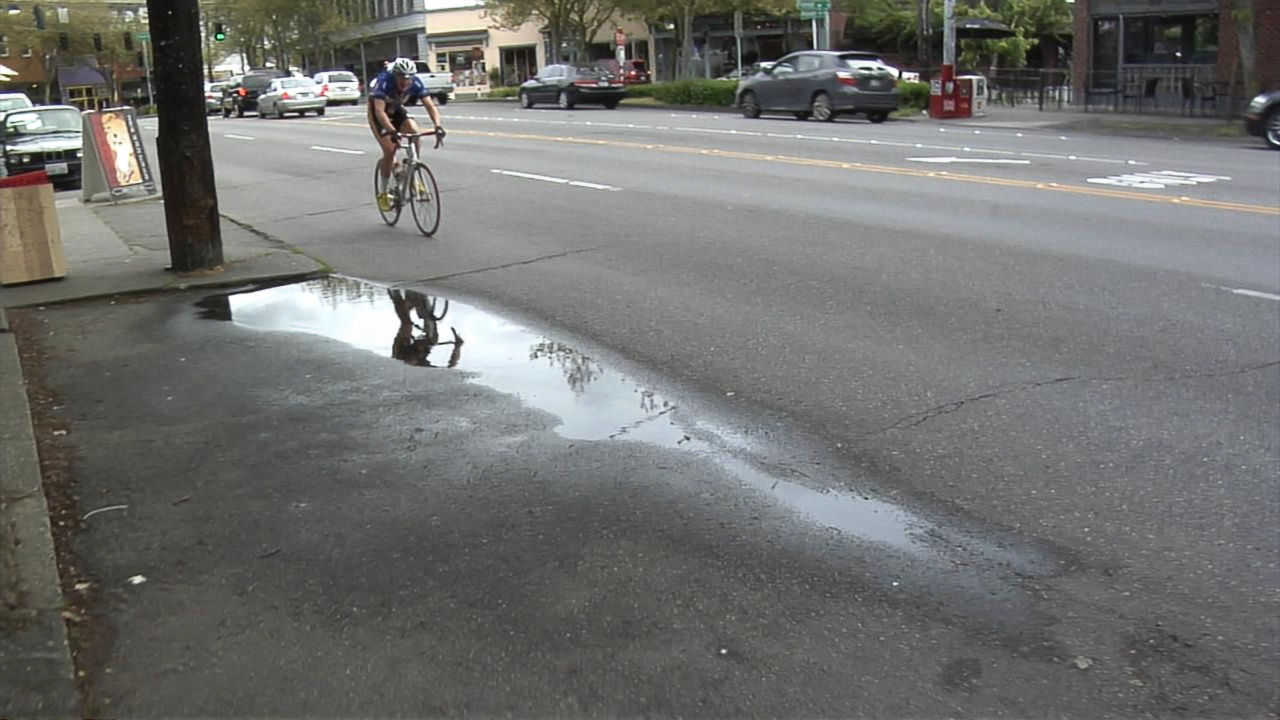} &
	\bike{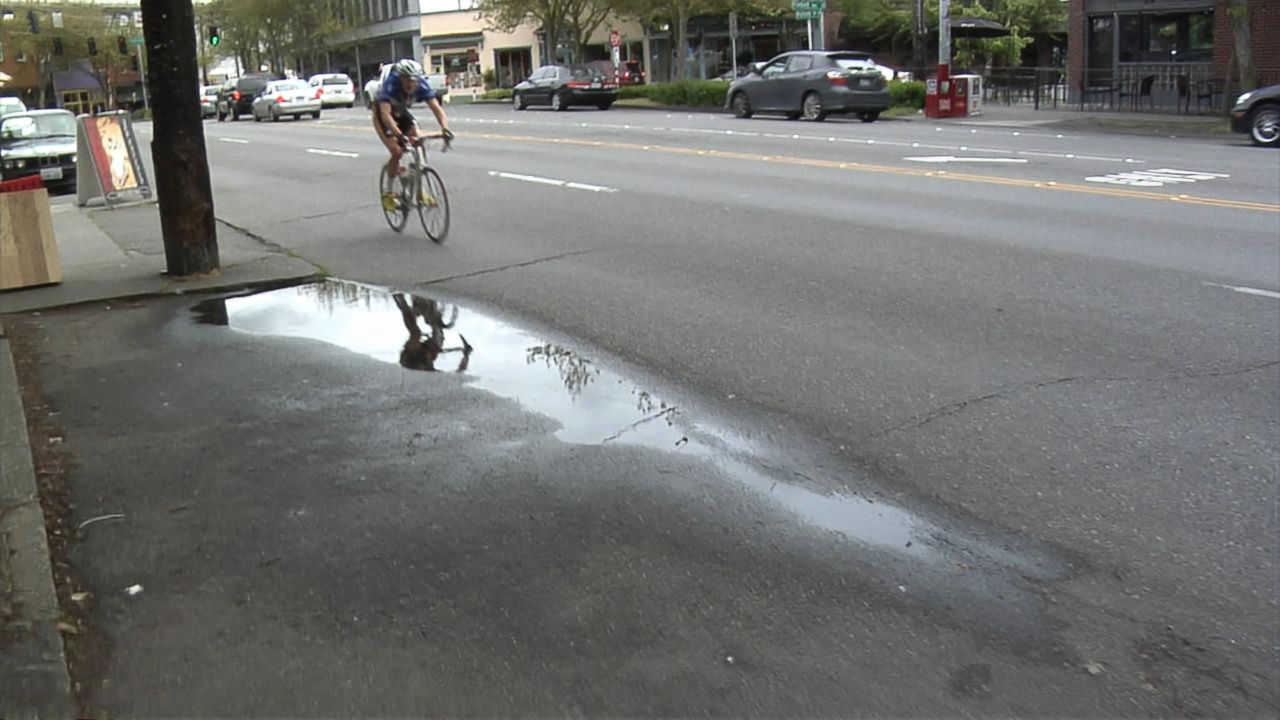} &
	\bike{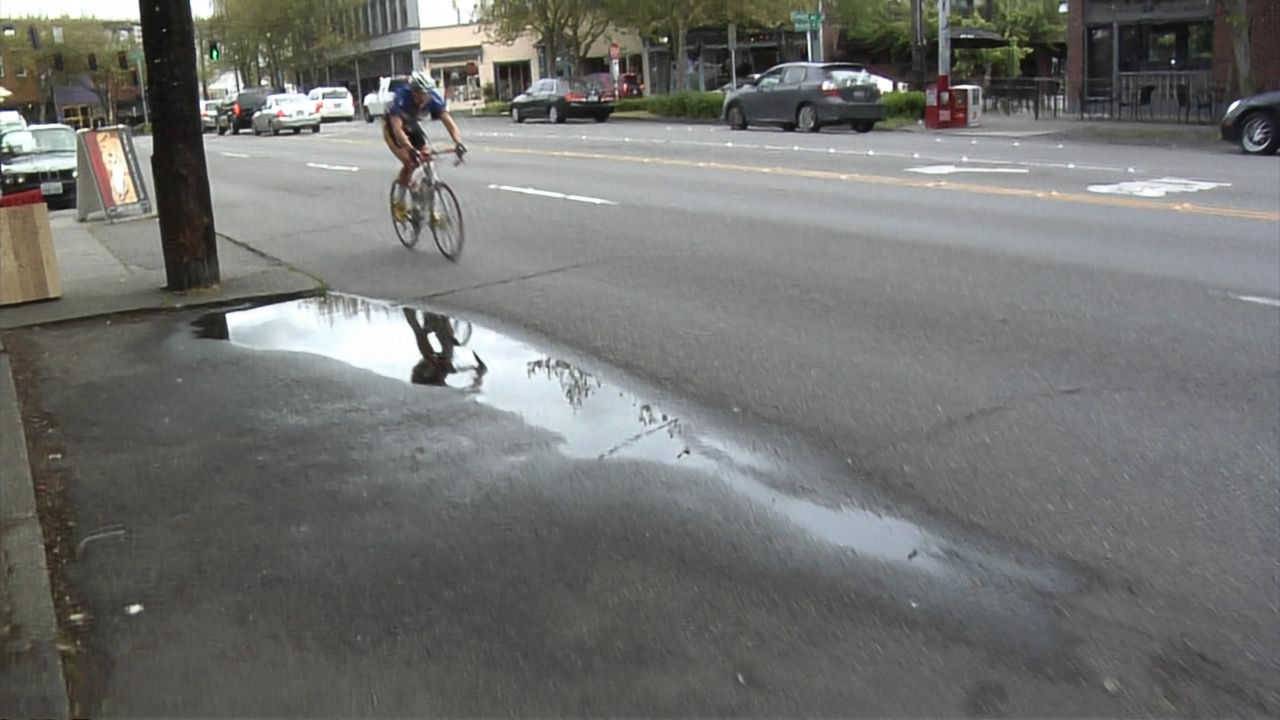} \\
\end{tabular}
\vspace{2em}
\begin{tabular}{*{6}{c@{\hspace{1.5px}}}}
\begin{sideways}\hspace{1.5em}input\end{sideways}& 
	\people{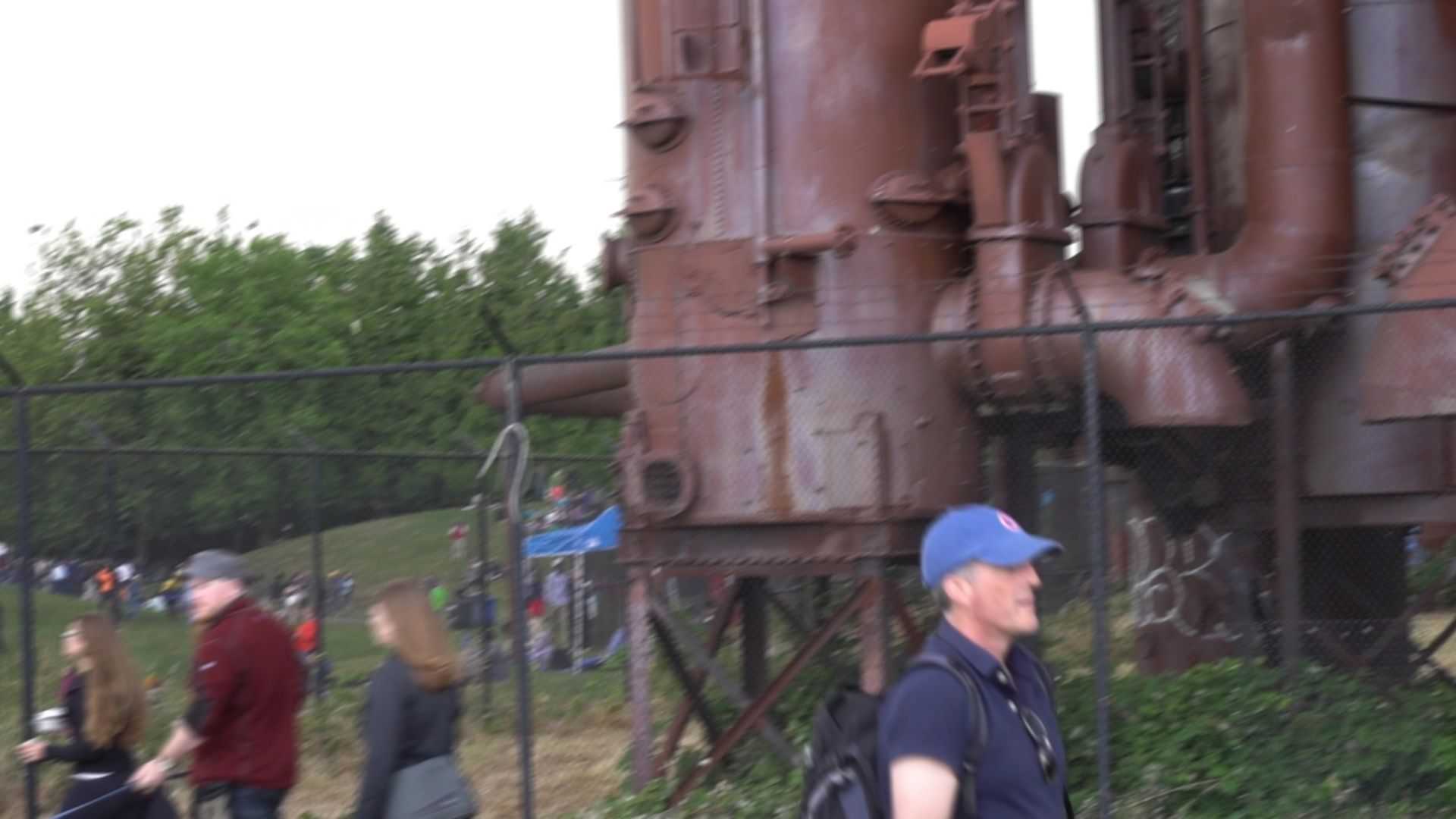} &
	\people{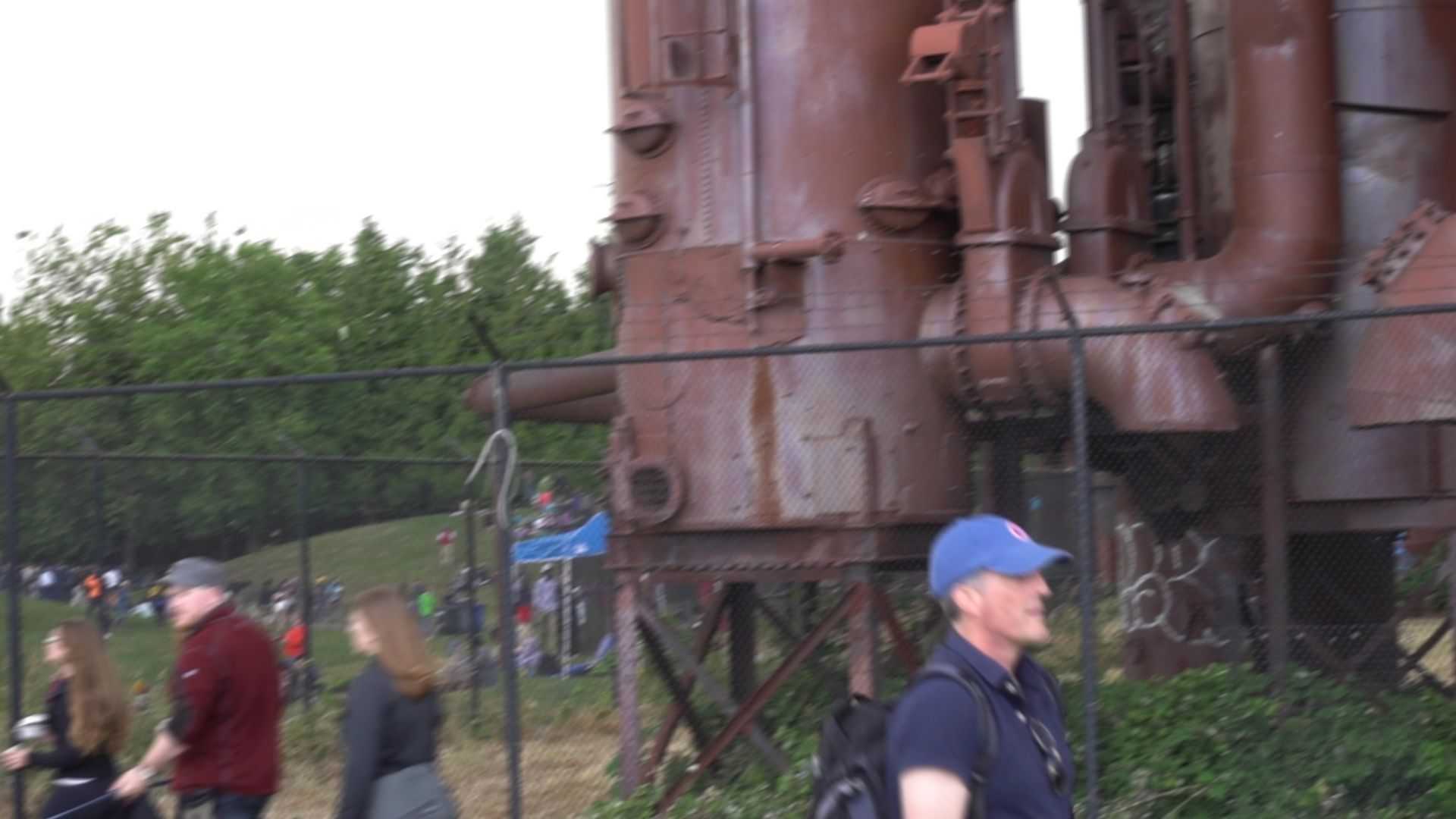} &
	\people{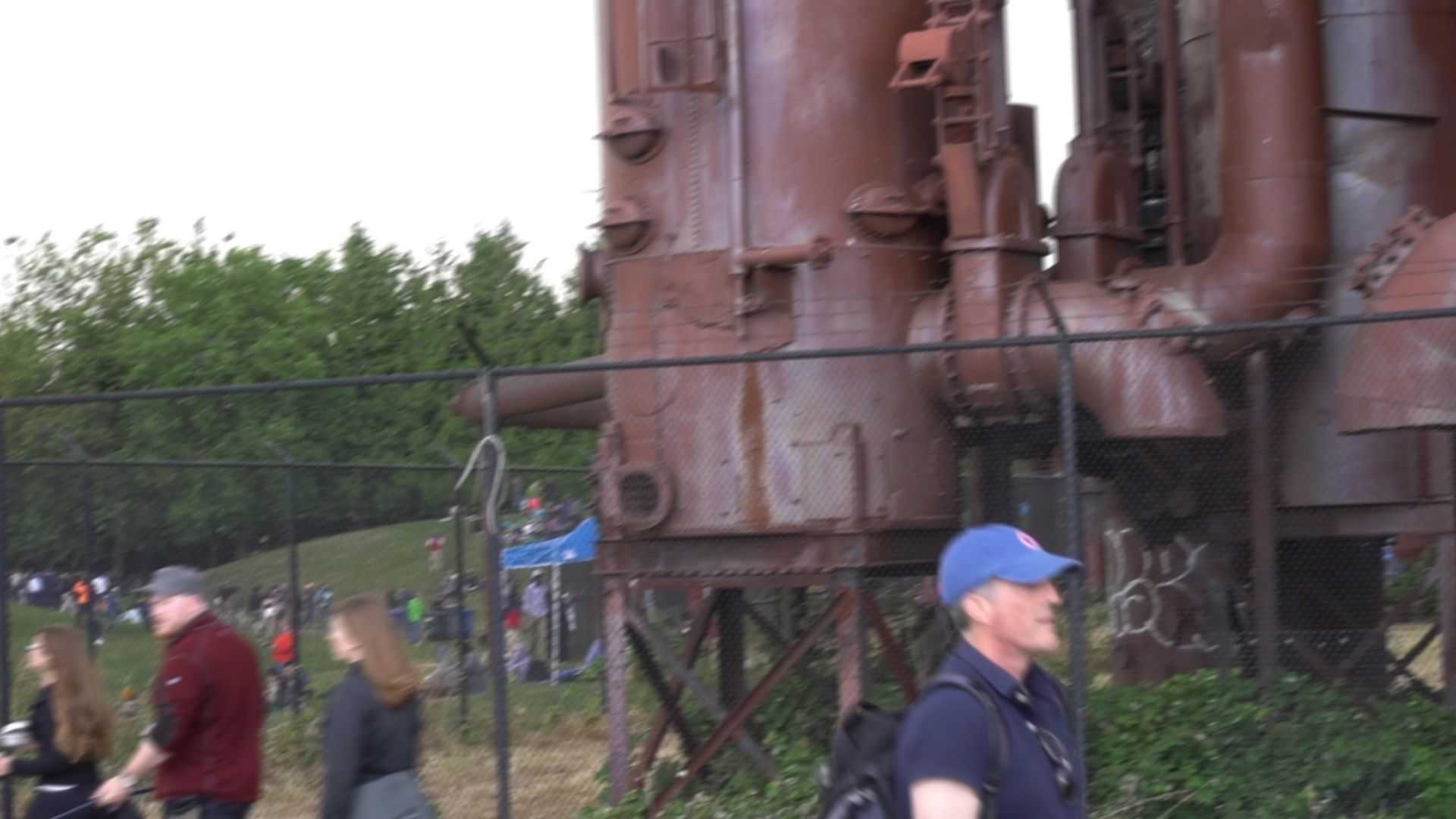} &
	\people{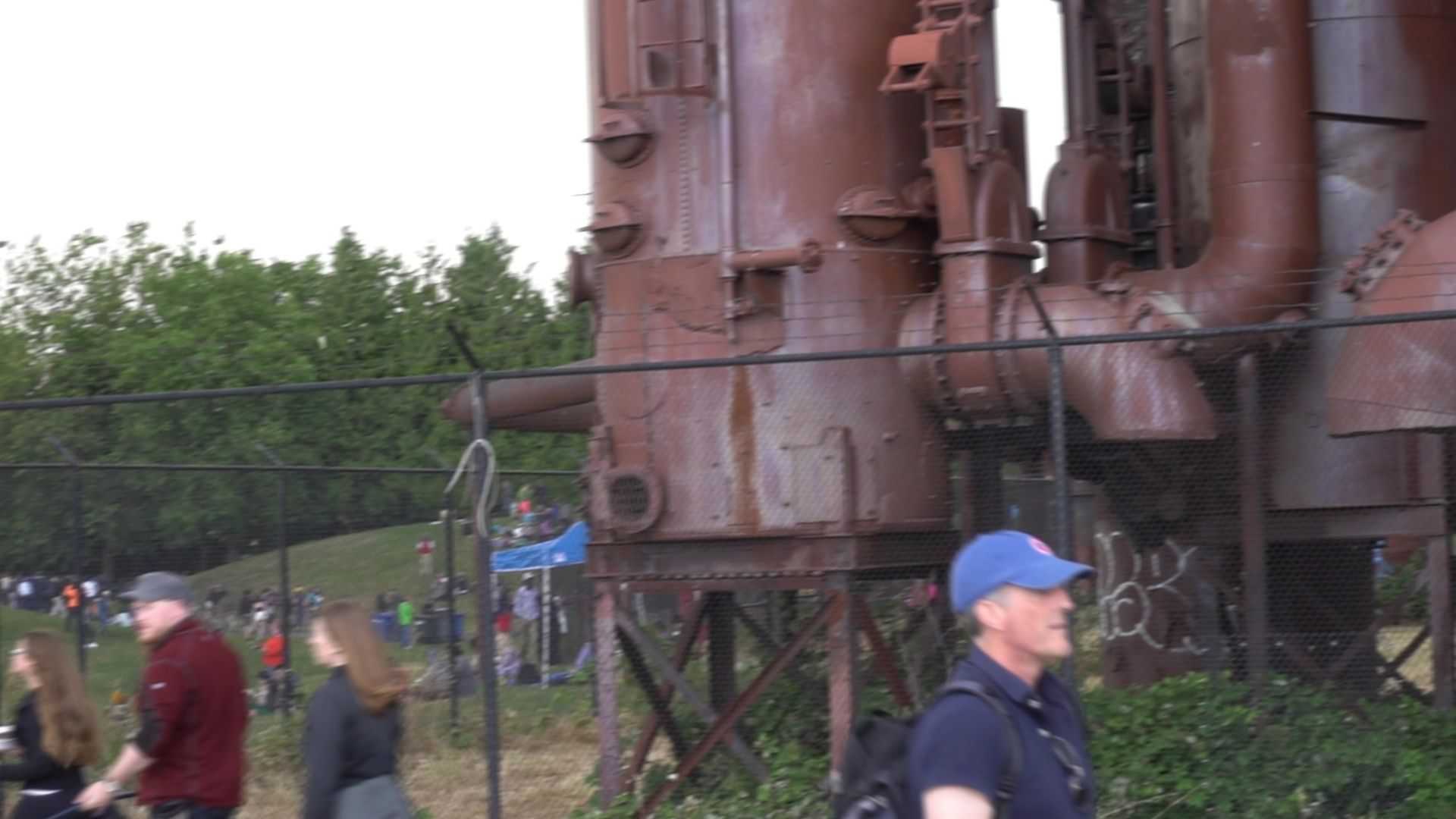} &
	\people{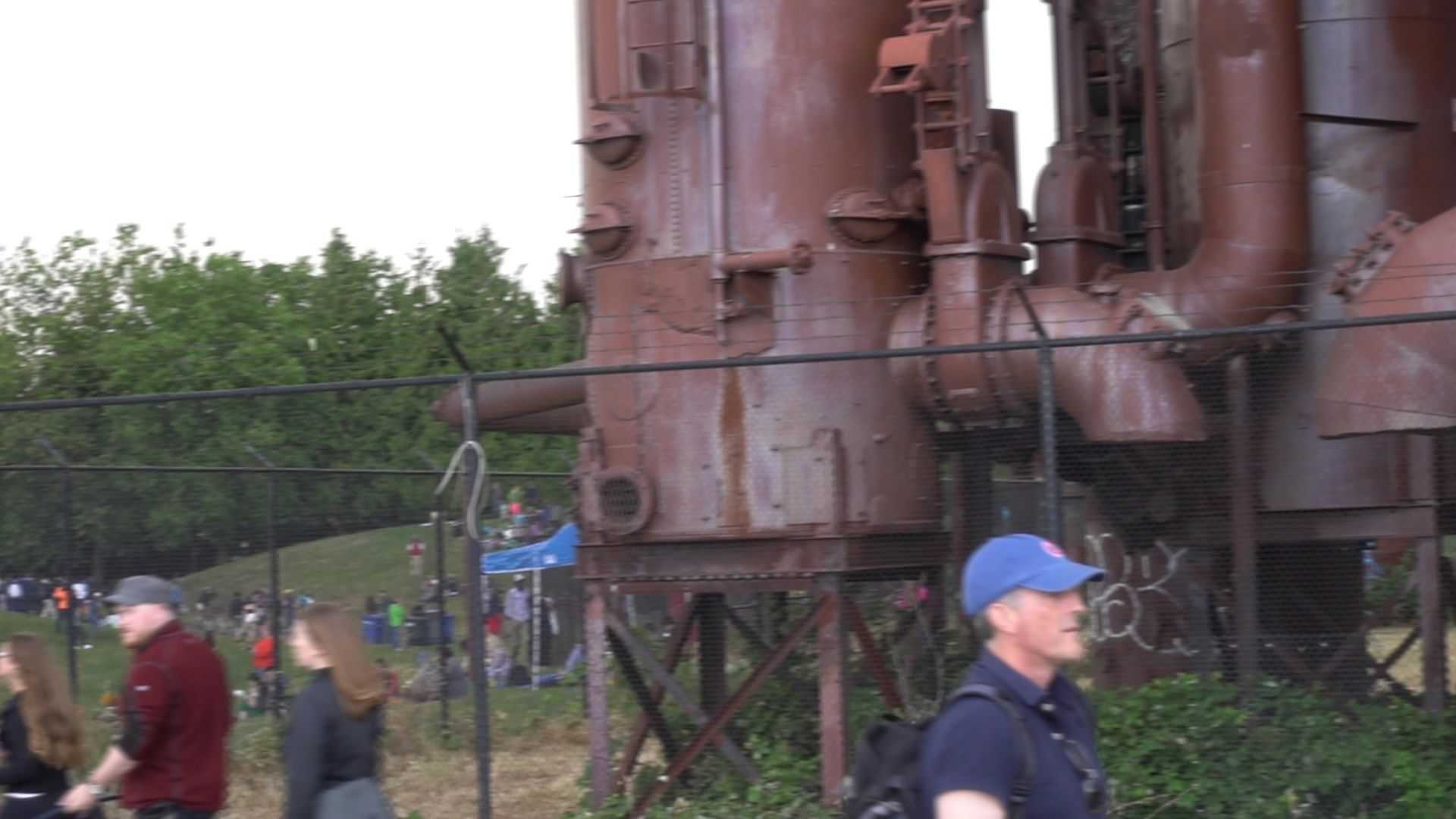} \\
\begin{sideways}\hspace{1.0em}\textbf{proposed}\end{sideways}& 
	\people{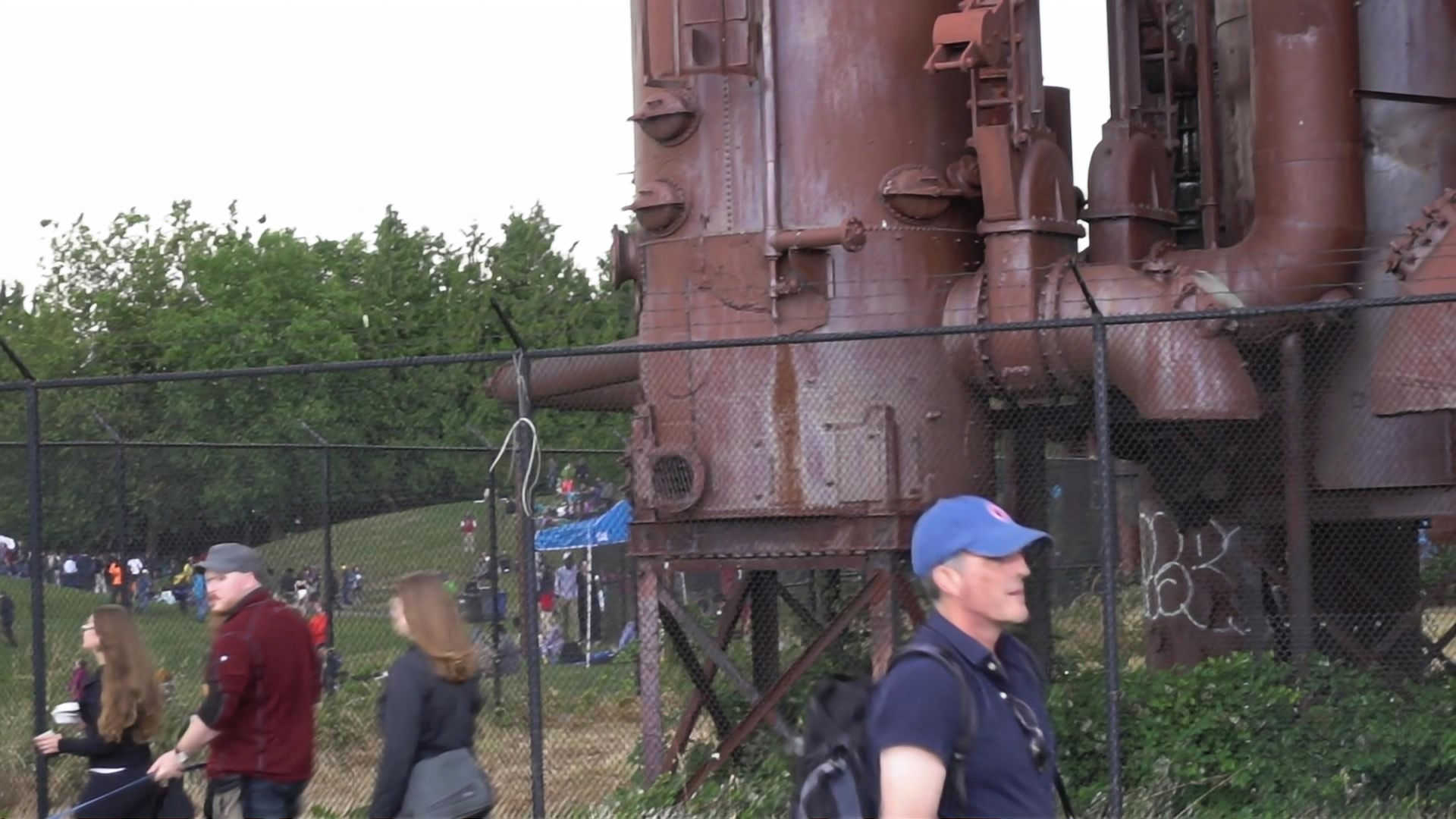} &
	\people{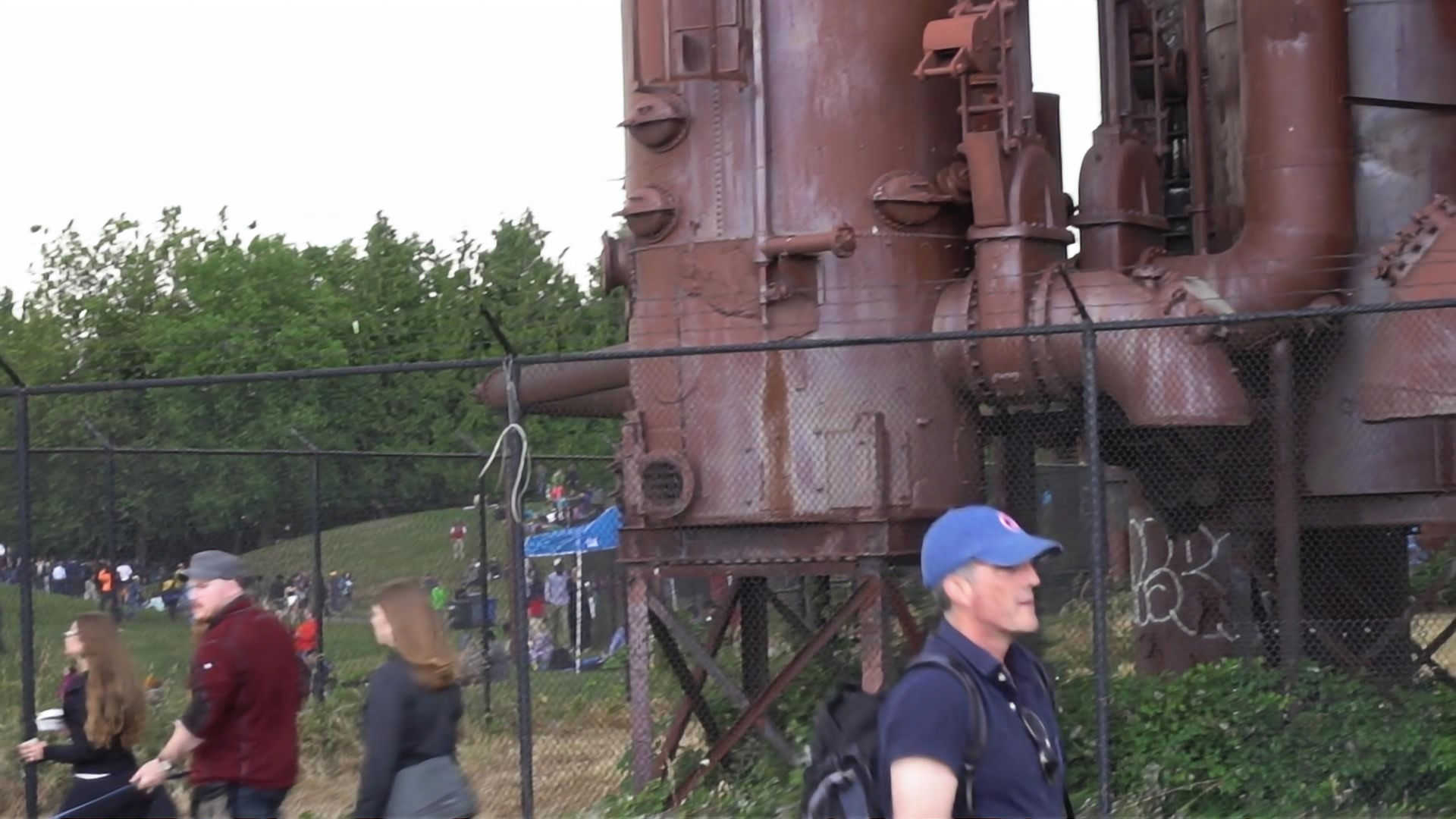} &
	\people{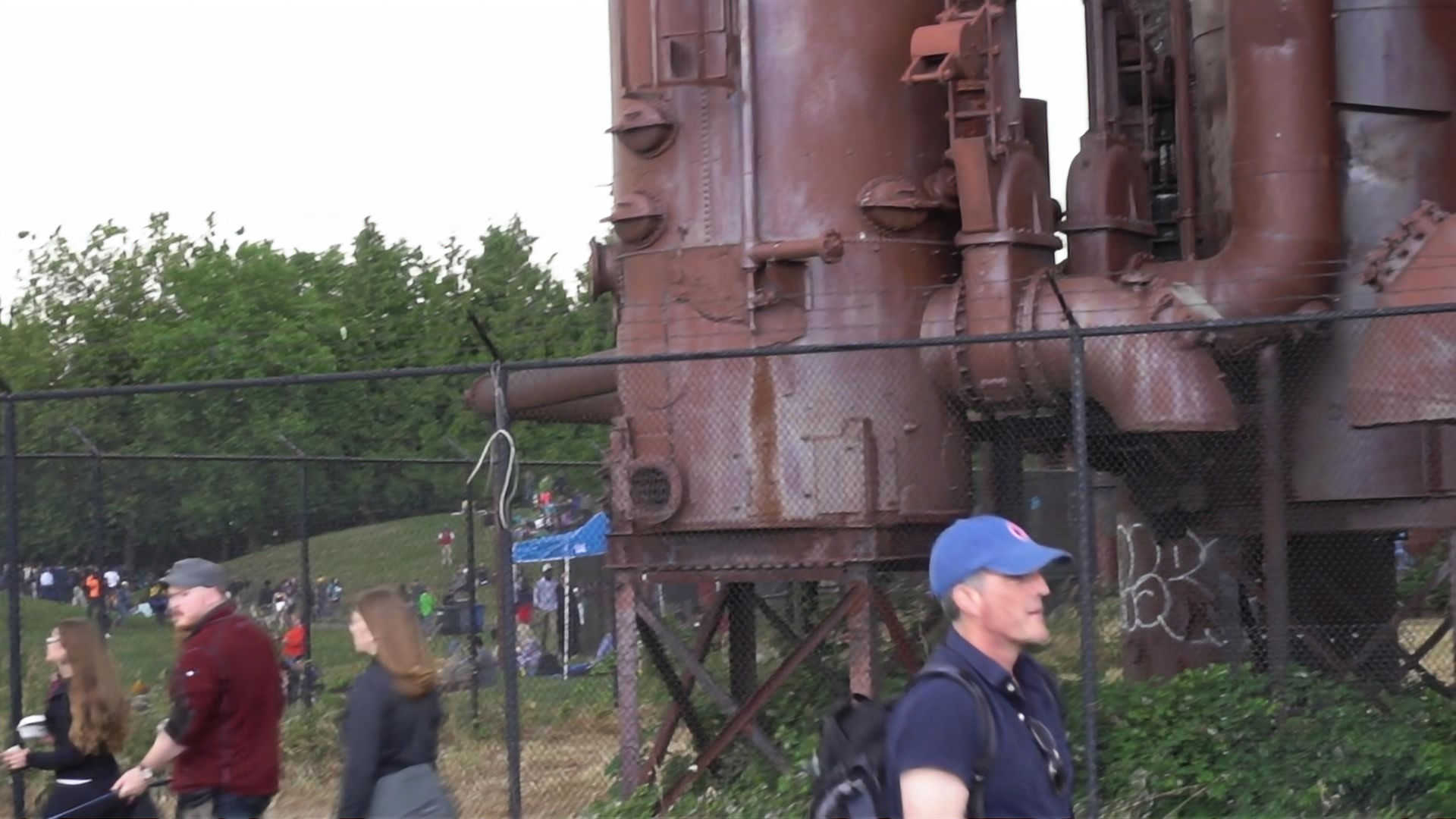} &
	\people{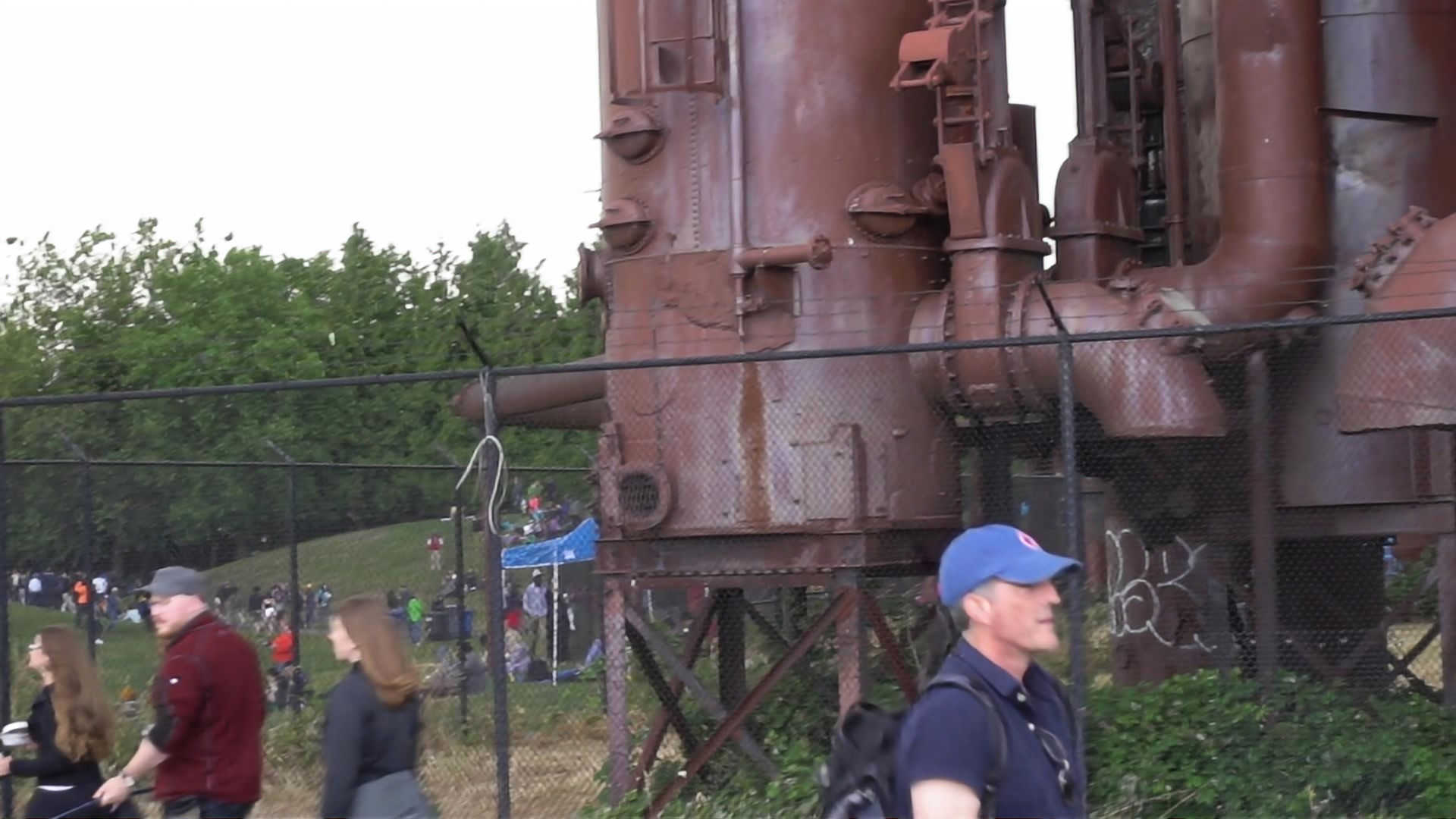} &
	\people{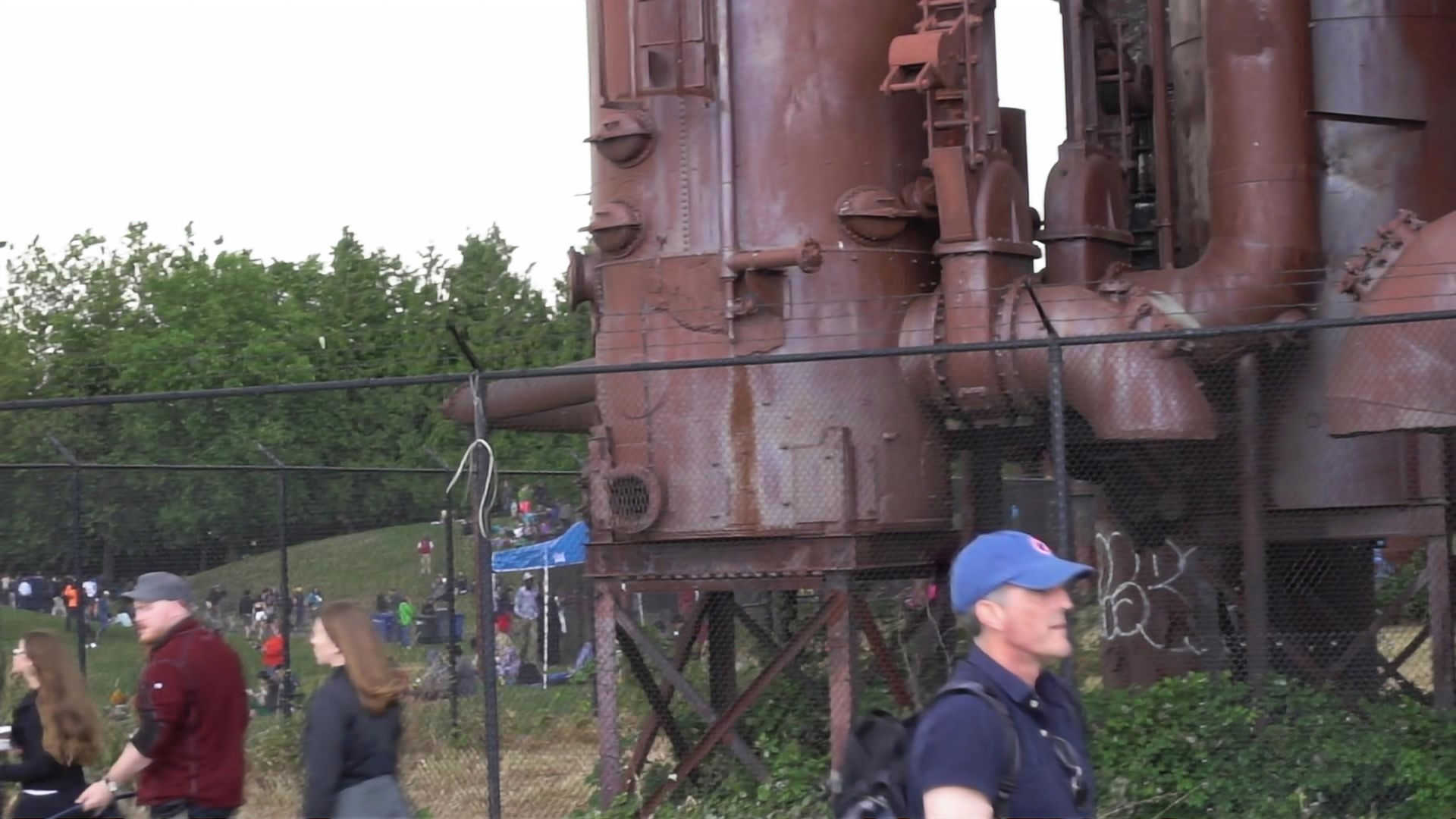} 
\end{tabular} \vspace{-1em}
\caption{Blur in videos can be significantly attenuated by learning
how to aggregate information from nearby frames\protect\footnotemark.
 Top: crops of consecutive frames from a shake blurry video; 
Bottom: the output from the proposed data-driven approach, using simple frame-wise homography alignment.}
\label{fig:teaser} 
\end{figure}

\footnotetext{Please see the supplemental videos for more results: \url{https://www.cs.ubc.ca/~shuochsu/deepvideodeblurring/videos.zip}}

Handheld video capture devices are now commonplace
As a result, video stabilization has become an essential step in video capture pipelines, often
performed automatically at capture time (e.g., iPhone, Google Pixel), or as a
service on sharing platforms (e.g., Youtube, Facebook). While stabilization
techniques have improved dramatically, the remaining motion blur is a major
problem with all stabilization techniques. 
This is because the blur becomes obvious when there is no motion to accompany it, yielding highly visible
``jumping'' artifacts. In the end,
the remaining camera shake motion blur limits the amount of stabilization that can be applied
before these artifacts become too apparent.

The most successful video deblurring approaches use information from
neighboring frames to sharpen blurry frames, taking advantage of the fact that
most hand-shake motion blur is both short and temporally uncorrelated. By
borrowing ``sharp" pixels from nearby frames, it is possible to
reconstruct a high quality output. Previous work has shown significant
improvement over traditional deconvolution-based deblurring approaches, via
patch-based synthesis that relies on either lucky imaging~\cite{cho2012video}
or weighted Fourier aggregation~\cite{delbracio2015hand}.

One of the main challenges associated with aggregating information
across multiple video frames is that the differently blurred frames must be aligned. This can
either be done, for example, by nearest neighbor patch
lookup~\cite{cho2012video}, or optical flow~\cite{delbracio2015hand}. However,
warping-based alignment is not robust around disocclusions and
areas with low texture, and often yields warping artifacts. 
In addition to the alignment computation cost, methods that rely on warping have to therefore disregard information 
from mis-aligned content or warping artifacts, which can be hard by looking at local image patches alone. 

To this end, we present the first end-to-end data driven approach to video deblurring, the results of which can be seen in Fig.~\ref{fig:teaser}.  
We address specifically blur that arises due to camera shake, e.g., is temporally uncorrelated, however we show that our deblurring extends to other types of blur as well, including motion blur from object motion.
We also show that by using an autoencoder-type network with skip
connections, we can create high quality results \emph{without} computing any
alignment or image warping, which makes our approach highly efficient and robust to
scene types. We present results with a number of different learned configurations based on different alignment types: no-alignment, homography, and optical flow alignment.
On average optical flow performs the best, although in many cases using a projective transform (i.e. homography) performs comparably with significantly less computation required, and we can even get good results without any alignment at all. 

Our main contribution is an end-to-end solution to train a deep neural network
to learn how to deblur images, given a short stack of neighboring video
frames. We describe the architecture we found to give the best results, and
the method we used to create a real-world dataset using high framerate
capture. 
We compare qualitatively to videos
previously used for video deblurring, and quantitatively with our ground truth
data set. We also present a test set of videos that shows that our method
generalizes to a wide range of videos. Both datasets will be made available to
the public to encourage follow up work.

\section{Related Work}
There exist two main approaches to deblurring: deconvolution-based methods that solve inverse problems, and methods that rely on multi-image aggregation/fusion. 
\vspace{.5em}
\noindent \textbf{Deblur using deconvolution.}
Modern single-image deblurring approaches jointly estimate a blurring kernel (either single or spatially varying) and the underlying sharp image via deconvolution~\cite{kundur1996blind}.
In recent years many successful methods have been introduced~\cite{fergus2006removing,shan2008high,cho2009fast,xu2010two,krishnan2011blind,xu2013unnatural,michaeli2014blind,su2015rolling}, see~\cite{wang2014recent} for a recent survey.
Multiple-image deconvolution methods use additional information to alleviate the severe  ill-posedness of single image deblurring.  
These approaches collect, for example, image bursts~\cite{ito2014blurburst}, blurry-noisy pairs~\cite{yuan2007image}, flash no-flash image pairs~\cite{petschnigg2004digital}, gyroscope information~\cite{park2014gyro}, high framerate sequences~\cite{tai2008image}, or stereo pairs~\cite{sellent2016stereo} for deblurring.
These methods generally assume static scenes and require the input images to be aligned. 
For video, temporal information~\cite{li2010generating}, optical flow~\cite{kim2015cvpr} and scene models~\cite{paramanand2013non,wulff2014modeling} have been used for improving both kernel and latent frame estimation.

All of the above approaches strongly rely on the accuracy of the assumed image degradation model (blur, motion, noise) and its estimation, thus may perform poorly when the simplified degradation models are insufficient to describe real data, or due to poor model estimation.
As a result, these approaches tend to be more fragile than aggregation-based methods~\cite{delbracio2015hand}, and often introduce undesirable artifacts such as ringing and amplified noise.

\vspace{.5em}
\noindent \textbf{Multi-image aggregation.} 
Multi-image aggregation methods directly combine multiple images in either spatial or frequency domain without solving any inverse problem. 
Lucky-imaging is a classic example, in which multiple low quality images are aligned and best pixels from different ones are selected and merged into the final result~\cite{law2006lucky,joshi2010seeing}.
For denoising, this has been extended to video using optical flow~\cite{liu2010high} or piecewise homographies~\cite{liu2014fast} for alignment.

For video deblurring, aggregation approaches rely on the observation that in general not all video frames are equally blurred.
Sharp pixels thus can be transferred from nearby frames to deblur the target frame, using for example homography alignment~\cite{matsushita2006full}. 
Cho et al. further extend this approach using patch-based alignment~\cite{cho2012video} for improved robustness against moving objects. The method however cannot handle large depth variations due to the underlying homography motion model, and the patch matching process is computationally expensive.
Klose et al.~\cite{klose2015sampling} show that 3D reconstruction can be used to project pixels into a single reference coordinate system for pixel fusion.  
Full 3D reconstruction however can be fragile for highly dynamic videos.

Recently, Delbracio and Sapiro~\cite{delbracio2015cvpr} show that aggregating multiple aligned images in the Fourier domain can lead to effective and computationally highly efficient deblurring.  
This technique was extended to video~\cite{delbracio2015hand}, where nearby frames are warped via optical flow for alignment. 
This method is limited by optical flow computation and evaluation, which is not reliable near occlusions and outliers. 

All above approaches have explicit formulations on how to fuse multiple images.
In this work, we instead adopt a data-driven approach to \emph{learn} how multiple images should be aggregated to generate an output that is as sharp as possible.

\begin{figure*}[t]
	\centering
	\includegraphics[width=0.95\linewidth]{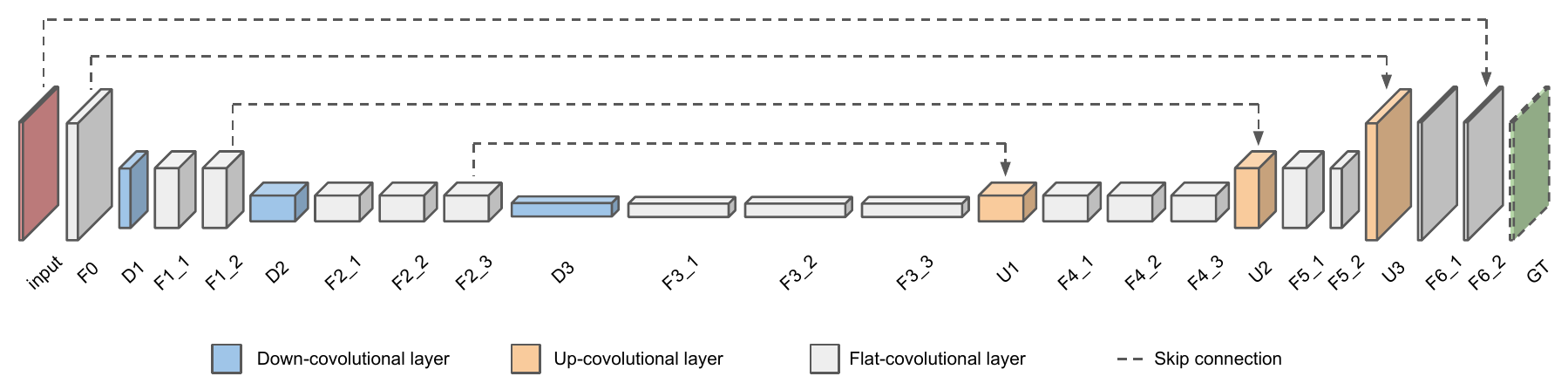}
	\caption{Architecture of the proposed \textsl{DeBlurNet} model, that takes the stacked nearby frames as input, and processes them jointly through a number of convolutional layers until outputing the deblurred central frame. The depth of each block represents the number of features after convolution.}
	\label{fig:architecture} 
\end{figure*}

\vspace{.5em}
\noindent \textbf{Data-driven approaches.} 
Recently, CNNs have been applied to achieve leading results on a wide variety of reconstruction problems. 
These methods tend to work best when large training datasets can be easily constructed, for example by adding synthetic noise for denoising~\cite{xie2012image}, removing content for inpainting~\cite{pathak2016context}, removing color information for colorization~\cite{iizuka2016let}, or downscaling for superresolution~\cite{dong2014learning,liu2016robust}. 
Super resolution networks have been applied to video sequences before~\cite{huang2015bidirectional, kappeler2016video, shi2016real}, but these approaches address a different problem, with its own set of challenges. 
In this work we focus on deblurring, where blurry frames can vary greatly in appearance from their neighbors, making information aggregation more challenging.

CNNs have also been used for single-image deblurring, using synthetic training data.~\cite{sun2015learning, chakrabarti2016neural}.
One problem with synthetic blur is that real blur has significantly different characteristics, as it depends on both the scene depth and object motion. 
In our experiments, we show that by leveraging multiple video frames, training on real blur, and directly estimating the sharp images, our method can produce better results than~\cite{chakrabarti2016neural}.

\section{Our Method}
\paragraph{Overview.}
Image alignment is inherently difficult as determining whether the aligned pixels in different images correspond to the same scene content can be difficult with only low-level features. 
High-level features, on the other hand, provide sufficient additional information to help separate incorrectly aligned image regions from correctly aligned ones. 
To make use of both low-level and high-level features, 
we therefore train an end-to-end system for video deblurring, where the input
is a stack of neighboring frames and the output is the deblurred
\emph{central} frame in the stack. 
Furthermore,  our network is trained using real video frames with realistically synthesized motion blur.
In the following, we first present our neural network architecture, then describe a number of experiments
for evaluating its effectiveness and comparing with existing methods. 
The key advantage of our method is the allowance of lessening the requirements for accurate alignment, a fragile component of prior work.

\subsection{Network Architecture}
\begin{table}[h]
\scriptsize
\centering
\begin{tabular}{@{}rccrc@{}}
\toprule
layer  & kernel size    & stride                                 & output size                                    & skip connection   \\ \midrule
input  & -              & -                                      & 15$\times H \times W$                          & to F6\_2$^{\ast}$ \\
F0     &  5$\times$5    & 1$\times$1                             & 64$\times H \times W$                          & to U3             \\ \midrule
D1     &  3$\times$3    & 2$\times$2                             & 64$\times \sfrac{H}{2} \times \sfrac{W}{2}$    & -                 \\
F1\_1  &  3$\times$3    & 1$\times$1                             & 128$\times \sfrac{H}{2} \times \sfrac{W}{2}$   & -                 \\
F1\_2  &  3$\times$3    & 1$\times$1                             & 128$\times \sfrac{H}{2} \times \sfrac{W}{2}$   & to U2             \\ \midrule
D2     &  3$\times$3    & 2$\times$2                             & 256$\times \sfrac{H}{4} \times \sfrac{W}{4}$   & -                 \\
F2\_1  &  3$\times$3    & 1$\times$1                             & 256$\times \sfrac{H}{4} \times \sfrac{W}{4}$   & -                 \\
F2\_2  &  3$\times$3    & 1$\times$1                             & 256$\times \sfrac{H}{4} \times \sfrac{W}{4}$   & -                 \\
F2\_3  &  3$\times$3    & 1$\times$1                             & 256$\times \sfrac{H}{4} \times \sfrac{W}{4}$   & to U1             \\ \midrule
D3     &  3$\times$3    & 2$\times$2                             & 512$\times \sfrac{H}{8} \times \sfrac{W}{8}$   & -                 \\
F3\_1  &  3$\times$3    & 1$\times$1                             & 512$\times \sfrac{H}{8} \times \sfrac{W}{8}$   & -                 \\
F3\_2  &  3$\times$3    & 1$\times$1                             & 512$\times \sfrac{H}{8} \times \sfrac{W}{8}$   & -                 \\
F3\_3  &  3$\times$3    & 1$\times$1                             & 512$\times \sfrac{H}{8} \times \sfrac{W}{8}$   & -                 \\ \midrule
U1     &  4$\times$4    & $\sfrac{1}{2}$$\times$$\sfrac{1}{2}$   & 256$\times \sfrac{H}{4} \times \sfrac{W}{4}$   & from F2\_3        \\ 
F4\_1  &  3$\times$3    & 1$\times$1                             & 256$\times \sfrac{H}{4} \times \sfrac{W}{4}$   & -                 \\
F4\_2  &  3$\times$3    & 1$\times$1                             & 256$\times \sfrac{H}{4} \times \sfrac{W}{4}$   & -                 \\
F4\_3  &  3$\times$3    & 1$\times$1                             & 256$\times \sfrac{H}{4} \times \sfrac{W}{4}$   & -                 \\ \midrule
U2     &  4$\times$4    & $\sfrac{1}{2}$$\times$$\sfrac{1}{2}$   & 128$\times \sfrac{H}{2} \times \sfrac{W}{2}$   & from F1\_2        \\ 
F5\_1  &  3$\times$3    & 1$\times$1                             & 128$\times \sfrac{H}{2} \times \sfrac{W}{2}$   & -                 \\
F5\_2  &  3$\times$3    & 1$\times$1                             & 64$\times \sfrac{H}{2} \times \sfrac{W}{2}$    & -                 \\ \midrule
U3     &  4$\times$4    & $\sfrac{1}{2}$$\times$$\sfrac{1}{2}$   & 64$\times H \times W$                          & from F0           \\ 
F6\_1  &  3$\times$3    & 1$\times$1                             & 15$\times H \times W$                          & -                 \\
F6\_2  &  3$\times$3    & 1$\times$1                             & 3$\times H \times W$                           & from input$^{\ast}$    \\ \midrule
\end{tabular}
\caption{Specifications of the \textsc{dbn} model. Each convolutional layer is followed by batch normalization and ReLU, except those that are skip connected to deeper layers, where only batch normalization has been applied, before the sum is rectified through a ReLU layer~\cite{he2015deep}. For example, the input to F4\_1 is the rectified summation of U1 and F2\_3. Note that for the skip connection from input layer to F6\_2, only the central frame of the stack is selected. At the end of the network a Sigmoid layer is applied to normalize the intensities. We use the Torch implementation of \texttt{SpatialConvolution} and \texttt{SpatialFullConvolution} for down- and up-convolutional layers. }
\label{table:architecture}
\end{table}

\begin{figure*}[t]
	\centering
	\begin{tabular}{*{8}{c@{\hspace{.5px}}}}
	\includegraphics[height=2.45cm,clip,trim=150 75 640 75]{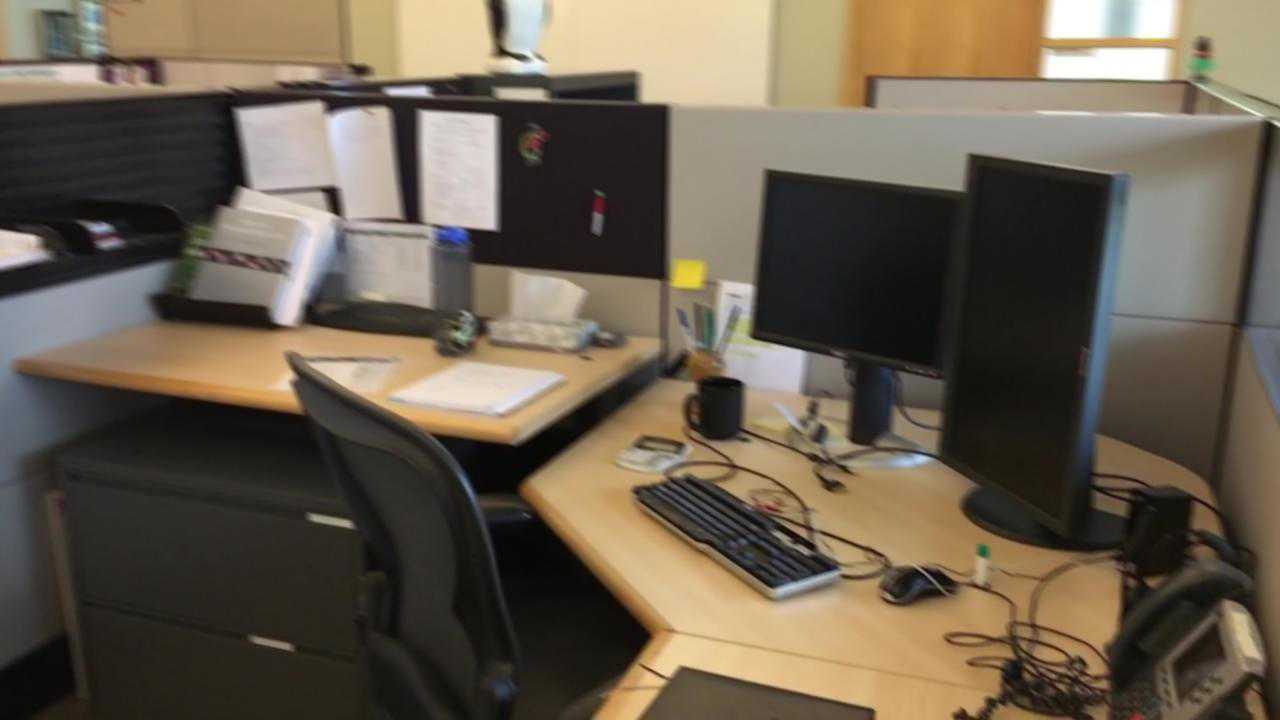} &
	\includegraphics[height=2.45cm,clip,trim=640 75 150 75]{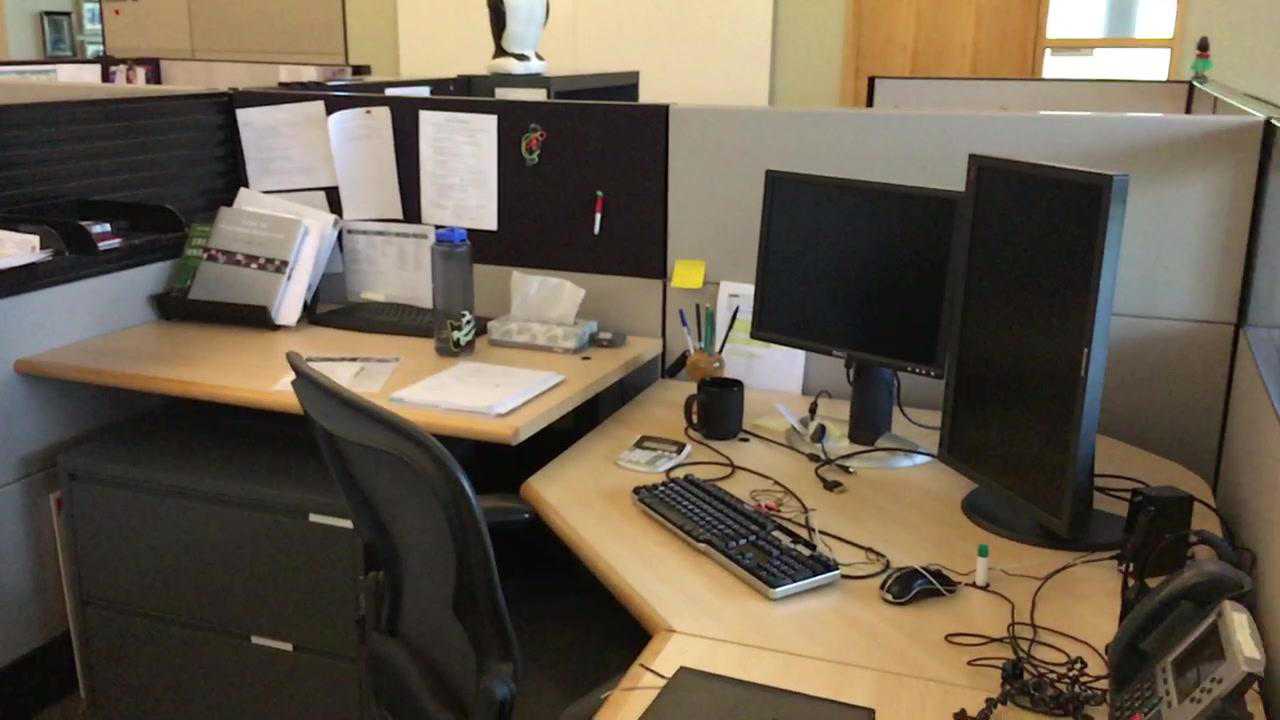} \hspace{3px} &
	\includegraphics[height=2.45cm,clip,trim=150 75 640 75]{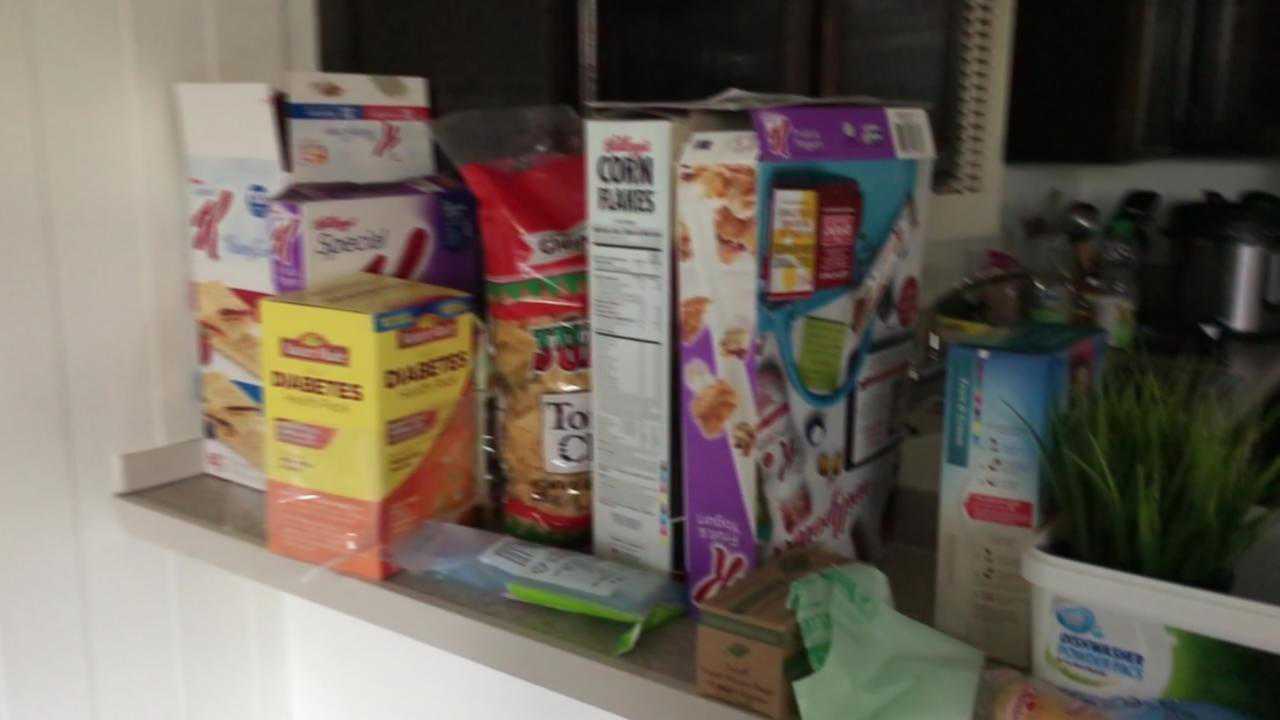} &
	\includegraphics[height=2.45cm,clip,trim=640 75 150 75]{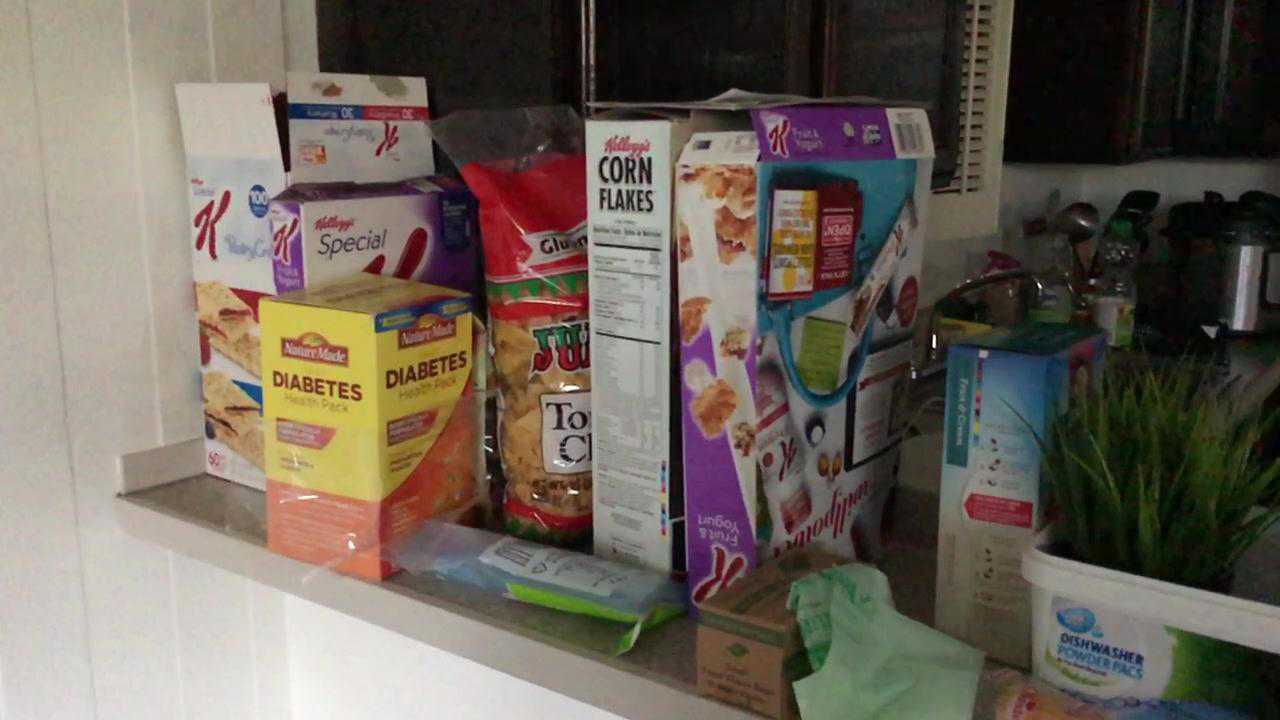} \hspace{3px} &
	\includegraphics[height=2.45cm,clip,trim=150 75 640 75]{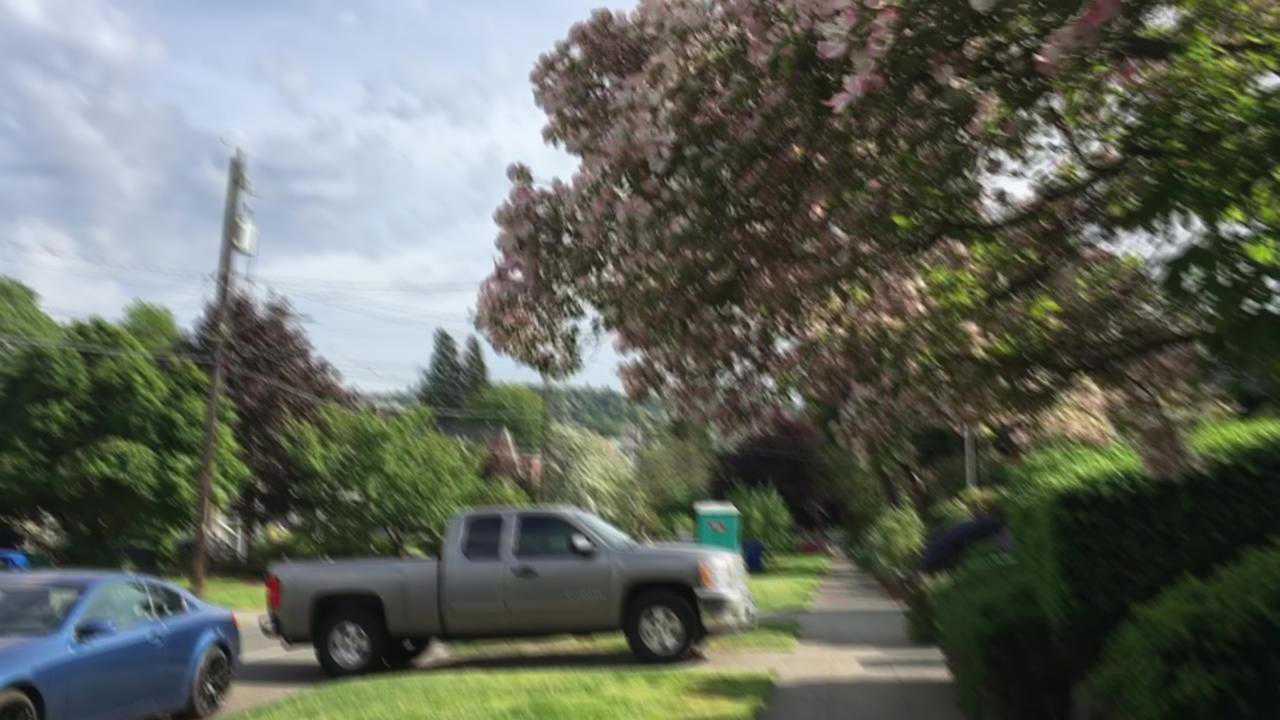} &
	\includegraphics[height=2.45cm,clip,trim=640 75 150 75]{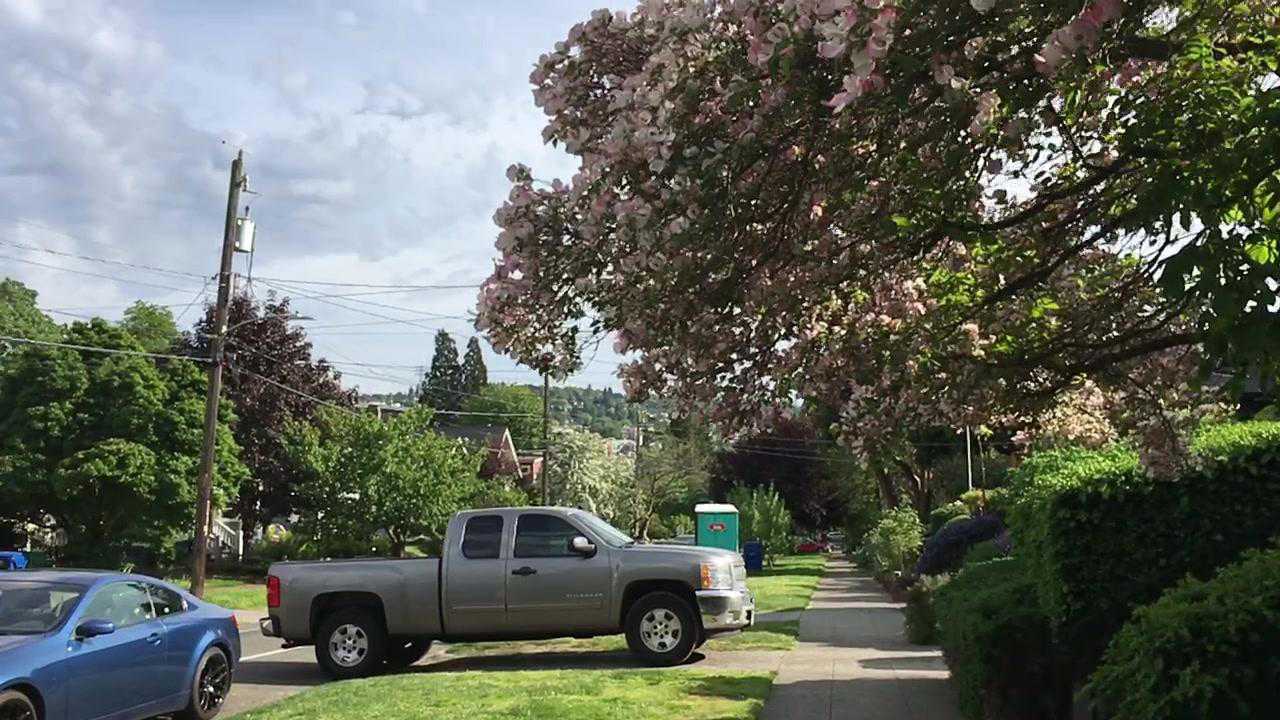} \hspace{3px} &
	\includegraphics[height=2.45cm,clip,trim=150 75 640 75]{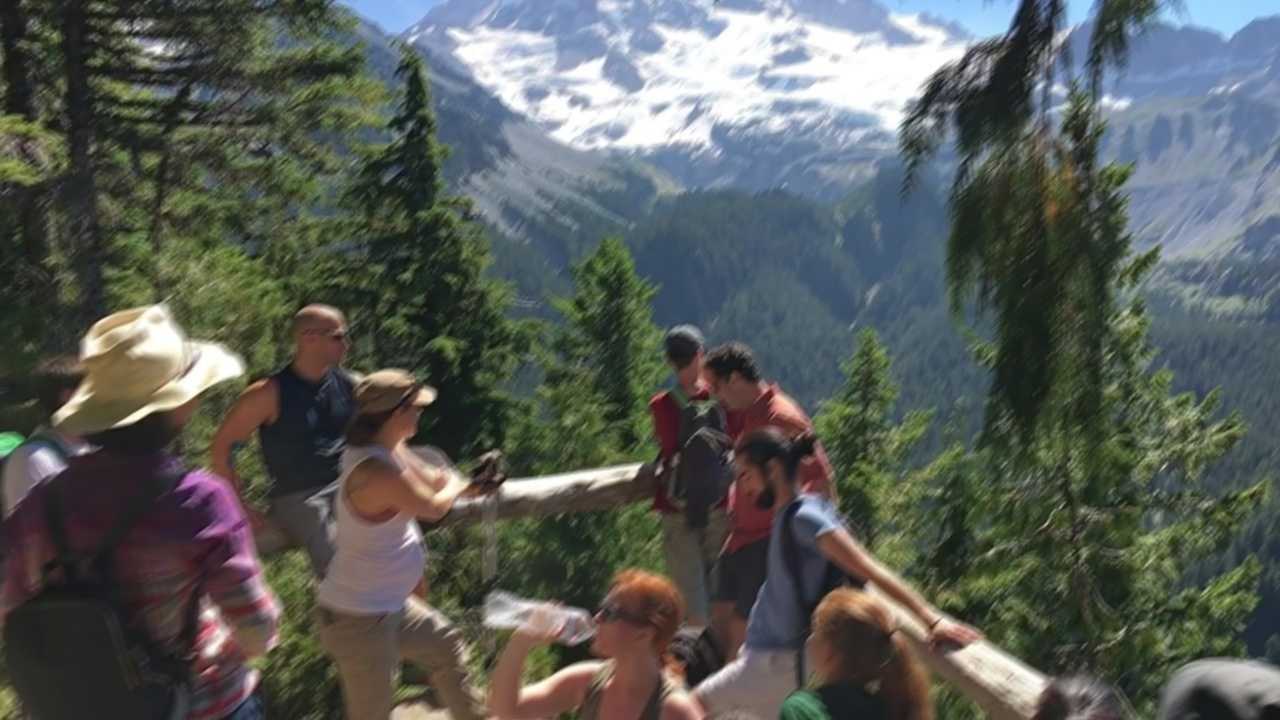} &
	\includegraphics[height=2.45cm,clip,trim=640 75 150 75]{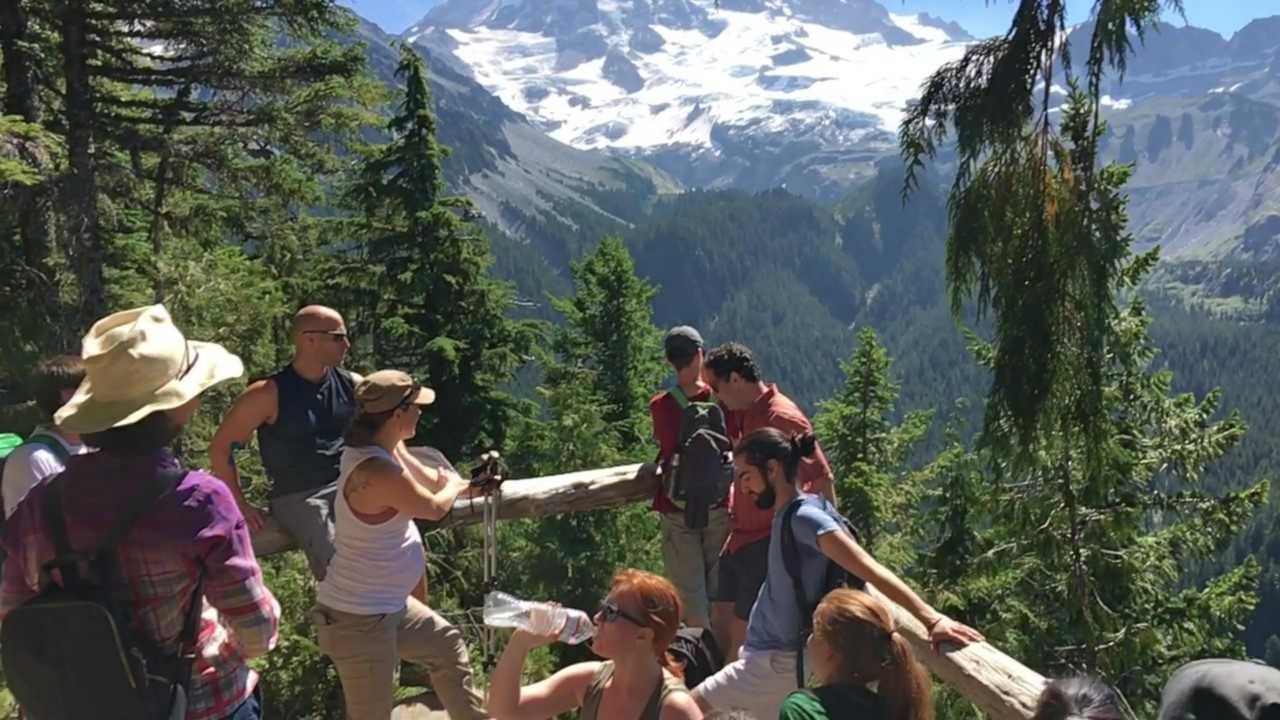} \\
	\end{tabular}
	\caption{A selection of blurry/sharp pairs (split left/right respectively) from our ground truth dataset. Images are best viewed on-screen and zoomed in.}
	\label{fig:deblurnet_dataset} 
\end{figure*}

We use an encoder-decoder style network, which have been shown to produce good results for a number of generative tasks~\cite{pathak2016context,simo2016learning}.
In particular, we choose a variation of the fully convolutional model proposed in~\cite{simo2016learning} for sketch cleanup. 
We add symmetric skip connections~\cite{mao2016image} between corresponding layers in encoder and decoder halves of the network, where features from the encoder side are added element-wise to each corresponding layer. 
This significantly accelerates the convergence and helps generate much sharper video frames. 
We perform an early fusion of neighboring frames that is similar to the FlowNetSimple model in~\cite{fischer2015flownet}, by concatenating all images in the input layer. 
The training loss is MSE to the ground truth sharp image, which will be discussed in more detail in Sec.~\ref{sec:dataset}. 
We refer to this network as \textsl{DeBlurNet}, or \textsc{dbn}, and show a diagram of it in Fig.~\ref{fig:architecture}. 
It consists of three types of convolutional layers: down-convolutional layers, that compress the spatial resolution of the features while increasing the spatial support of subsequent layers; flat-convolutional layers, that perform non-linear mapping and preserve the size of the image; and finally up-convolutional layers, that increase the spatial resolution.
Please refer to Tab.~\ref{table:architecture} for detailed configurations of the layers. 

\paragraph{Alignment.}
One of the main advantages of our method is the ability to work well without accurate frame-to-frame alignment.
To this end, we create three versions of our dataset with varying degrees of alignment, and use these to train \textsc{dbn}.
At one end, we use no alignment at all, relying on the network to abstract spatial information through a series of down-convolution layers. 
This makes the method significantly faster, as alignment usually dominates running time in multi-frame aggregation methods.
We refer to this network as \textsc{dbn+noalign}.
We also use optical flow~\cite{ipol.2013.26} to align stacks (\textsc{dbn+flow}), which is slow to compute and prone to errors (often introducing additional warping artifacts), but allows pixels to be aggregated more easily by removing the spatial variance of corresponding features.
Finally, we use a single global homography to align frames, which provides a compromise in approaches, in terms of computational complexity and alignment quality (\textsc{dbn+homog}). The homography is estimated using SURF features and a variant of RANSAC~\cite{torr2000mlesac} to reject outliers.

\paragraph{Implementation details.}
During training we use a batch size of 64, and patches of 15$\times$128$\times$128, where 15 is the total number of RGB channels stacked from the crops of 5 consecutive video frames.
We observed that a patch size of 128 was sufficient to provide enough overlapping content in the stack even if the frames are not aligned.
We use ADAM~\cite{kingma2014adam} for optimization, and fix the learning rate to be $0.005$ in the first 24,000 iterations, then halves for every subsequent 8,000 iterations until it reaches the lower bound of $10^{-6}$. 
For all the results reported in the paper we train the network for 80,000 iterations, which takes about 45 hours on a NVidia Titan X GPU. 
Default values of $\beta_1$, $\beta_2$ and $\epsilon$ are used, which are 0.9, 0.999, and $10^{-8}$ respectively, and we set weight decay to 0. 

As our network is fully convolutional, the input resolution is restricted only by GPU memory. 
At test time, we pass a $960\times540$ frame into the network, and tile this if the video frame is of larger resolution. 
Since our approach deblurs images in a single forward pass, it is computationally very efficient. 
Using a NVidia Titan X GPU, we can process a 720p frame within 1s without alignment. 
Previous approaches took on average 15s~\cite{delbracio2015hand} and 30s~\cite{cho2012video} per frame on CPUs. 
The recent neural deblurring method~\cite{chakrabarti2016neural} takes more than 1 hour to fully process each frame, and the approach of Kim et al~\cite{kim2015cvpr} takes several minutes per frame.

\section{Training Dataset}
\label{sec:dataset}
\begin{table*}[t]
\footnotesize
\rowcolors{2}{white}{rowblue}
\setlength{\tabcolsep}{4pt} % general space between cols (6pt standard)
\setlength\extrarowheight{2pt}
\centering
\resizebox{.99\linewidth}{!}{
\begin{tabular}{@{}lccccccccccc@{}}
\toprule
      Method                   & \#1             & \#2             & \#3             & \#4             & \#5             & \#6             & \#7             & \#8             & \#9             & \#10          & Average    \\ 
      \midrule
Input                          &    24.14 / .859    &    30.52 / .958    &    28.38 / .914    &    27.31 / .900    &    22.60 / .852    &    29.31  / .951   &    27.74  / .939   &    23.86  / .906   &    30.59  / .976   &    26.98 / .926    &    27.14  / .918 \\
\textsc{PSdeblur}                       &    24.42 / .908    &    28.77 / .952    &    25.15 / .928    &    27.77 / .928    &    22.02 / .890    &    25.74  / .932   &    26.11  / .948   &    19.75  / .822   &    26.48  / .963   &    24.62  / .938    &    25.08  / .921 \\
\textsc{wfa}~\cite{delbracio2015hand}   &    25.89 / .910    &    32.33 / .974    &    28.97 / .931    &    28.36 / .925    &    23.99 / .910    &    31.09 / .975    &    28.58 / .955    &    24.78 / .926    &    31.30 / .981    &    28.20 / .960    &    28.35 / .944  \\
\textsc{dbn+single}  &    25.75 / .901   &    31.15 / .966    &    29.30 / .946    &    28.38 / .922   &    23.63 / .885   &    30.70 / .962    &    29.23 / .959    &    25.62 / .936   &    31.92 / .983   &    28.06 / .949   &    28.37 / .941  \\
\textsc{dbn+noalign}  &    27.83 / .940    &    \textbf{33.11} / .980   &    \textbf{31.29} / \textbf{.973}   &    29.73 / .948  &    25.12 / .930    &    \textbf{32.52} / .978   &    \textbf{30.80} / \textbf{.975}   &    \textbf{27.28} / .962   &    \textbf{33.32} / \textbf{.989}   &    29.51 / .969   &    \textbf{30.05} / .964 \\
\textsc{dbn+homog.}                     &    27.93 / .945    &    32.39 / .975   &    30.97 / .969    &    29.82/ .948  &    24.79 / .925   &    31.84 / .972    &    30.46 / .972   &    26.64 / .955   &    33.15 / \textbf{.989}    &    29.30 / .969    &    29.73 / .962  \\
\textsc{dbn+flow} &    \textbf{28.31}  / \textbf{.956}  &    33.14 / \textbf{.982}  &    30.92 / \textbf{.973}   &    \textbf{29.99} / \textbf{.954}    &    \textbf{25.58} / \textbf{.944}   &    32.39 / \textbf{.981}   &    30.56 / \textbf{.975}    &    27.15 / \textbf{.963}    &    32.95 / \textbf{.989}    &    \textbf{29.53} / \textbf{.975}    &    \textbf{30.05} / \textbf{.969}  \\ 
\bottomrule
\end{tabular}
}
\caption{\label{tab:quantitative} PSNR/MSSIM~\cite{kohler2012recording} measurements for each approach, averaged over all frames, for 10 test datasets (\#1$\rightarrow$\#10).}

\end{table*}

Generating realistic training data is a major challenge for tasks where ground truth data cannot be easily collected/labeled. For training our neural network, we require two video sequences of exactly the same content: one blurred by camera shake motion blur, and its corresponding sharp version. 
Capturing such data is extremely hard. One could imagine using a beam-splitter and multiple cameras to build a special capturing system, but this setup would be challenging to construct robustly, and would present a host of other calibration issues.

One solution would be to use rendering techniques to create synthetic videos for training. However if not done properly, this often leads to a domain gap, where models trained on synthetic data do not generalize well to real world data. For example, we could apply synthetic motion blur on sharp video frames to simulate camera shake blur. 
However, in real world scenarios the blur not only depends on camera motion, but also is related to scene depth and object motion, thus is very difficult to be rendered properly.

In this work, we propose to collect real-world sharp videos at very high framerate, and synthetically create blurred ones by accumulating a number of short exposures to approximate a longer exposure~\cite{telleen2007synthetic}. 
In order to simulate realistic motion blur at 30fps, we capture videos at 240fps, and subsample every eighth frame to create the 30fps ground
truth sharp video.  We then average together a temporally centered window of 7 frames (3 on either side of the ground truth frame) to generate synthetic motion blur at the target frame rate.

Since there exists a time period between adjacent exposures (the ``duty cycle''), simply averaging consecutive frames will yield ghosting artifacts. 
To avoid this, \cite{kim2016dynamic} proposed to only use frames whose relative motions in-between are smaller than 1 pixel. To use all frames for rendering, we compute optical flow between adjacent high fps frames, and generate an additional 10 evenly spaced inter-frame images, which we then average together. 
Examples of the dataset are shown in Fig.~\ref{fig:deblurnet_dataset}.  
We plan to release this dataset publicly for future research.

In total, we collect 71 videos, each with 3-5s average running time. 
These are used to generate 6708 synthetic blurry frames with corresponding ground truth.
We subsequently augment the data by flipping, rotating ($0\degree$, $90\degree$, $180\degree$, $270\degree$), and scaling ($\sfrac{1}{4},\sfrac{1}{3},\sfrac{1}{2}$) the images, and from this we draw on average 10 random 128$\times$128 crops. 
In total, this gives us 2,146,560 pairs of patches. 
We split our dataset into 61 training videos and 10 testing videos. 
For each video, its frames are used for either training or testing, but not both, meaning that the scenes used for testing have not been seen in the training data. 

The training videos are capture at 240fps with an iPhone 6s, GoPro Hero 4 Black, and Canon 7D. The reason to use multiple devices is to avoid bias towards a specific capturing device that may generate videos with some unique characteristics.  
We test on videos captured by other devices, including Nexus 5x and Moto X mobile phones and a Sony a6300 consumer camera.

\paragraph{Limitations.}
We made an significant effort to capture a wide range of situations, including long pans, selfie videos, scenes with moving content  (people, water, trees), recorded with a number of different capture devices. 
While it is quite diverse, it also has some limitations. 
As our blurry frames are averaged from multiple input frames, the noise characteristics will be different in the
ground truth image. 
To reduce this effect, we recorded input videos in high light situations, where there was minimal visible noise even in the original 240fps video, meaning that our dataset only contains scenes with sufficient light. 
An additional source of error is that using optical-flow for synthesizing motion blur adds possible artifacts which would not exist in real-world data. We found that however, as the input video is recorded at 240fps, the motion between frames is small, and we did not observe visual artifacts from this step. 

As we will show in Sec.~\ref{sec:experiments}, despite these limitations, our trained model still generalizes well to new capture devices and scene types, notably on low-light videos.
We believe future improvements to the training data set will further improve the performance of our method.

\begin{figure*}[t]
	\centering
		\resizebox{1.0\linewidth}{!}{
		\hspace{-20px}
	\centering
\scriptsize
\newcommand{\boyincludea}[1]{\includegraphics[height=.07\textheight,clip,trim=540 360 640 260]{#1}}
\newcommand{\boyincludeb}[1]{\includegraphics[height=.07\textheight,clip,trim=330 220 800 350]{#1}}
\newcommand{\yachtincludea}[1]{\includegraphics[height=.07\textheight,clip,trim=680 200 460 380]{#1}}
\newcommand{\yachtincludeb}[1]{\includegraphics[height=.07\textheight,clip,trim=380 320 740 240]{#1}}
\newcommand{\leninincludea}[1]{\includegraphics[height=.07\textheight,clip,trim=230 250 910 330]{#1}}
\newcommand{\leninincludeb}[1]{\includegraphics[height=.07\textheight,clip,trim=880 180 240 380]{#1}}

\newcommand{\peopleincludea}[1]{\includegraphics[height=.07\textheight,clip,trim=200 220 1520 660]{#1}}
\newcommand{\peopleincludeb}[1]{\includegraphics[height=.07\textheight,clip,trim=1200 220 520 660]{#1}}

\def\arraystretch{0.5}%
\begin{tabular}{*{9}{c@{\hspace{1.5px}}}}

\includegraphics[height=.07\textheight,clip,trim=70 0 70 0]{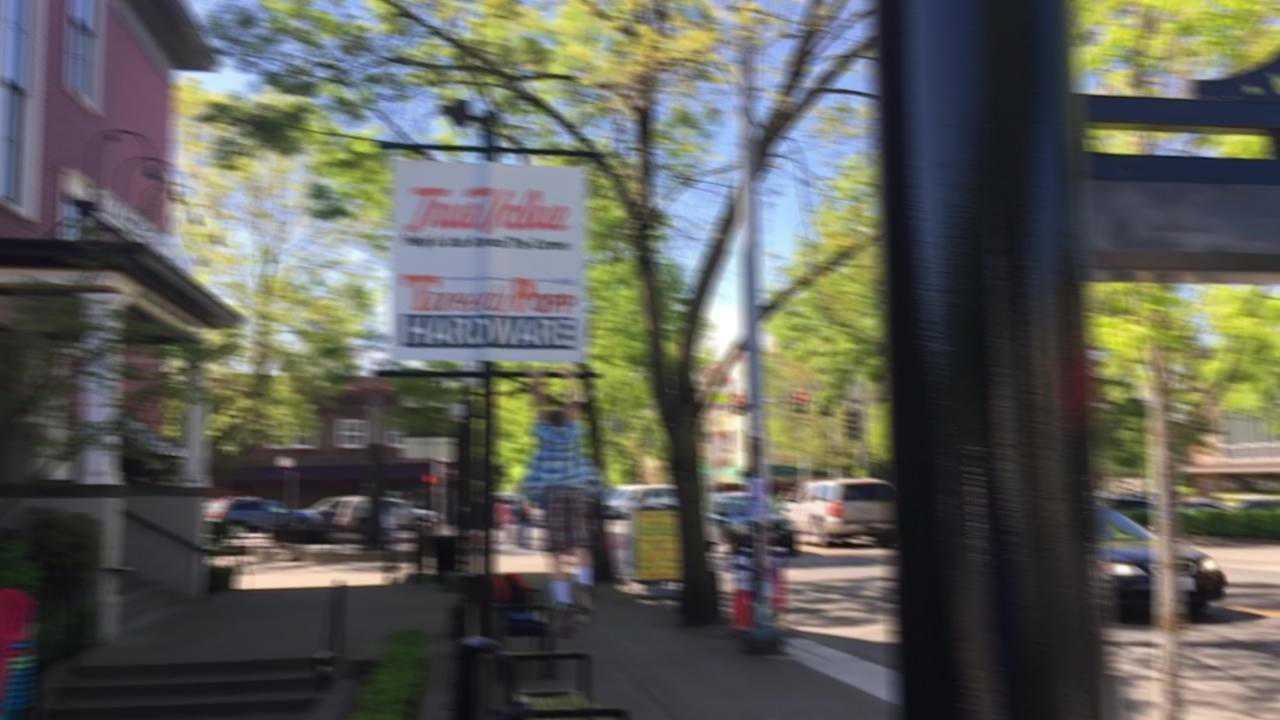} &
\boyincludea{figures/qualitative_comparisons/quantest_01_input} &
\boyincludea{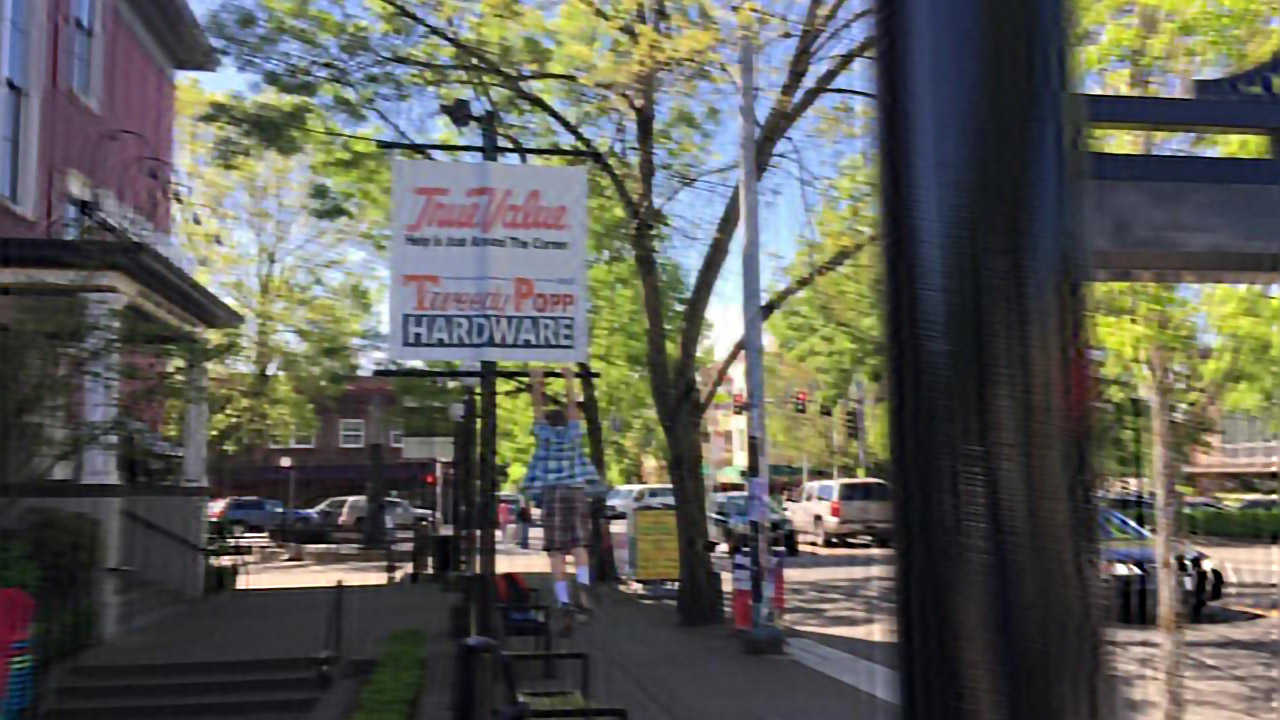} &
\boyincludea{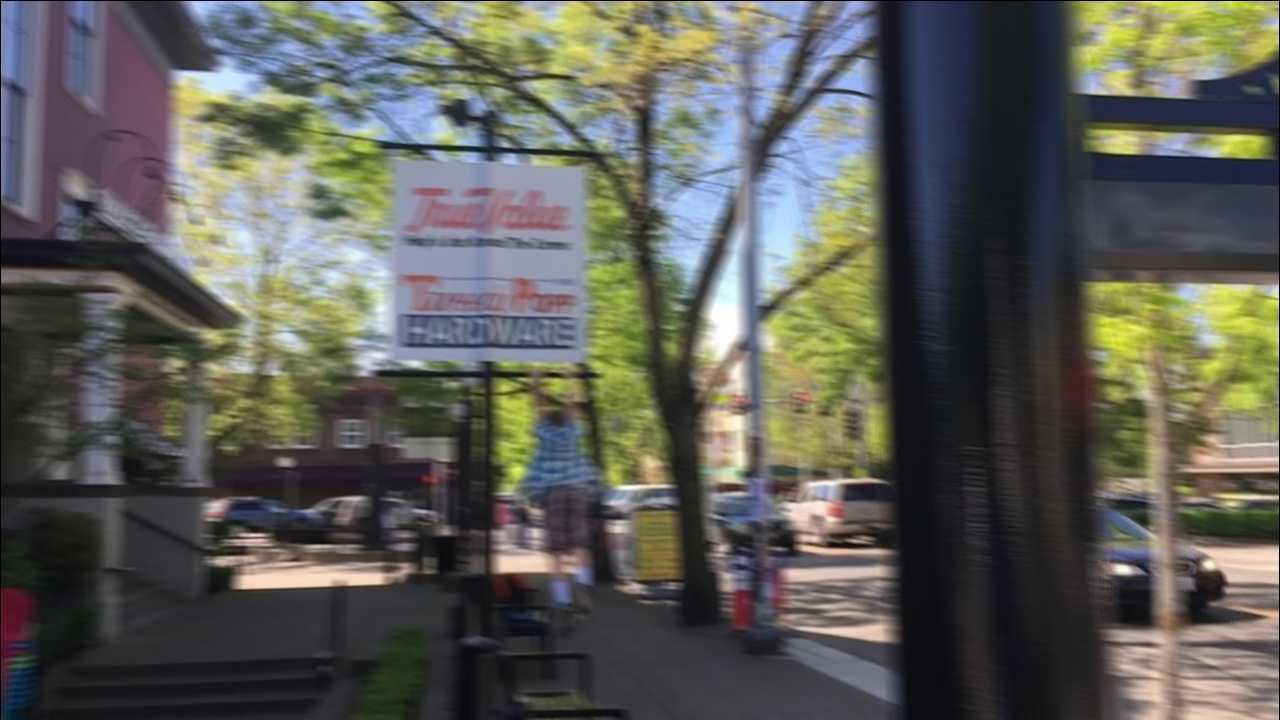} &
\boyincludea{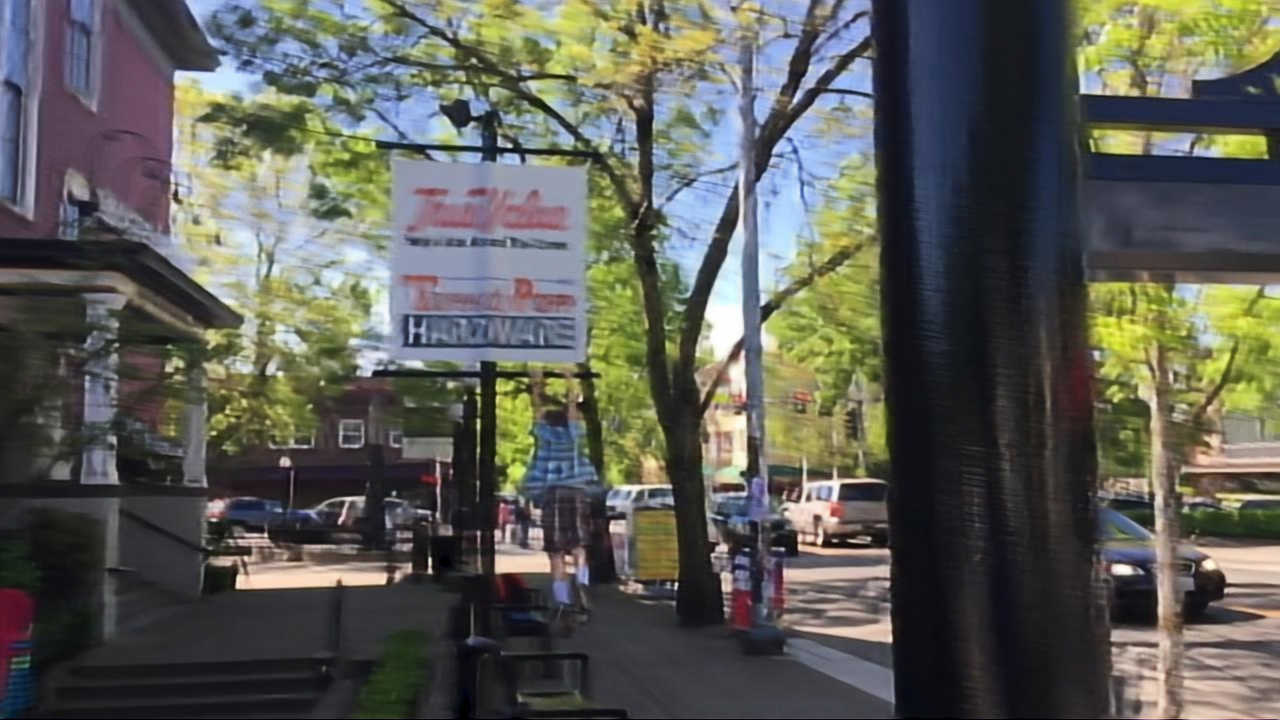} &
\boyincludea{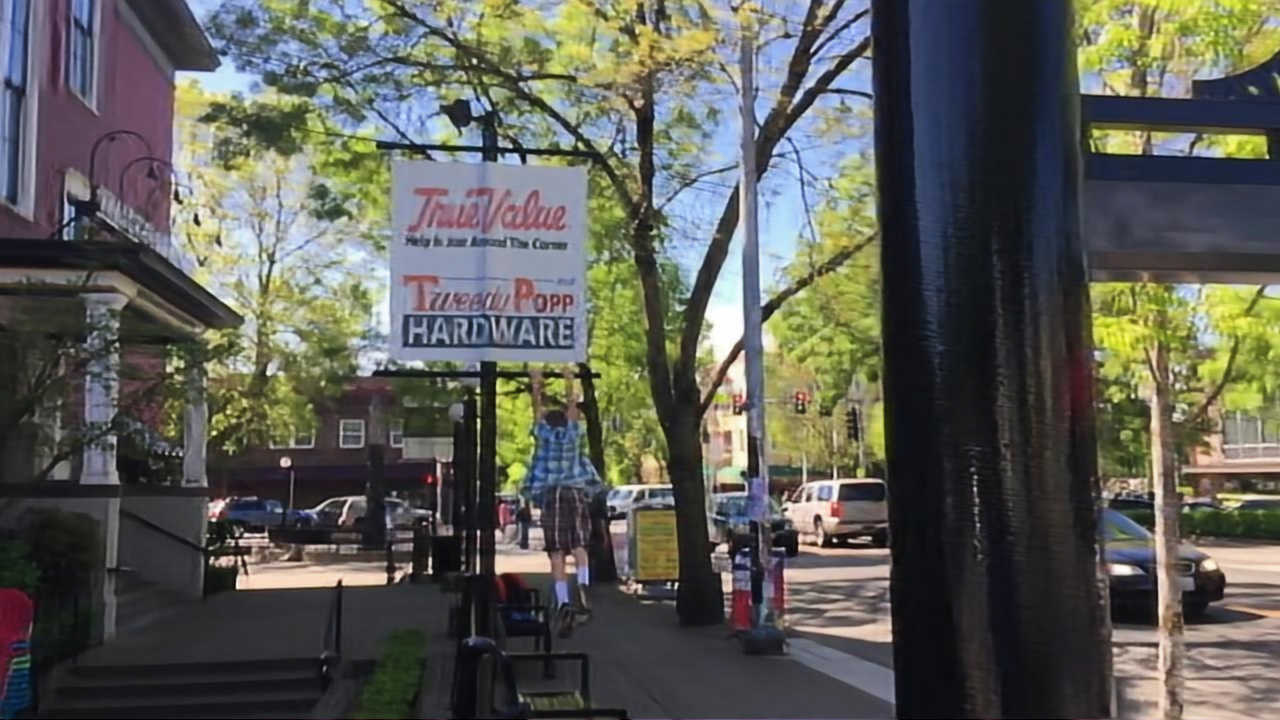} &
\boyincludea{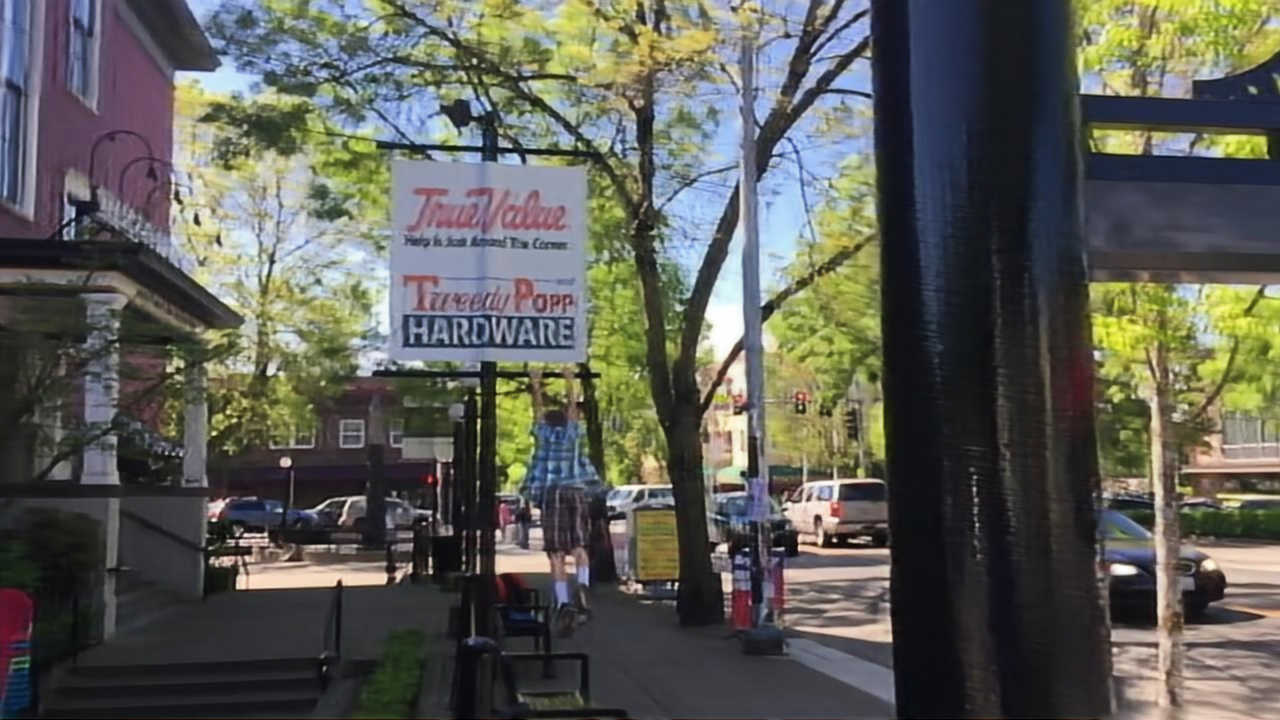} &
\boyincludea{figures/qualitative_comparisons/quantest_01_OF_baseline_deeper} &
\boyincludea{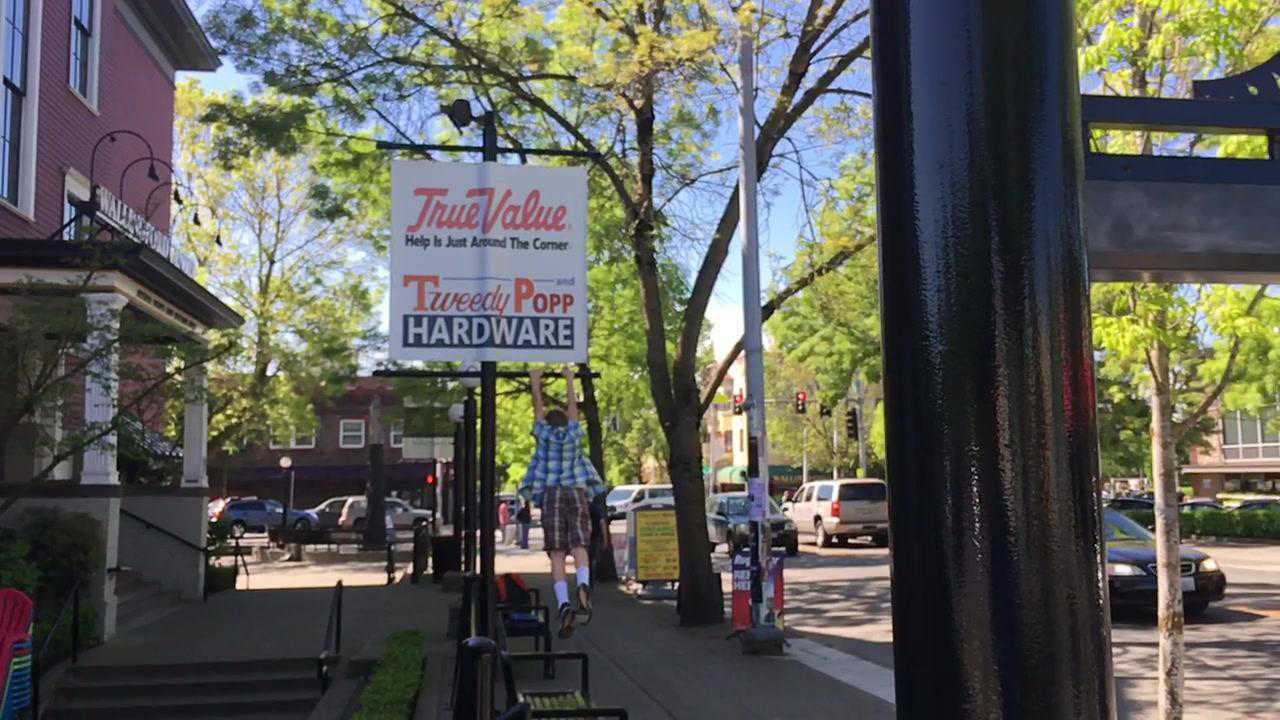} \\

\includegraphics[height=.07\textheight,clip,trim=70 0 70 0]{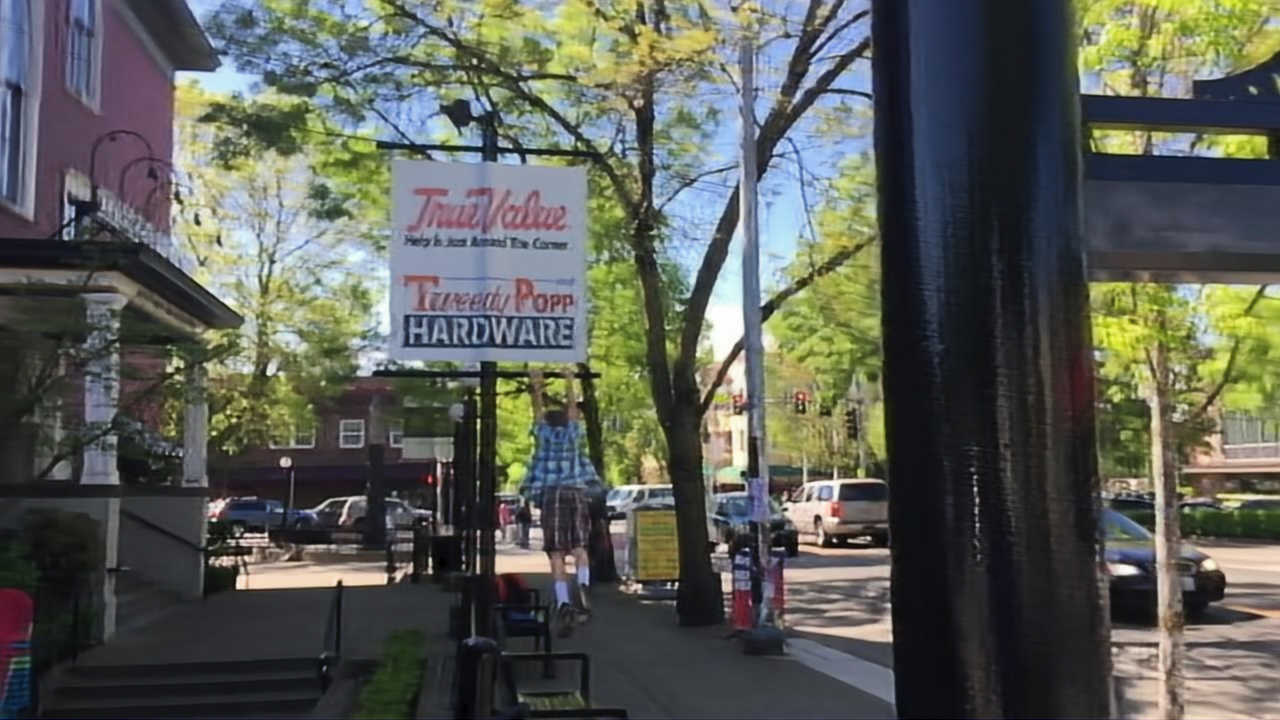} &
\boyincludeb{figures/qualitative_comparisons/quantest_01_input} &
\boyincludeb{figures/qualitative_comparisons/quantest_01_ps_deblur} &
\boyincludeb{figures/qualitative_comparisons/quantest_01_wfa_oliver} &
\boyincludeb{figures/qualitative_comparisons/quantest_01_single_baseline_deeper} &
\boyincludeb{figures/qualitative_comparisons/quantest_01_none_baseline_deeper} &
\boyincludeb{figures/qualitative_comparisons/quantest_01_homography_baseline_deeper} &
\boyincludeb{figures/qualitative_comparisons/quantest_01_OF_baseline_deeper} &
\boyincludeb{figures/qualitative_comparisons/quantest_01_gt} \\

\includegraphics[height=.07\textheight,clip,trim=70 0 70 0]{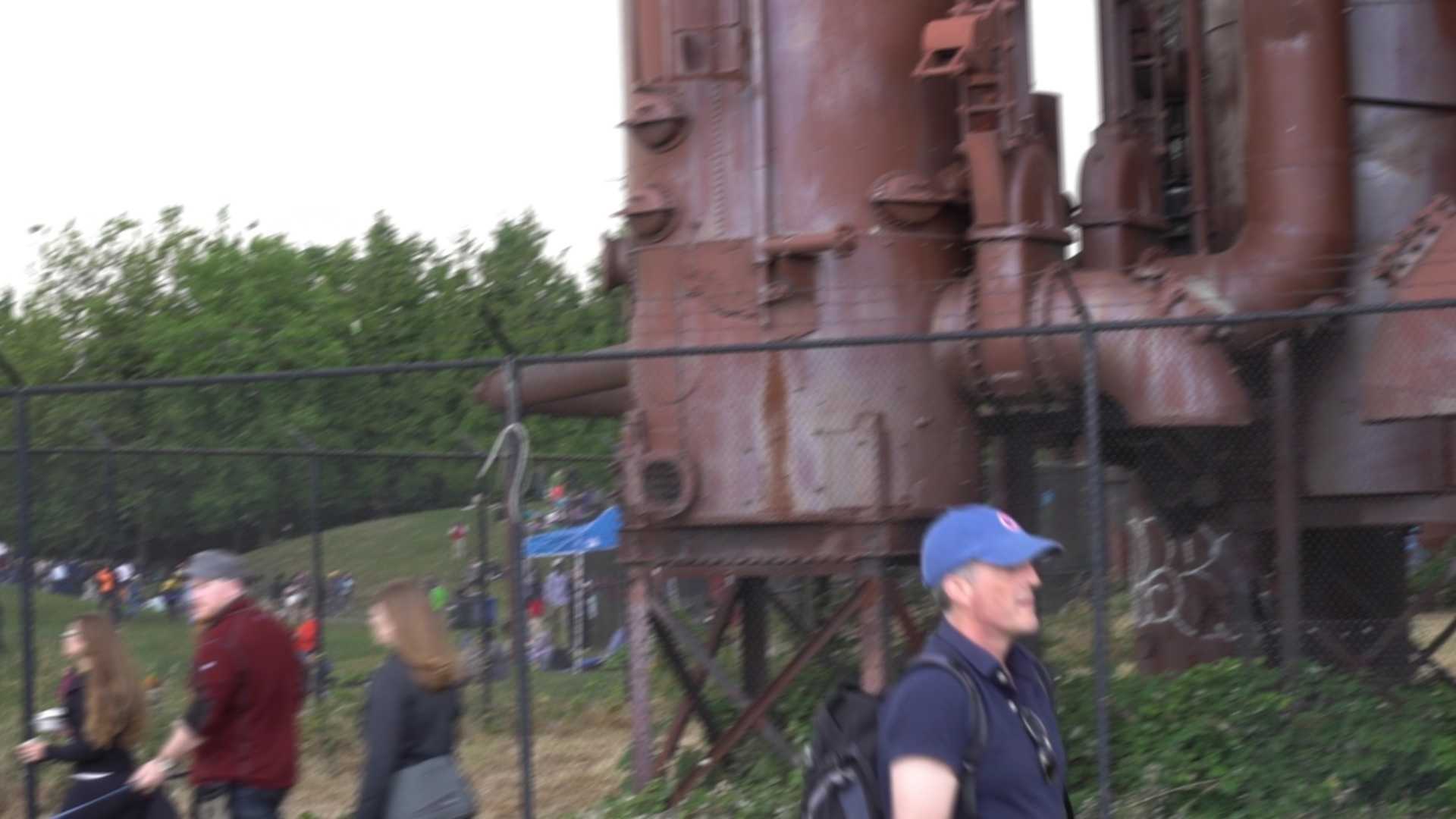} &
\peopleincludea{figures/qualitative_comparisons/quantest_04_input} &
\peopleincludea{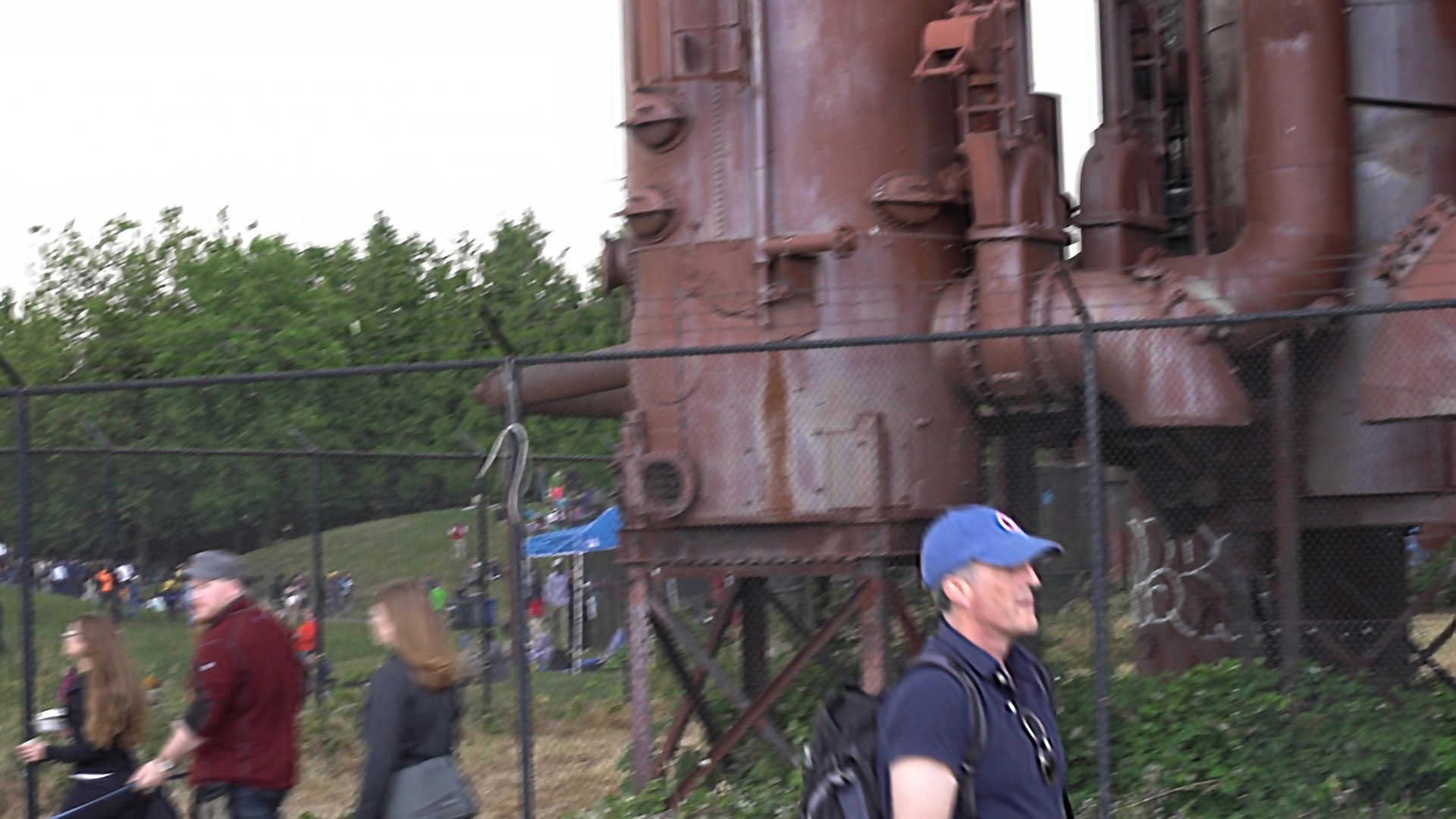} &
\peopleincludea{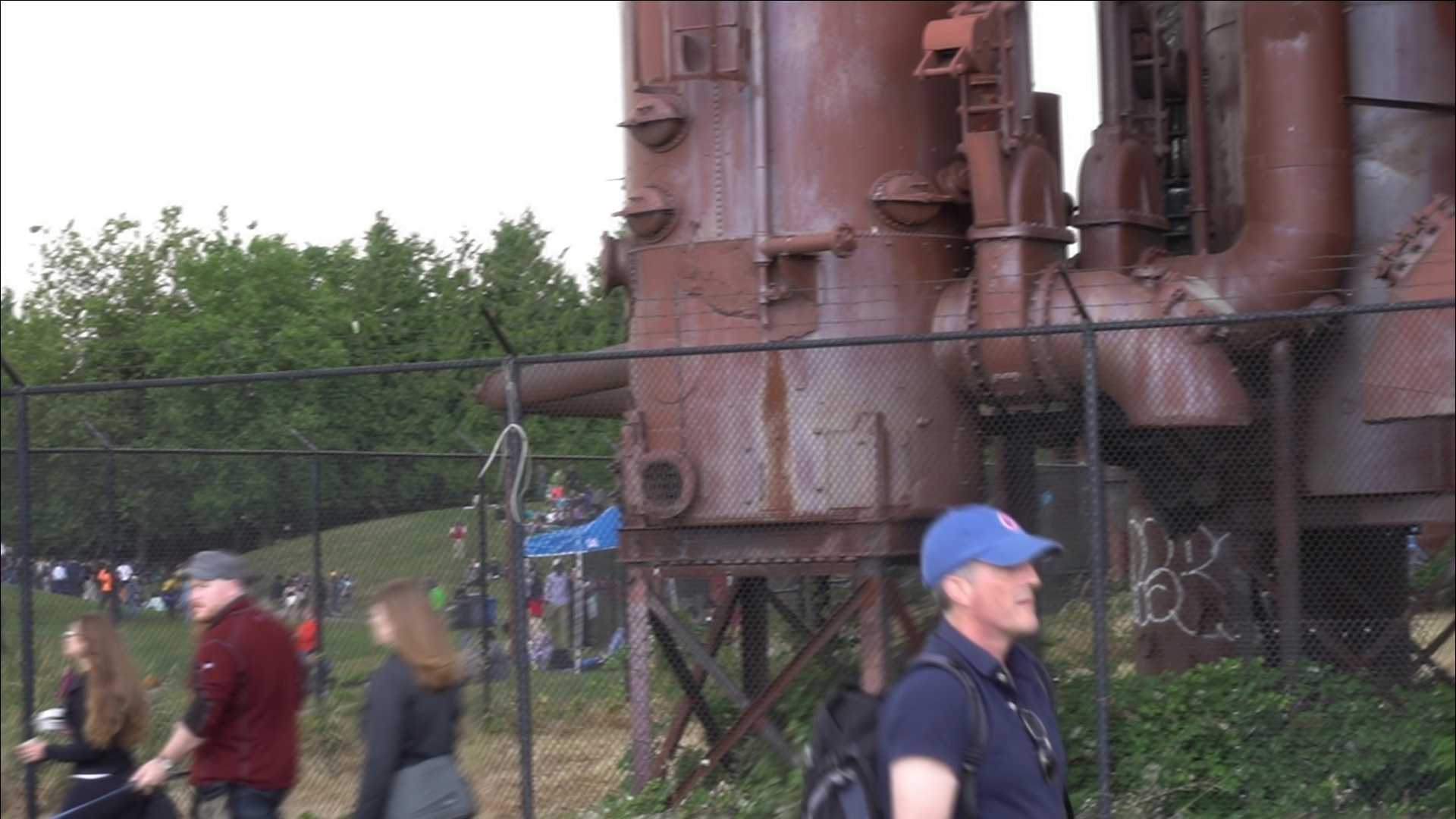} &
\peopleincludea{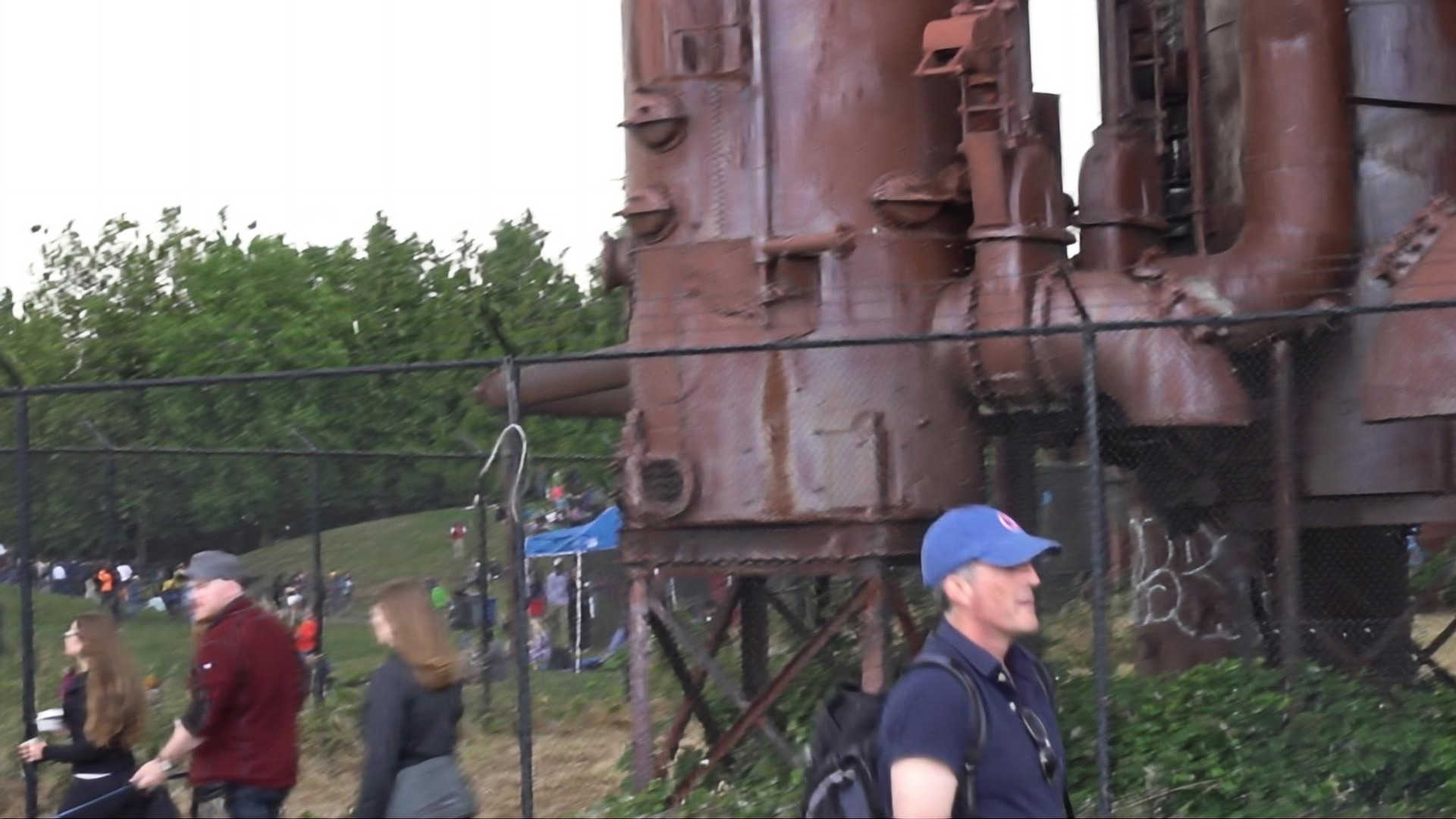} &
\peopleincludea{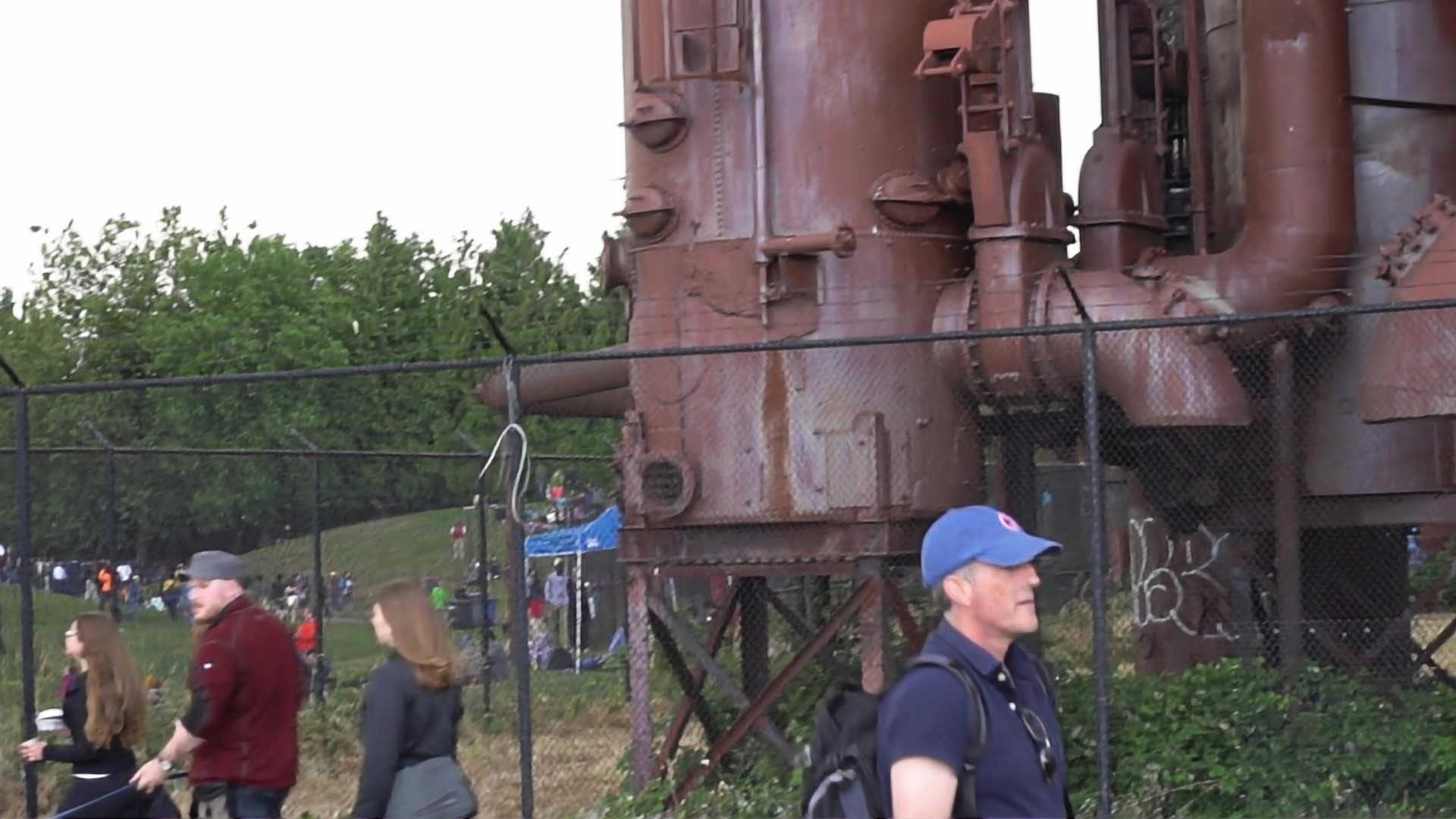} &
\peopleincludea{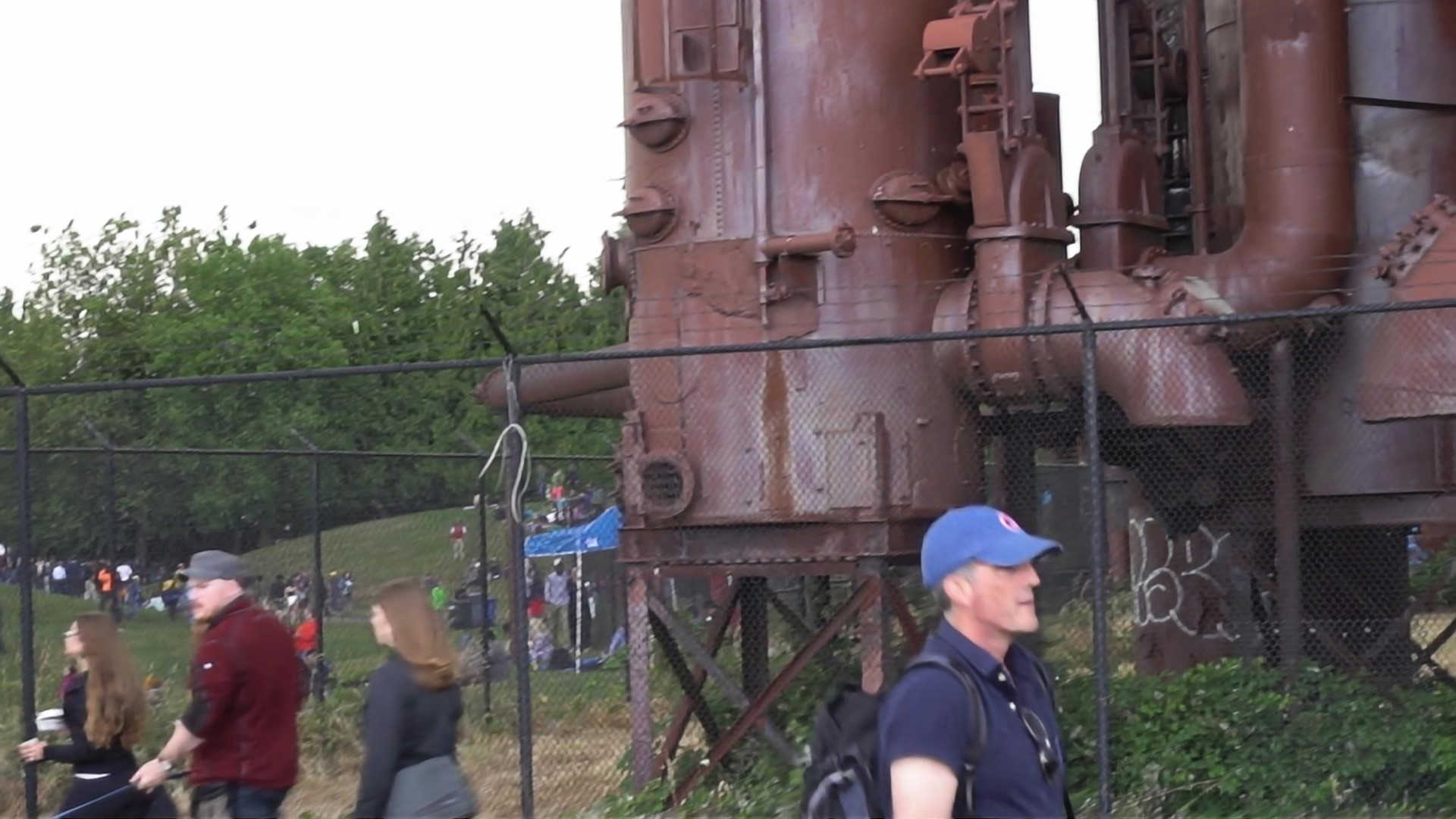} &
\peopleincludea{figures/qualitative_comparisons/quantest_04_OF_baseline_deeper} &
\peopleincludea{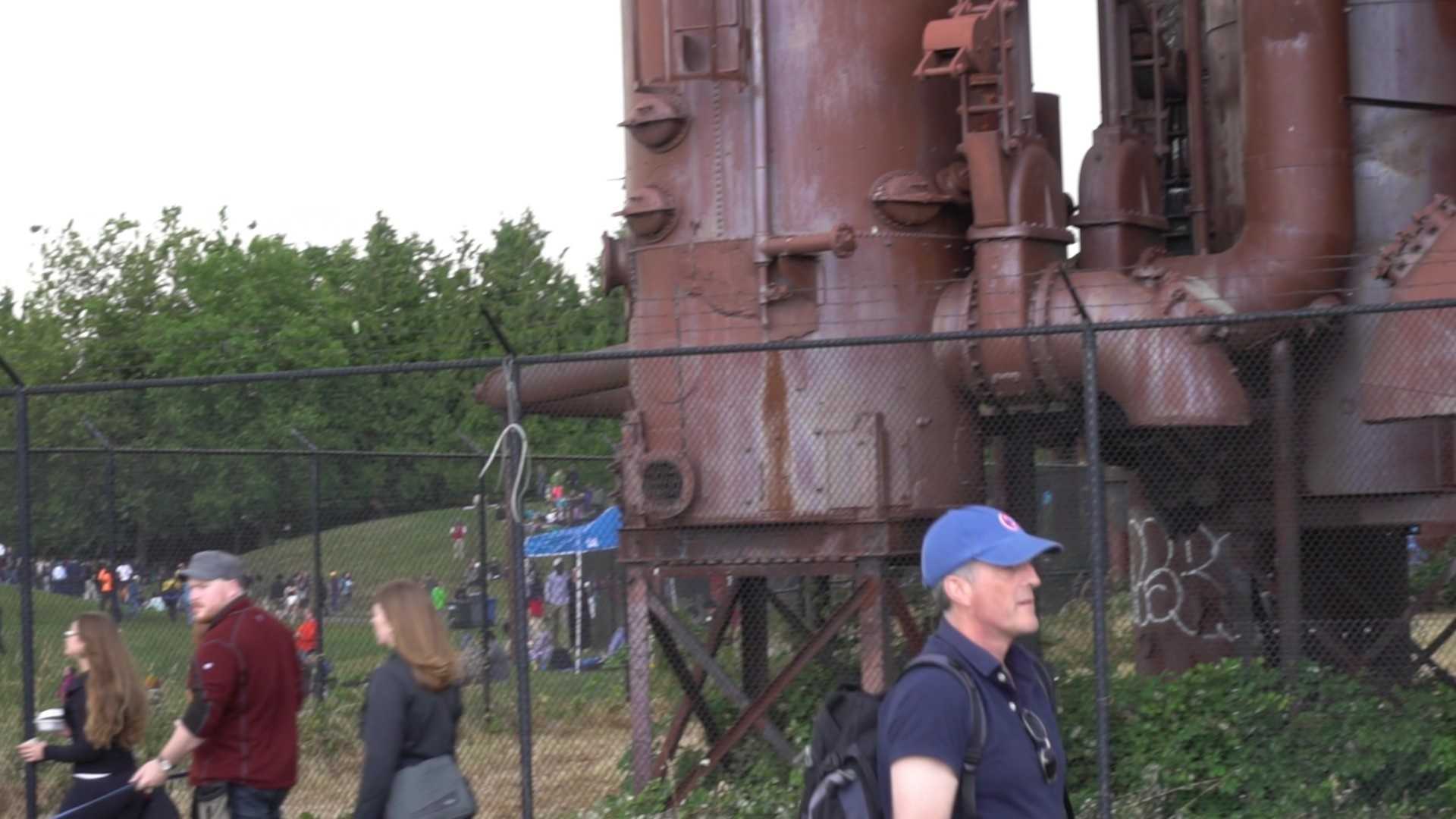} \\

\includegraphics[height=.07\textheight,clip,trim=70 0 70 0]{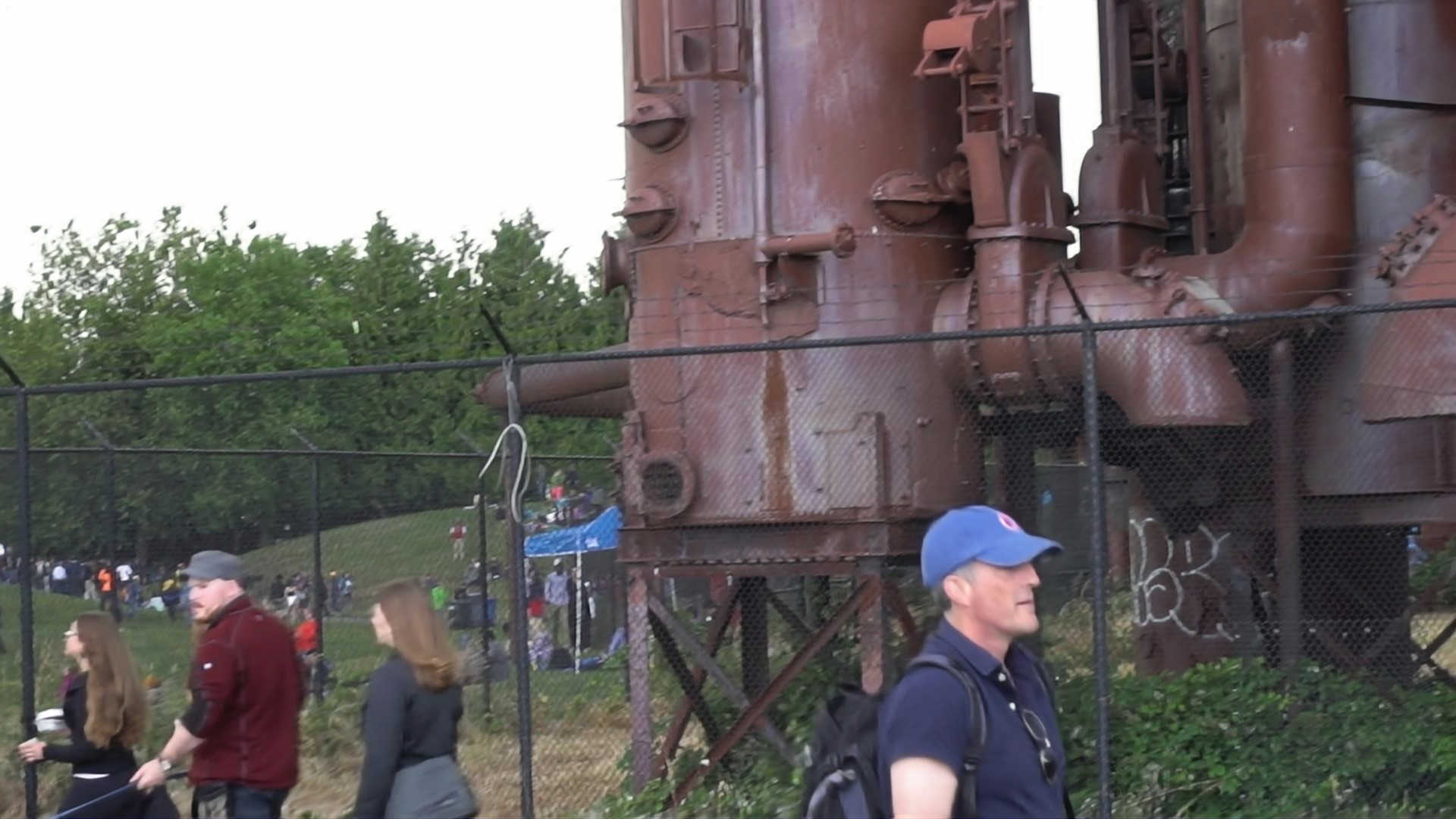} &
\peopleincludeb{figures/qualitative_comparisons/quantest_04_input} &
\peopleincludeb{figures/qualitative_comparisons/quantest_04_ps_deblur} &
\peopleincludeb{figures/qualitative_comparisons/quantest_04_wfa_oliver} &
\peopleincludeb{figures/qualitative_comparisons/quantest_04_single_baseline_deeper} &
\peopleincludeb{figures/qualitative_comparisons/quantest_04_none_baseline_deeper} &
\peopleincludeb{figures/qualitative_comparisons/quantest_04_homography_baseline_deeper} &
\peopleincludeb{figures/qualitative_comparisons/quantest_04_OF_baseline_deeper} &
\peopleincludeb{figures/qualitative_comparisons/quantest_04_gt} \\

Input (top) / ours (bottom) & Input & \textsc{PSdeblur} & \textsc{wfa}\cite{delbracio2015hand} & \textsc{dbn+single} & \textsc{dbn+noalign} & \textsc{dbn+homog} & \textsc{dbn+flow}  & ground-truth\\[.25em]
 & 21.79dB & 24.09dB & 21.53dB & 24.51dB & 27.24dB & 26.66dB & 26.69dB  & \\[.25em]
 & 31.72dB & 31.13dB & 29.83dB & 31.49dB & 32.89dB & 34.76dB & 34.87dB  & \\
\end{tabular}

	}
	\caption{Quantitative results from our test set, with PSNRs relative to the ground truth. Here we compare \textsc{dbn} with a \emph{single-image} approach, \textsc{PSdeblur}, and a start-of-the-art multi-frame video deblurring method, \textsc{wfa}~\cite{delbracio2015hand}. \textsc{dbn}, achieves comparable results to ~\cite{delbracio2015hand} without alignment, and improved results with alignment. }
	\label{fig:deblurnet_comparison_3} 
\end{figure*}

\section{Experiments and Results}
\label{sec:experiments}

We conduct a series of experiments to evaluate the effectiveness of the learned model, and also the importance of individual components. 

\begin{figure}[t]
	\centering	
	\includegraphics[width=0.99\linewidth]{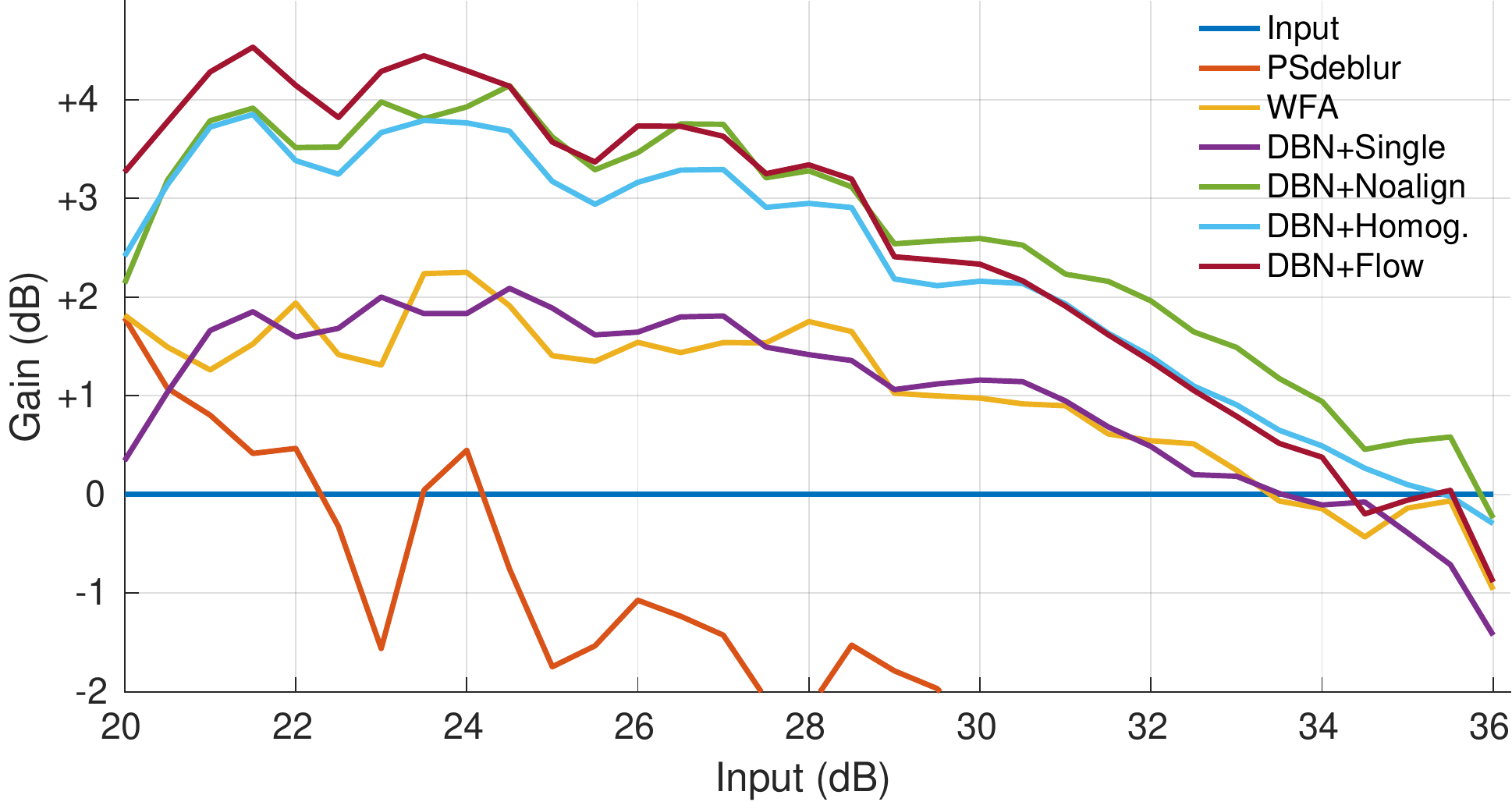}
	\caption{\label{fig:quantitative} Quantitative comparison of different approaches. In this plot, the PSNR gain of applying different methods and configurations is plotted versus the sharpness of the input frame. We observe that all multi-frame methods provide a quality improvement for blurry input frames, with diminishing improvements as the input frames get sharper. \textsc{dbn+noalign} and \textsc{dbn+flow} perform the best, but qualitatively, \textsc{dbn+flow} and \textsc{dbn+homog.} are often comparable, and superior to no alignment. We provide a \emph{single-image} uniform blur kernel deblurring method as reference (\textsc{PSdeblur}).}
\end{figure}

\begin{figure*}[t]
	\centering
	\resizebox{1.0\linewidth}{!}{
			\hspace{-20px}
	\centering
\scriptsize
\newcommand{\pianoincludea}[1]{\includegraphics[height=.07\textheight,clip,trim=560 780 1160 100]{#1}}
\newcommand{\pianoincludeb}[1]{\includegraphics[height=.07\textheight,clip,trim=880 550 840 330]{#1}}
\newcommand{\supincludea}[1]{\includegraphics[height=.07\textheight,clip,trim=900 500 820 380]{#1}}
\newcommand{\supincludeb}[1]{\includegraphics[height=.07\textheight,clip,trim=380 380 1400 560]{#1}}
\newcommand{\walkincludea}[1]{\includegraphics[height=.07\textheight,clip,trim=330 660 1420 250]{#1}}
\newcommand{\walkincludeb}[1]{\includegraphics[height=.07\textheight,clip,trim=1170 330 550 550]{#1}}

\def\arraystretch{0.5}%
\begin{tabular}{*{9}{c@{\hspace{1.5px}}}}
\scriptsize

\includegraphics[height=.07\textheight,clip,trim=70 0 70 0]{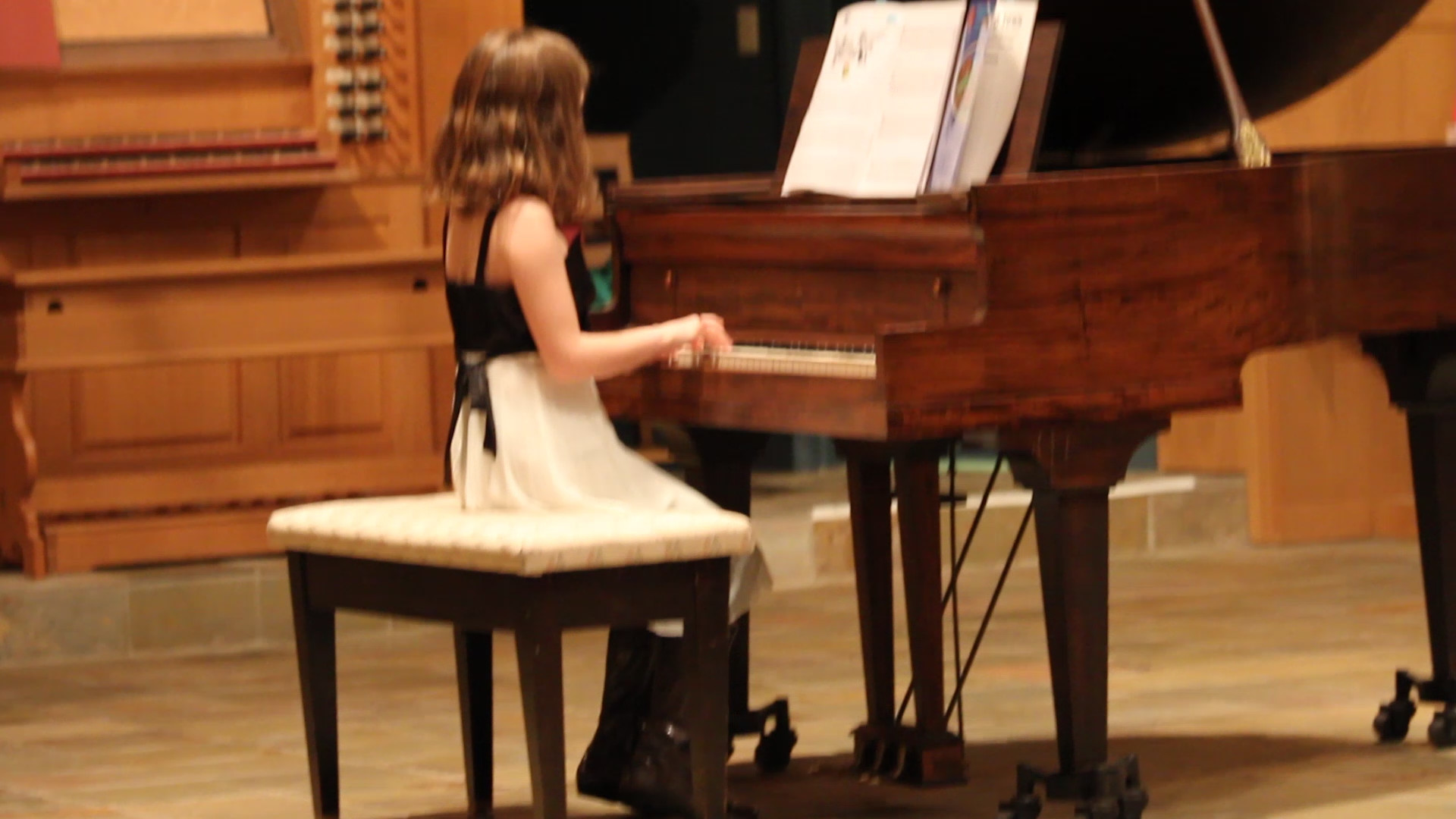} &
\pianoincludea{figures/qualitative_comparisons/08_input} &
\includegraphics[height=.07\textheight]{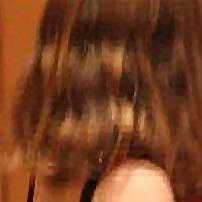} &
\pianoincludea{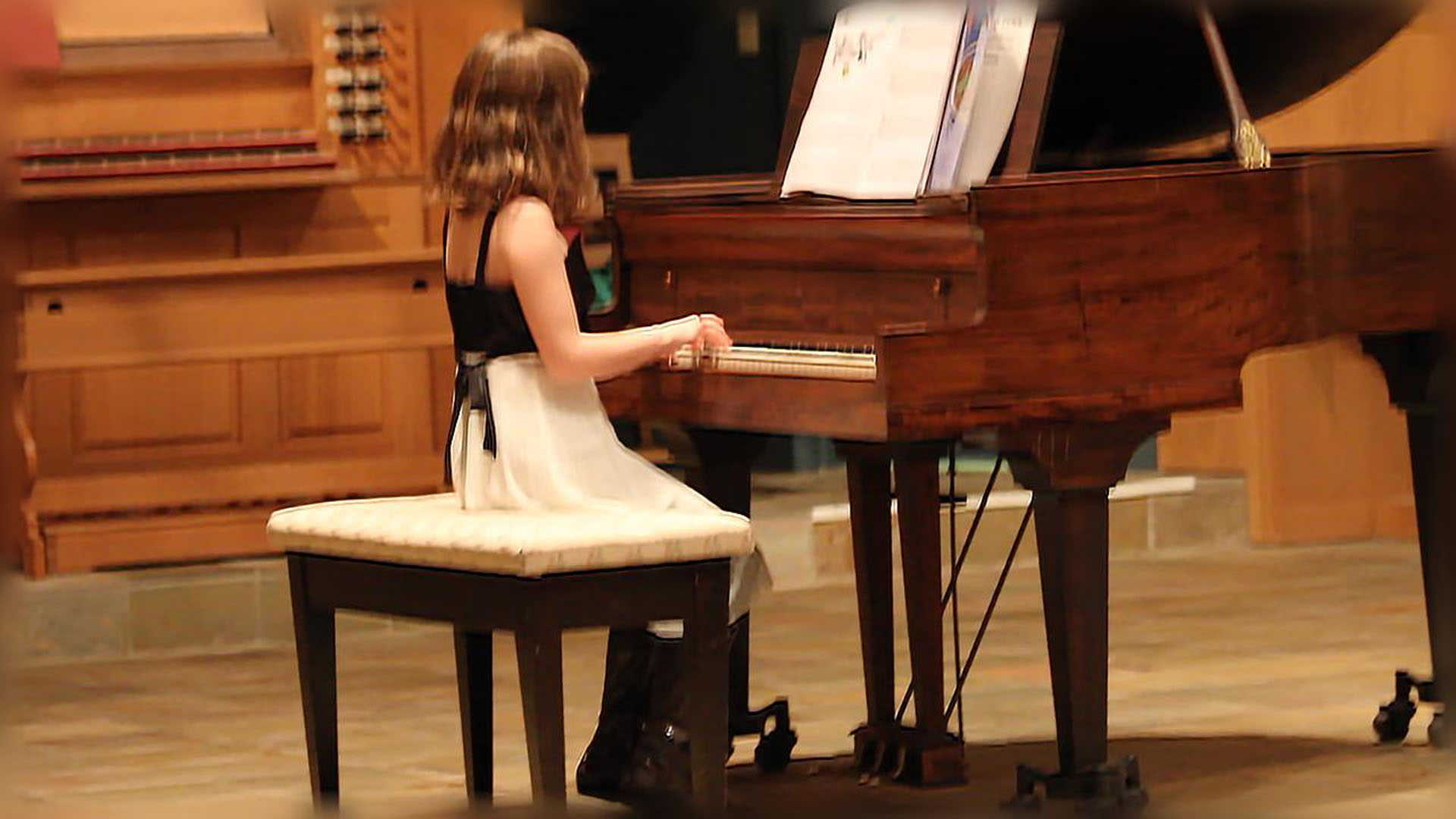} &
\pianoincludea{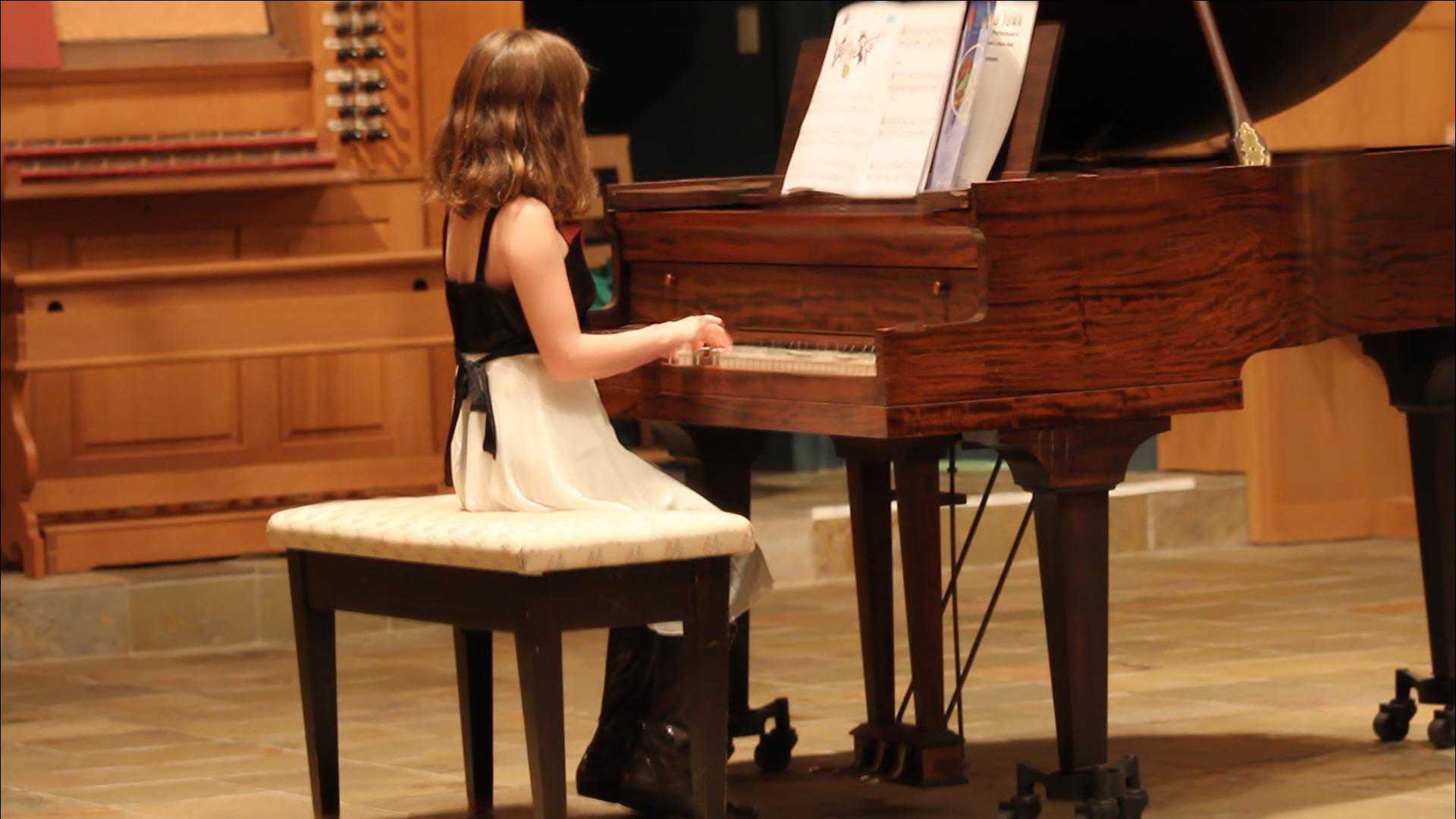} &
\pianoincludea{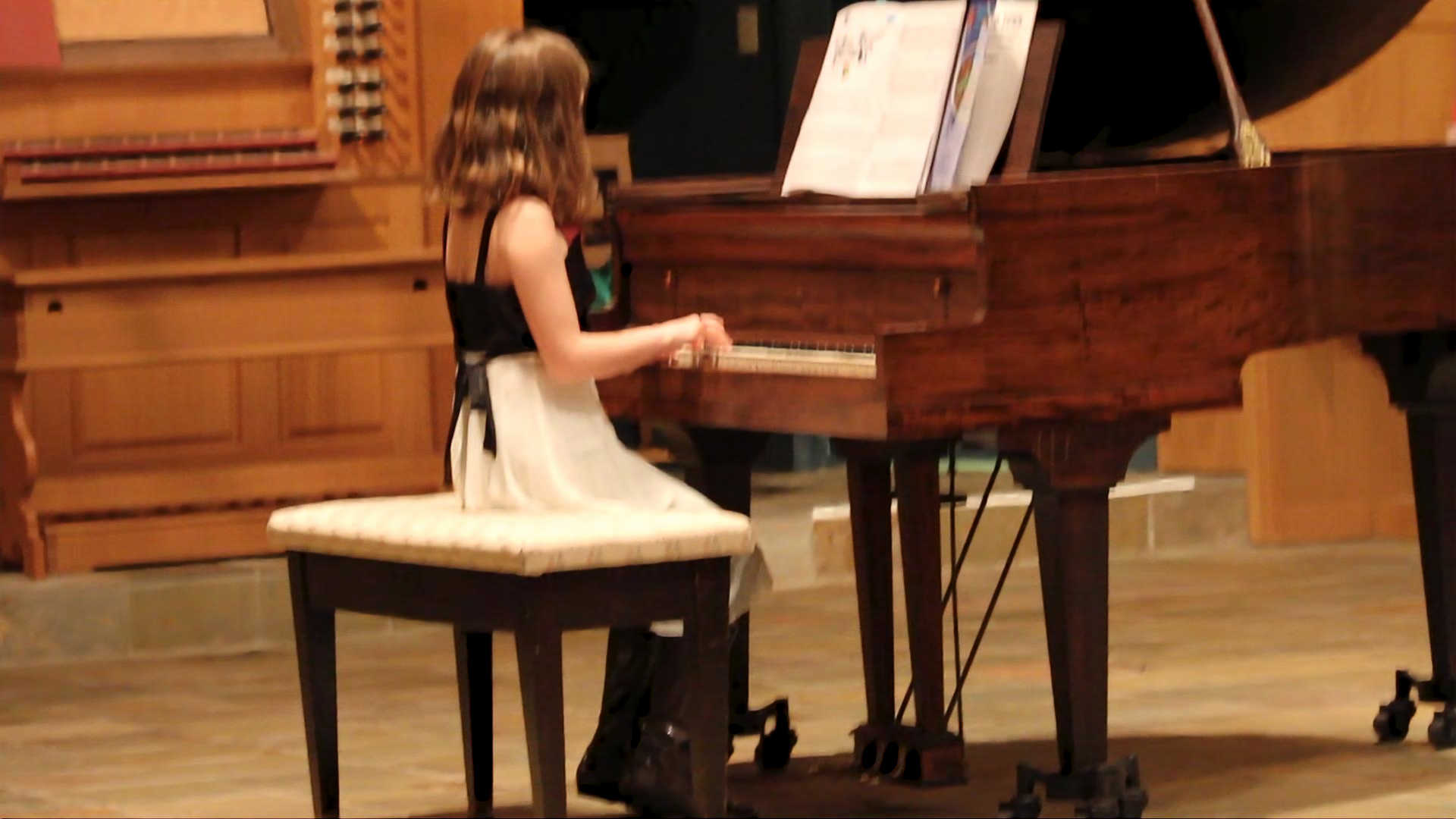} &
\pianoincludea{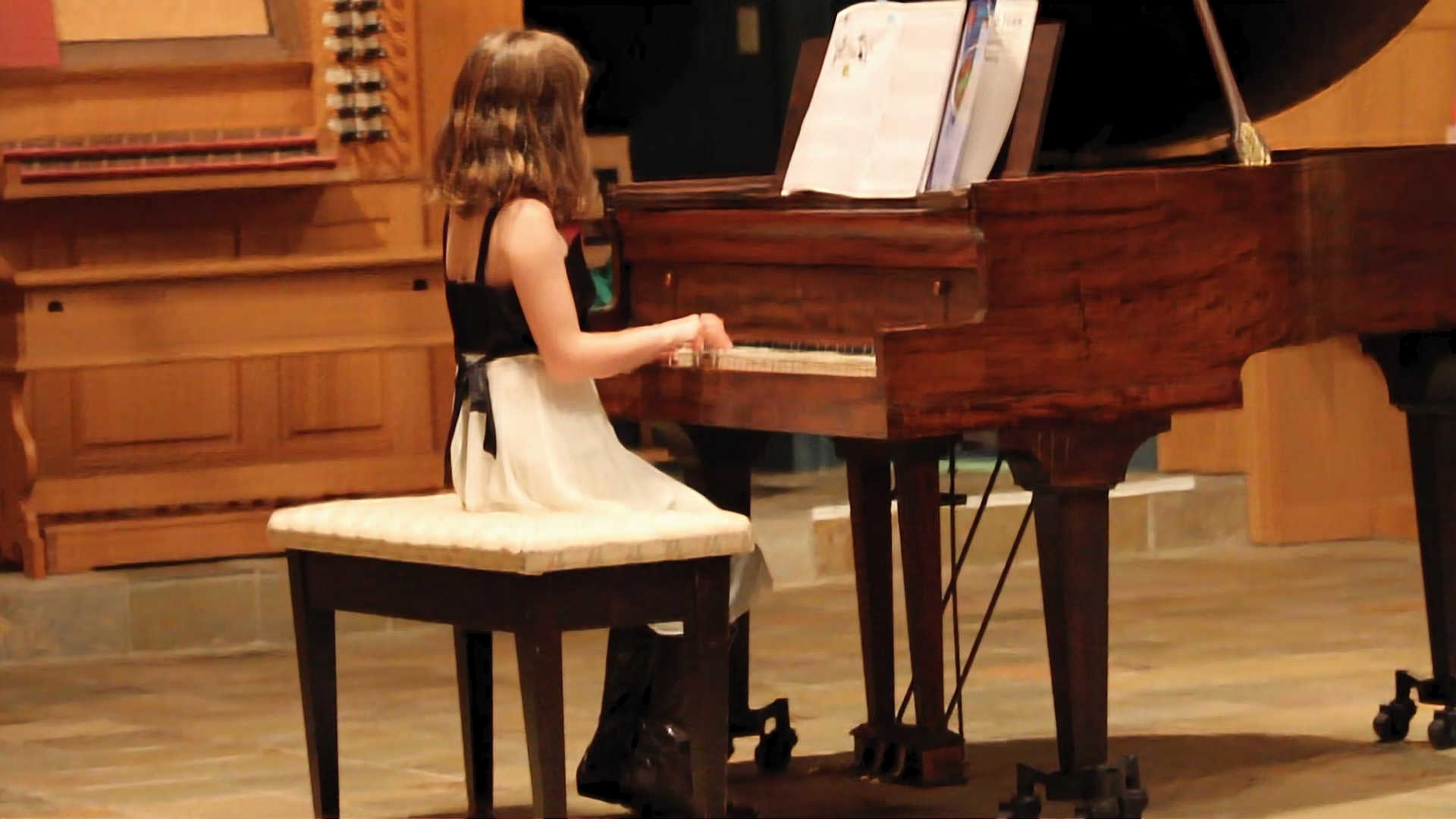} &
\pianoincludea{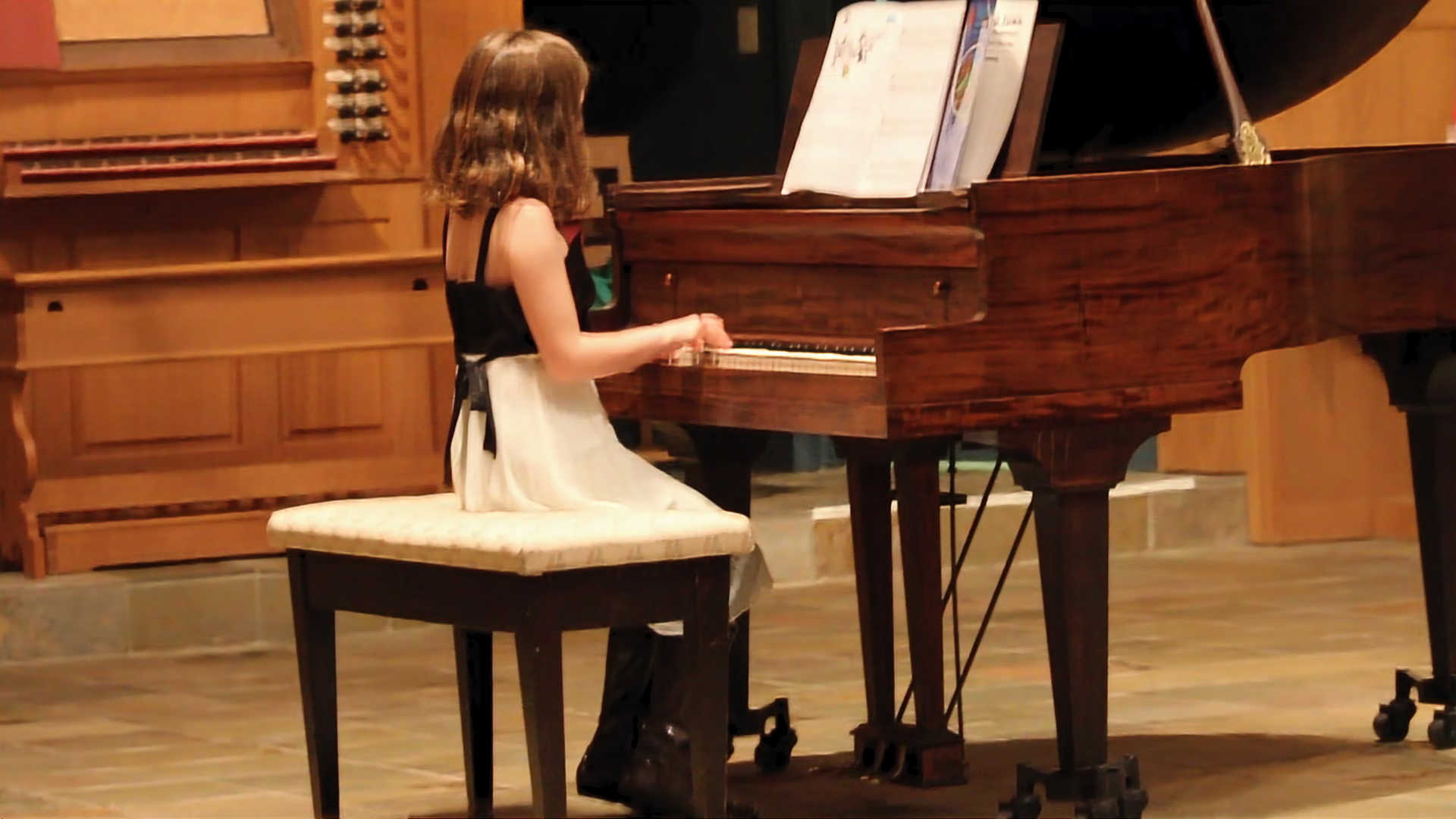} &
\pianoincludea{figures/qualitative_comparisons/08_OF_baseline_deeper2} \\

\includegraphics[height=.07\textheight,clip,trim=70 0 70 0]{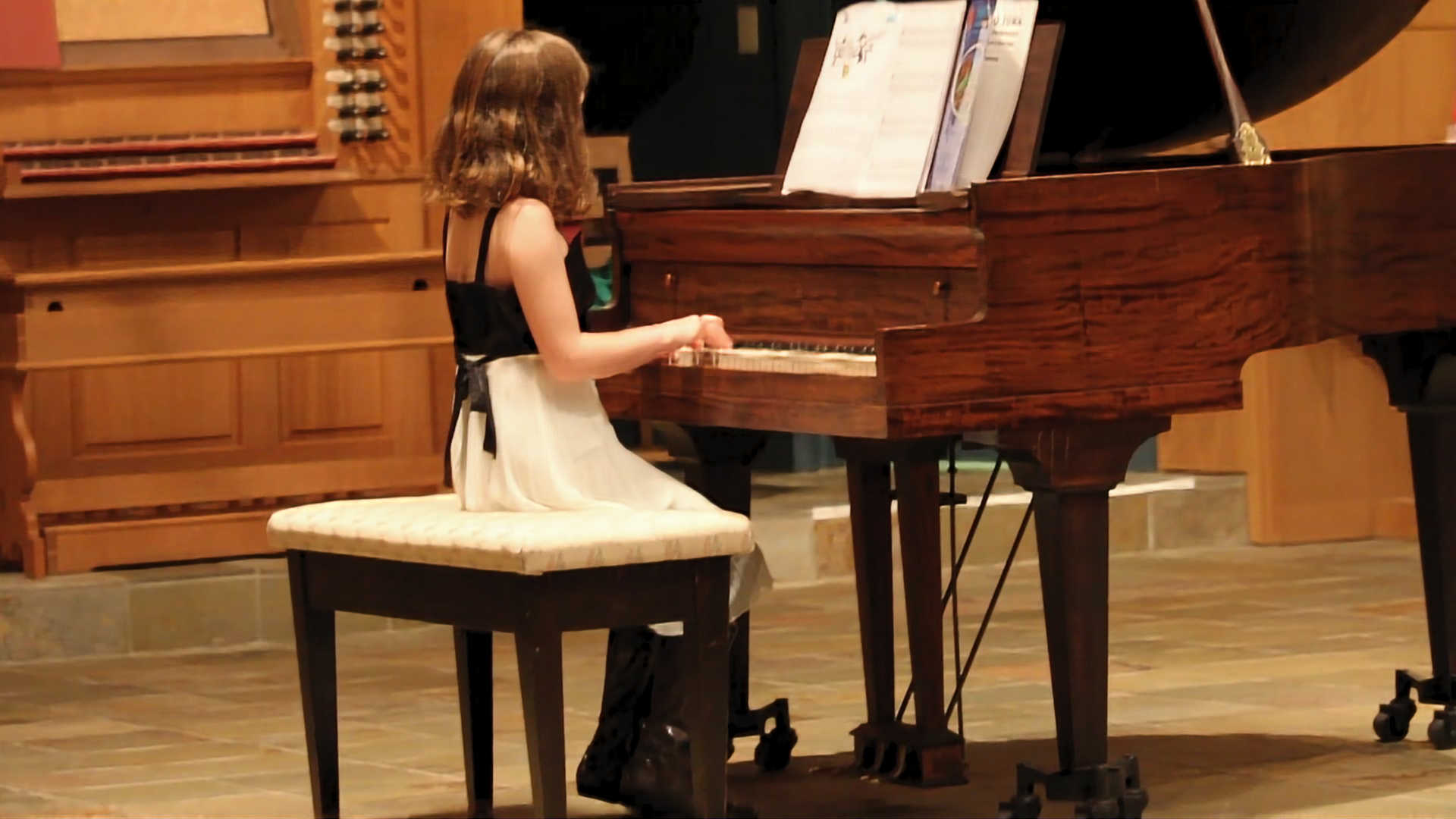} &
\pianoincludeb{figures/qualitative_comparisons/08_input} &
\includegraphics[height=.07\textheight]{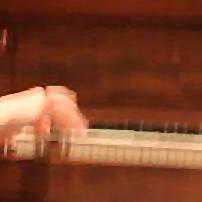} &
\pianoincludeb{figures/qualitative_comparisons/08_nndeblur} &
\pianoincludeb{figures/qualitative_comparisons/08_wfa_oliver} &
\pianoincludeb{figures/qualitative_comparisons/08_single_ms} &
\pianoincludeb{figures/qualitative_comparisons/08_none_baseline_deeper2} &
\pianoincludeb{figures/qualitative_comparisons/08_homography_baseline_deeper} &
\pianoincludeb{figures/qualitative_comparisons/08_OF_baseline_deeper2} \\

\includegraphics[height=.07\textheight,clip,trim=70 0 70 0]{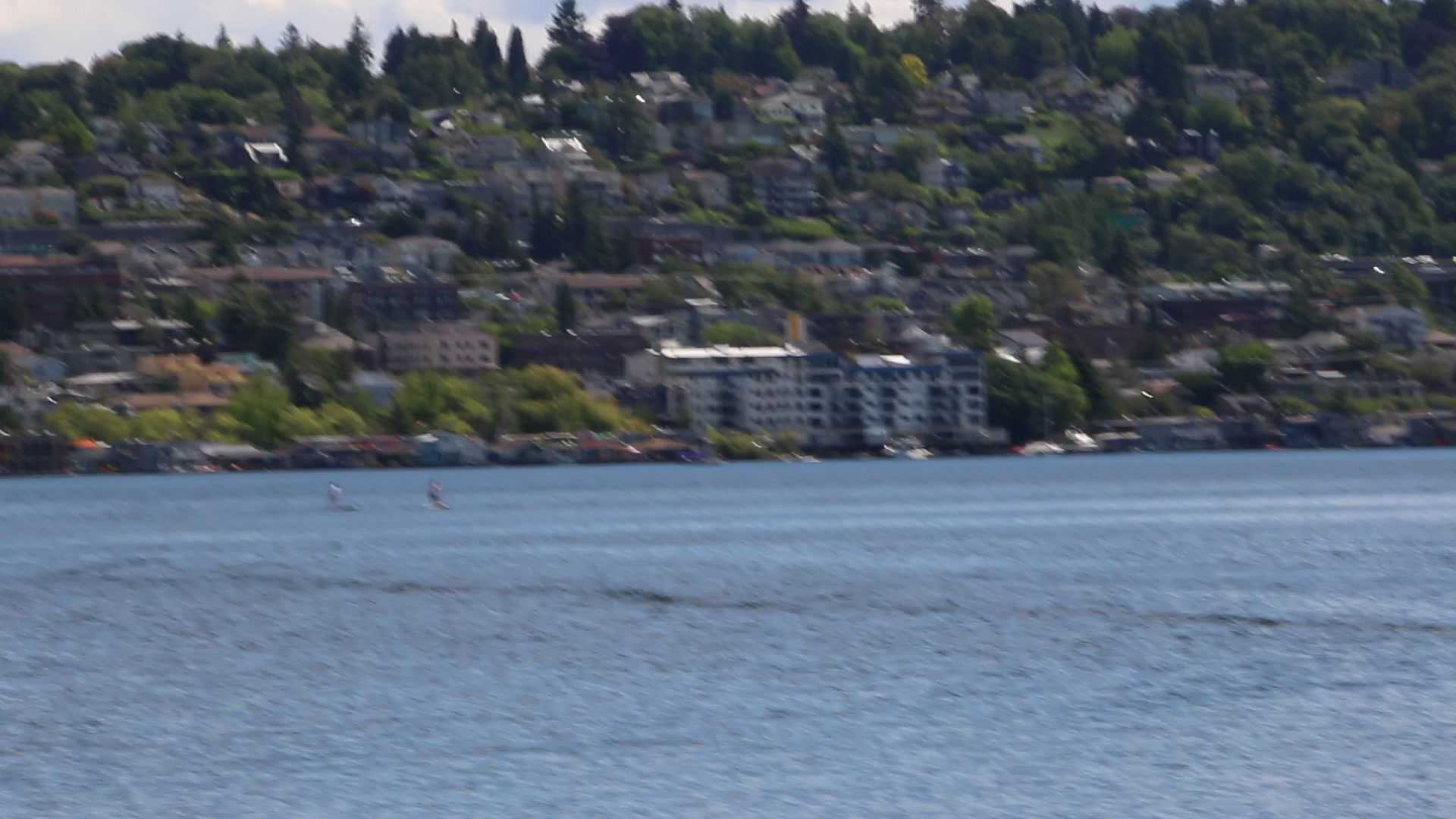} &
\supincludea{figures/qualitative_comparisons/04_input} &
\includegraphics[height=.07\textheight]{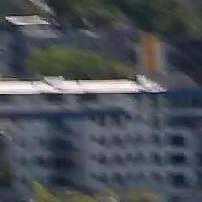} &
\supincludea{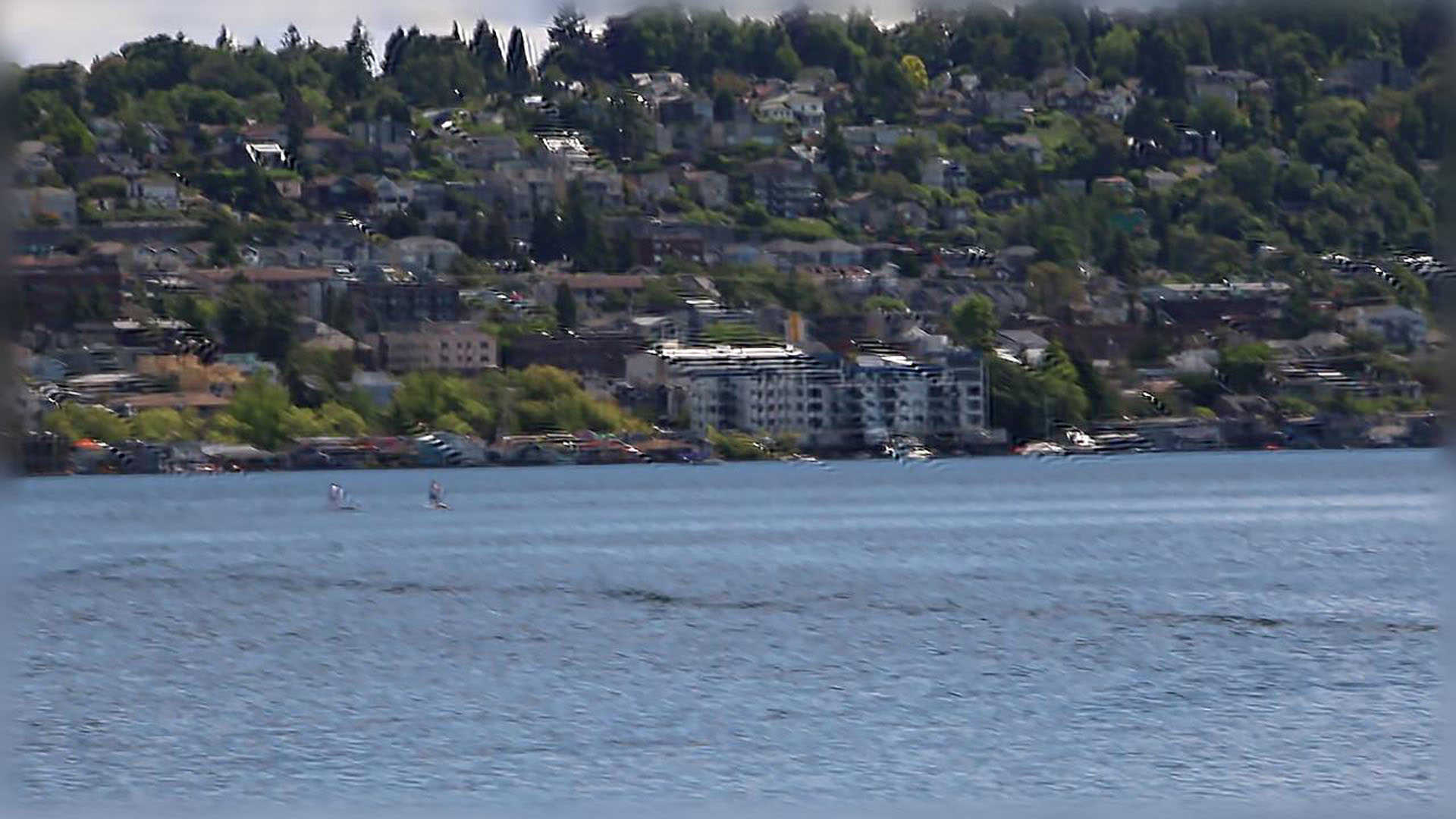} &
\supincludea{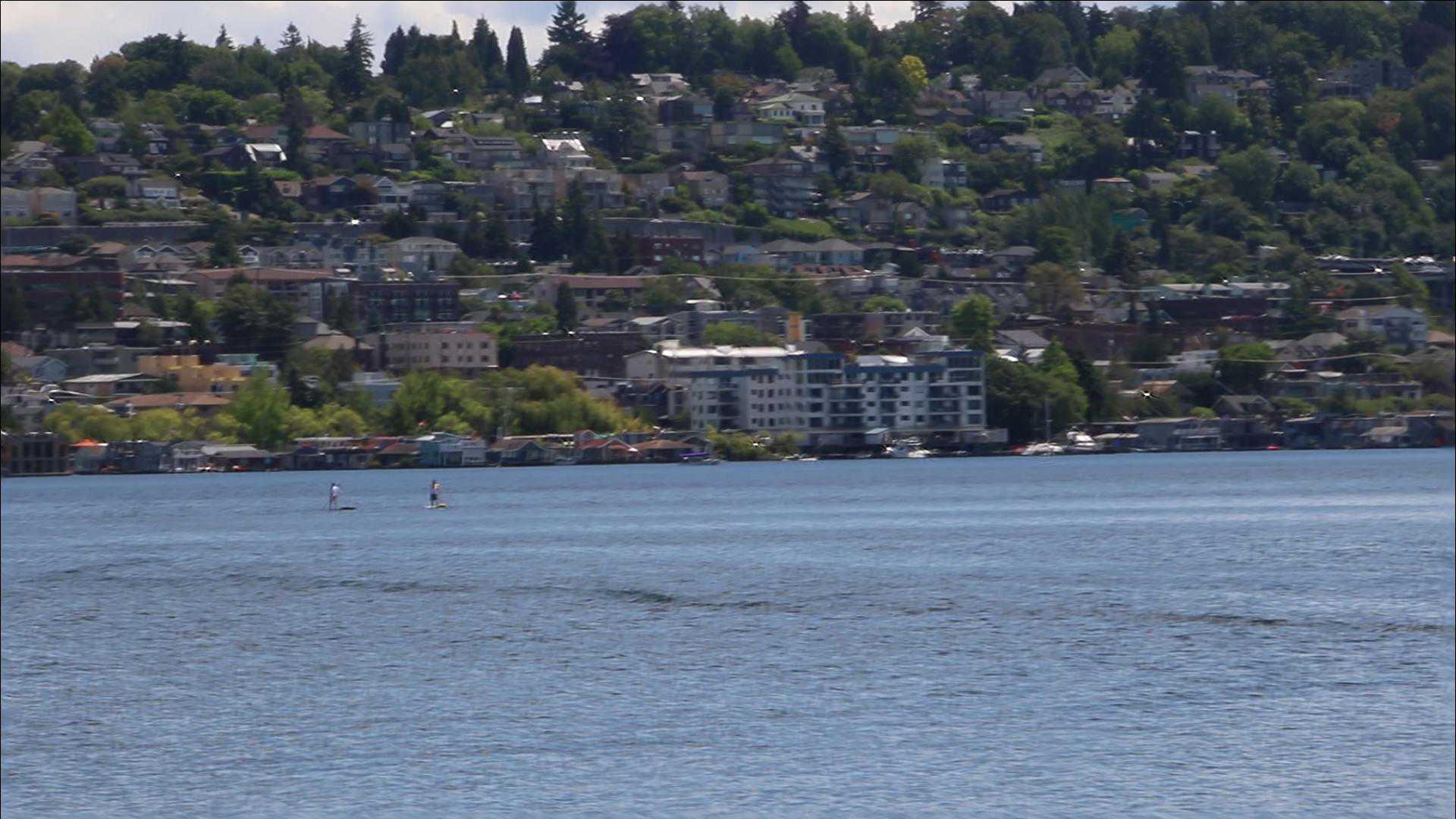} &
\supincludea{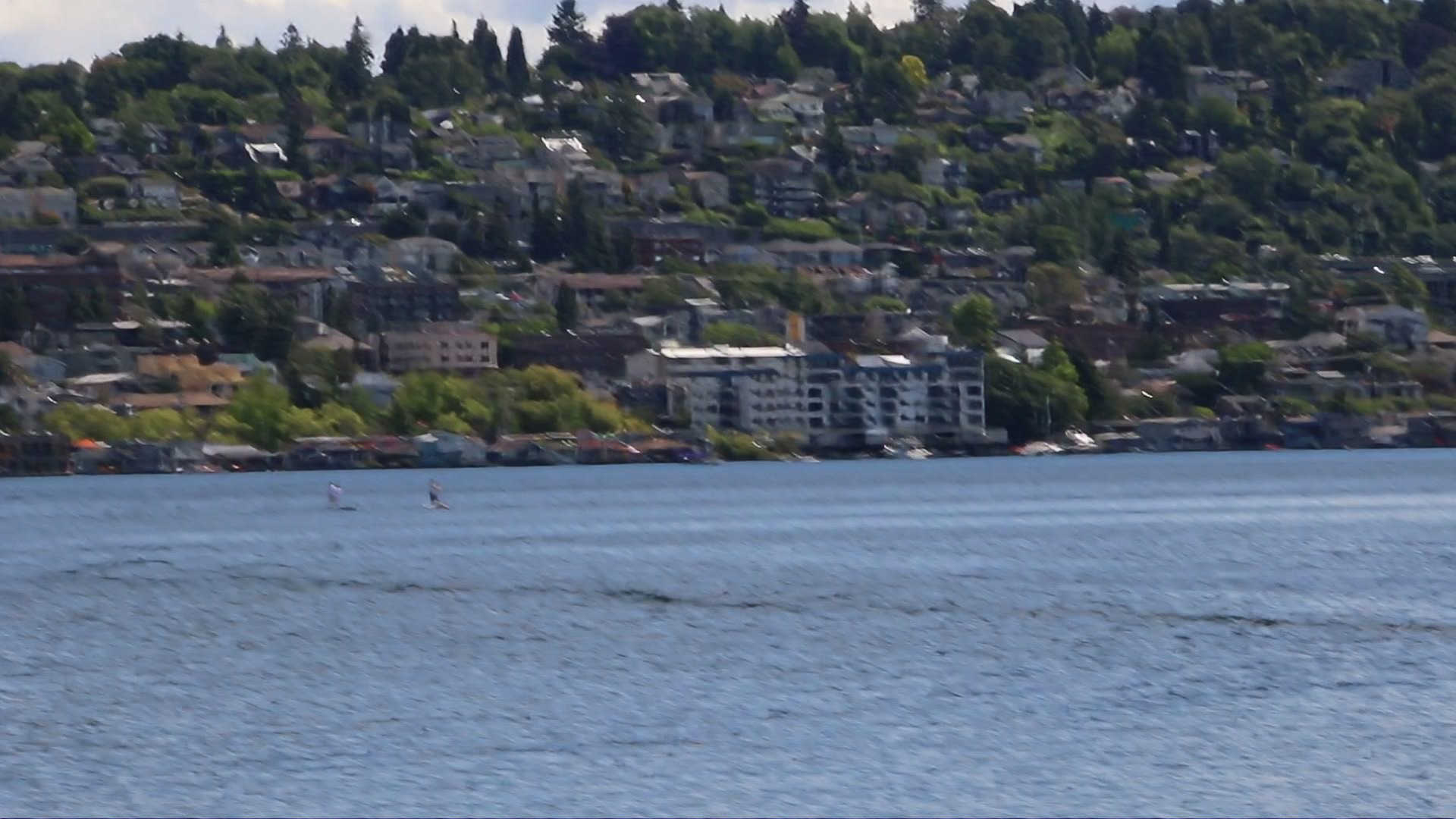} &
\supincludea{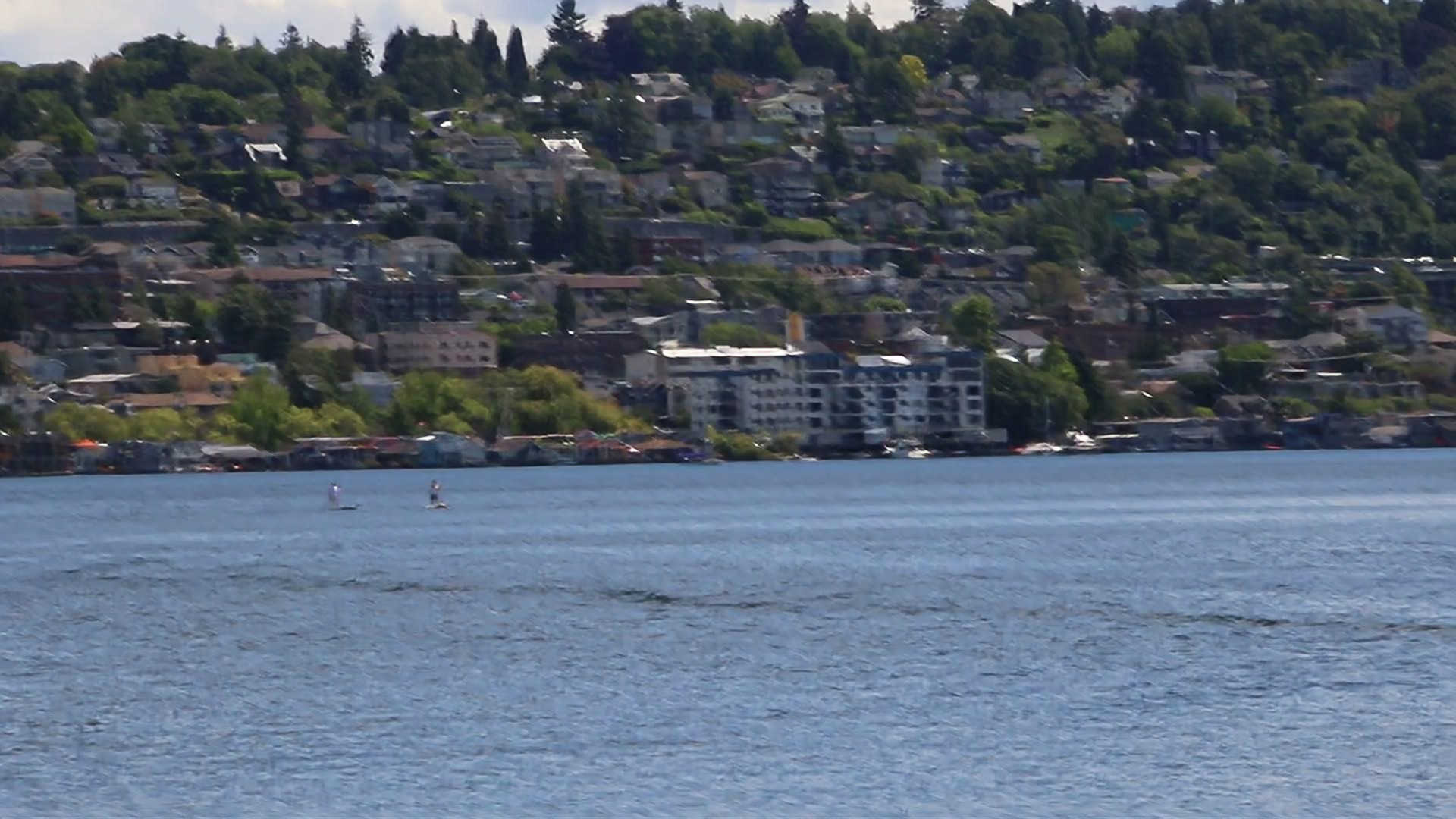} &
\supincludea{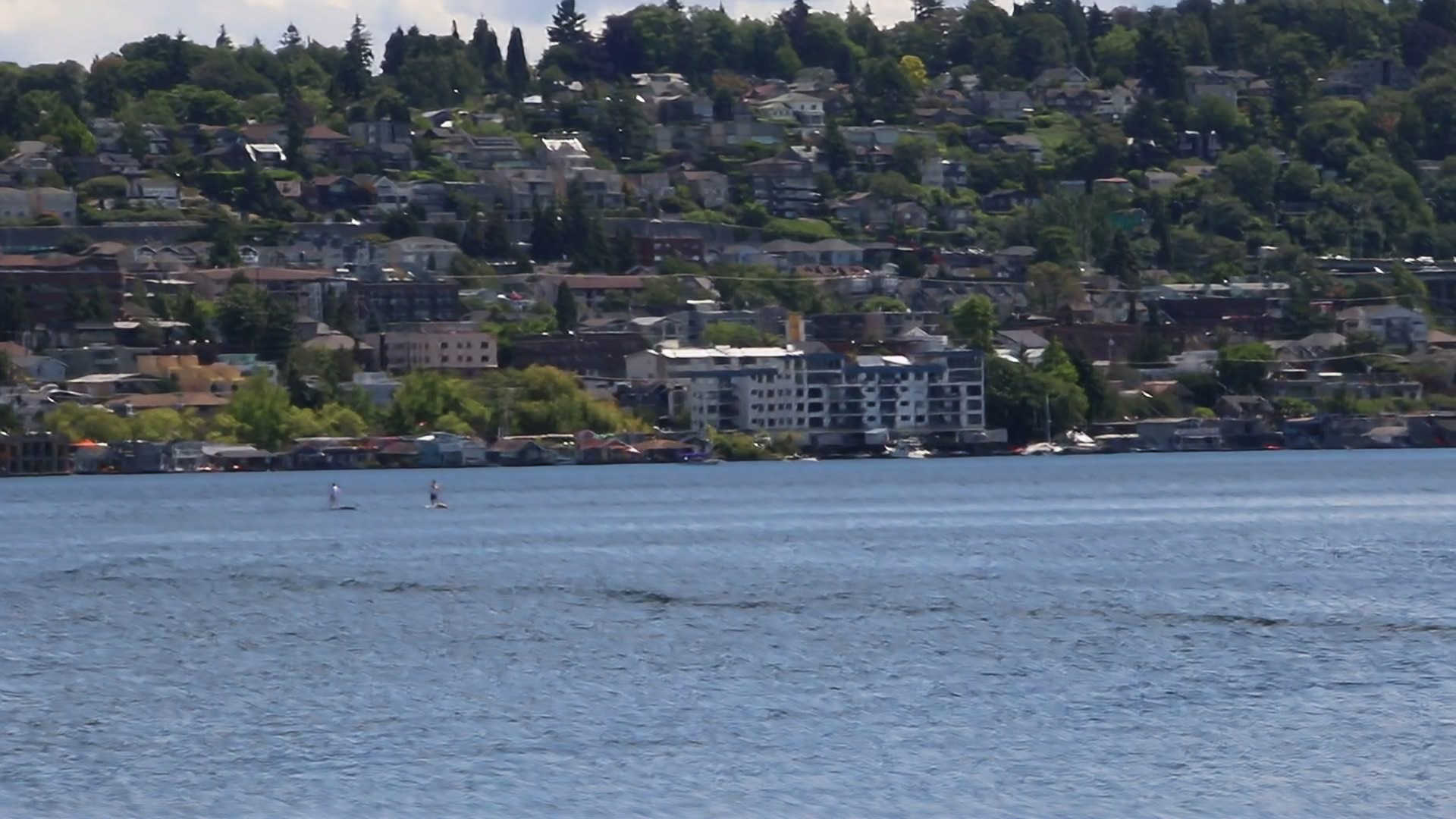} &
\supincludea{figures/qualitative_comparisons/04_OF_baseline_deeper2} \\

\includegraphics[height=.07\textheight,clip,trim=70 0 70 0]{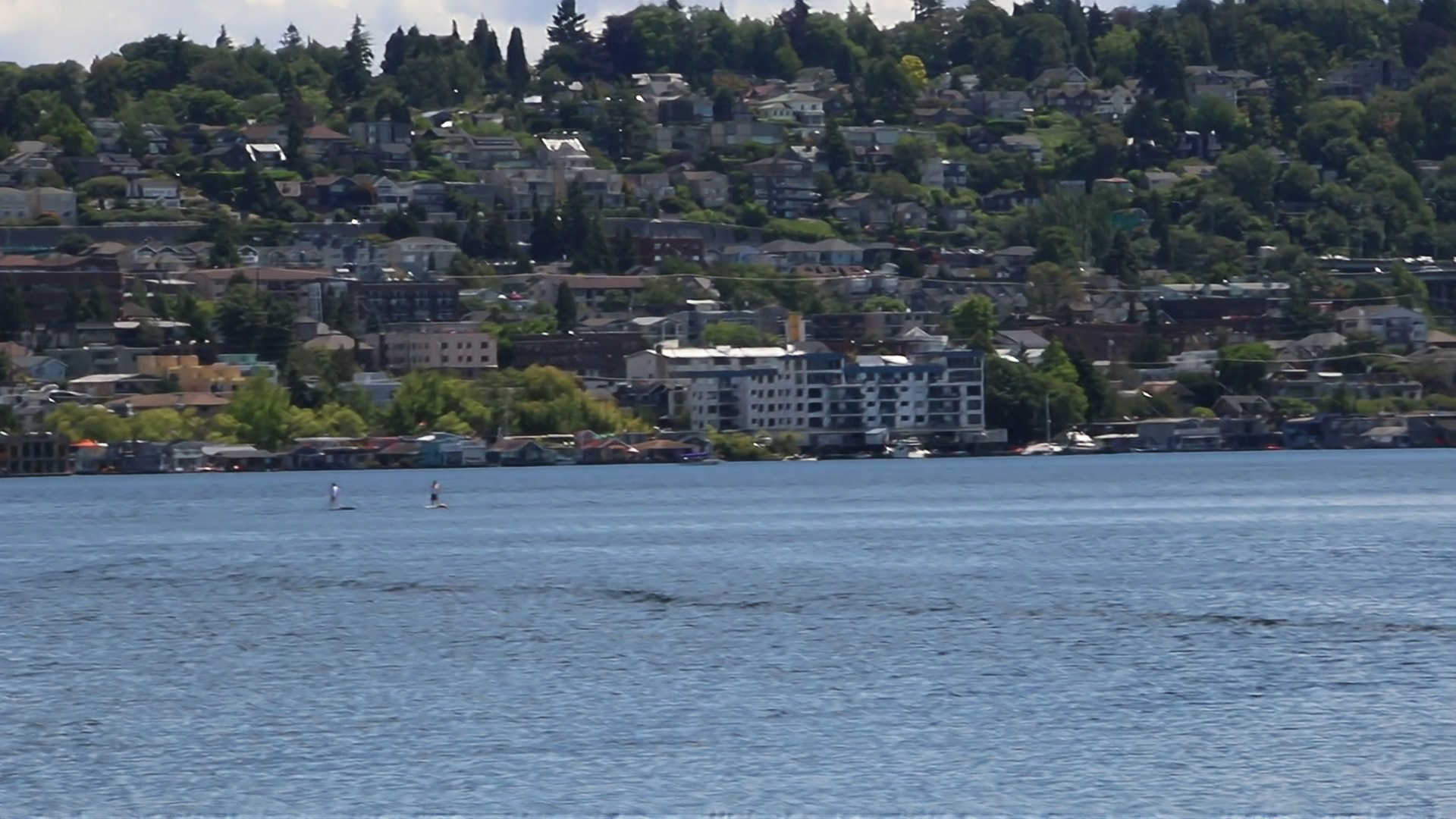} &
\supincludeb{figures/qualitative_comparisons/04_input} &
\includegraphics[height=.07\textheight]{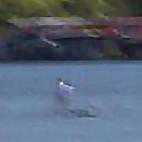} &
\supincludeb{figures/qualitative_comparisons/04_nndeblur} &
\supincludeb{figures/qualitative_comparisons/04_wfa_oliver} &
\supincludeb{figures/qualitative_comparisons/04_single_ms} &
\supincludeb{figures/qualitative_comparisons/04_none_baseline_deeper2} &
\supincludeb{figures/qualitative_comparisons/04_homography_baseline_deeper} &
\supincludeb{figures/qualitative_comparisons/04_OF_baseline_deeper2} \\

Input (top) / ours (bottom) & Input & \textsc{L0Deblur}\cite{xu2013unnatural} & \textsc{neural}~\cite{chakrabarti2016neural} & \textsc{wfa}~\cite{delbracio2015hand} & \textsc{dbn+single} & \textsc{dbn+noalign} & \textsc{dbn+homog} & \textsc{dbn+flow} \\

\end{tabular}
	}
	\vspace{1em}
	
	\resizebox{1.0\linewidth}{!}{
					\hspace{-20px}
	\centering
\scriptsize
\newcommand{\streetincludea}[1]{\includegraphics[height=.08\textheight,clip,trim=540 360 640 260]{#1}}
\newcommand{\streetincludeb}[1]{\includegraphics[height=.08\textheight,clip,trim=330 220 800 350]{#1}}
\newcommand{\bridgeincludea}[1]{\includegraphics[height=.08\textheight,clip,trim=600 200 540 380]{#1}}
\newcommand{\bridgeincludeb}[1]{\includegraphics[height=.08\textheight,clip,trim=880 200 240 360]{#1}}
\newcommand{\bookincludea}[1]{\includegraphics[height=.08\textheight,clip,trim=230 50 910 530]{#1}}
\newcommand{\bookincludeb}[1]{\includegraphics[height=.08\textheight,clip,trim=880 180 240 380]{#1}}

\def\arraystretch{0.5}
\begin{tabular}{*{10}{c@{\hspace{1.5px}}}}

\includegraphics[height=.08\textheight,clip,trim=50 0 50 0]{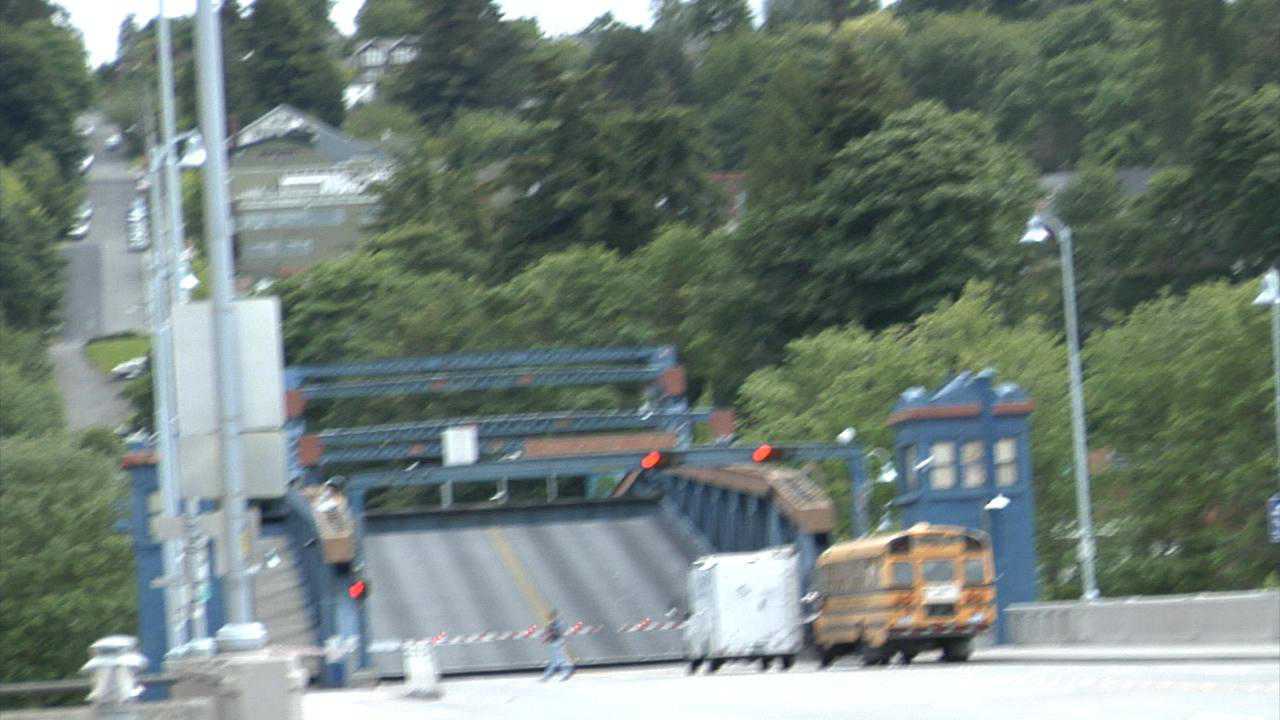} &
\bridgeincludea{figures/qualitative_comparisons/16_input} &
\bridgeincludea{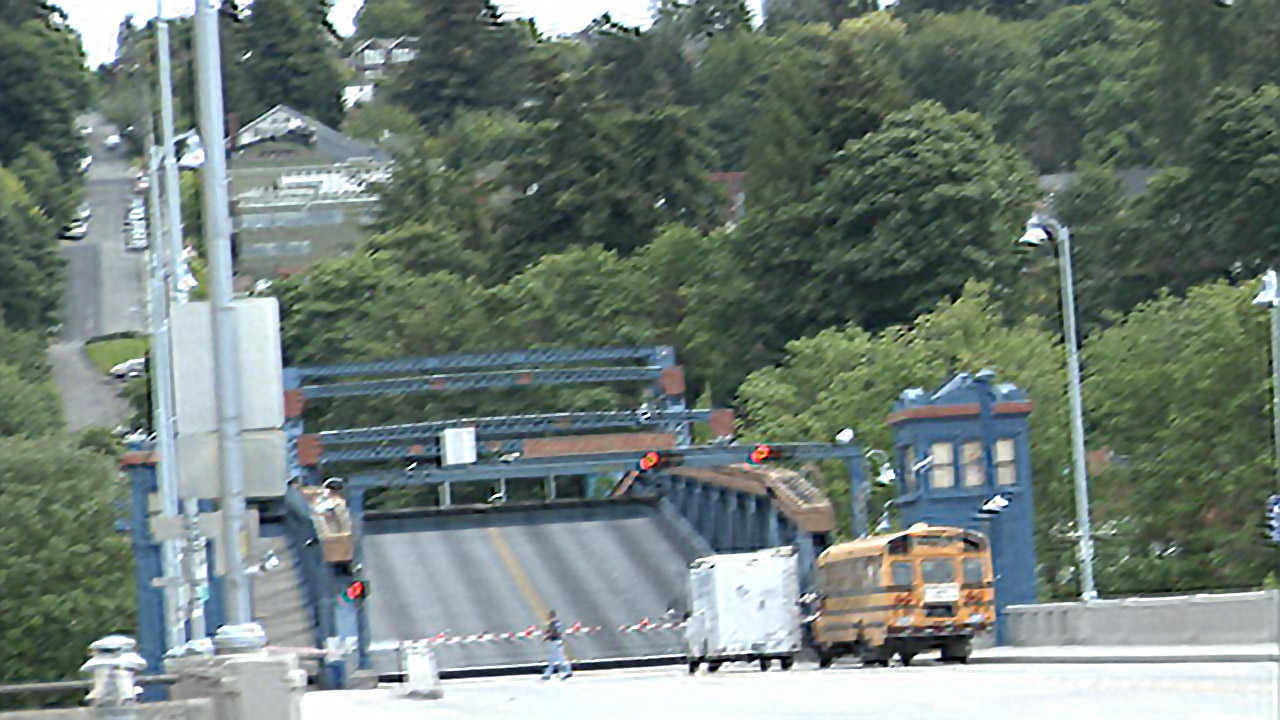} &
\bridgeincludea{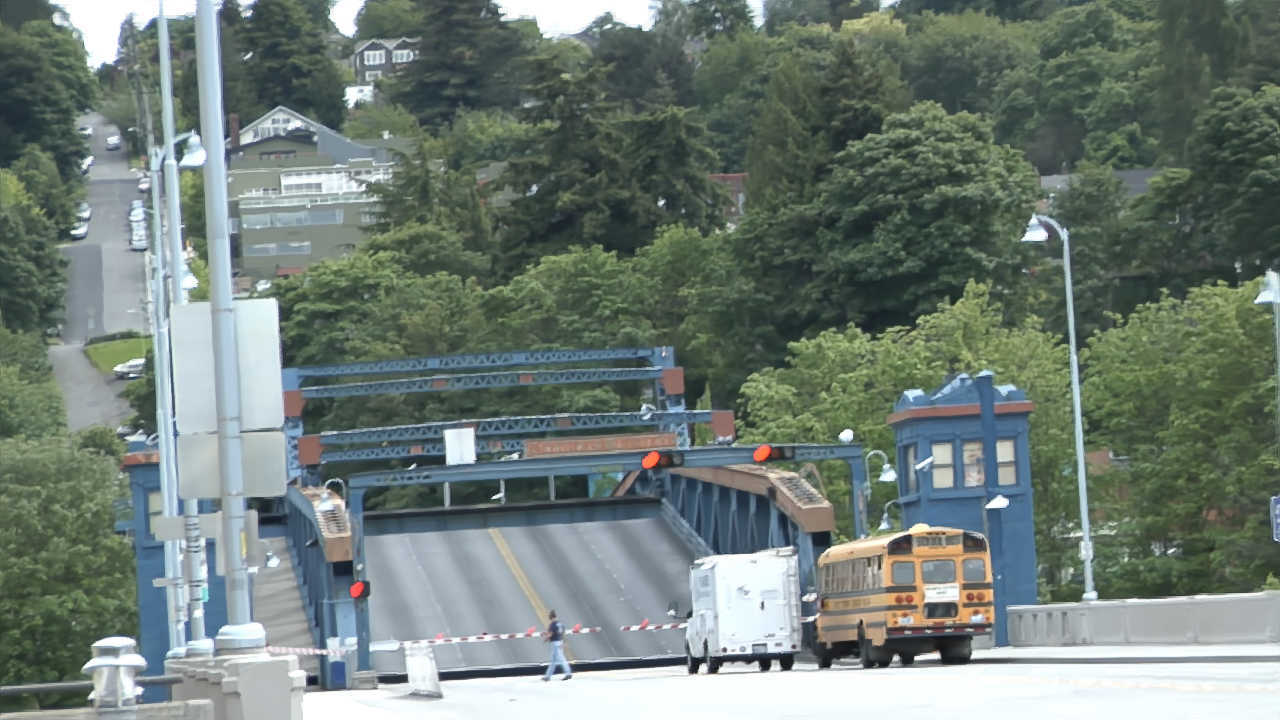} &
\bridgeincludea{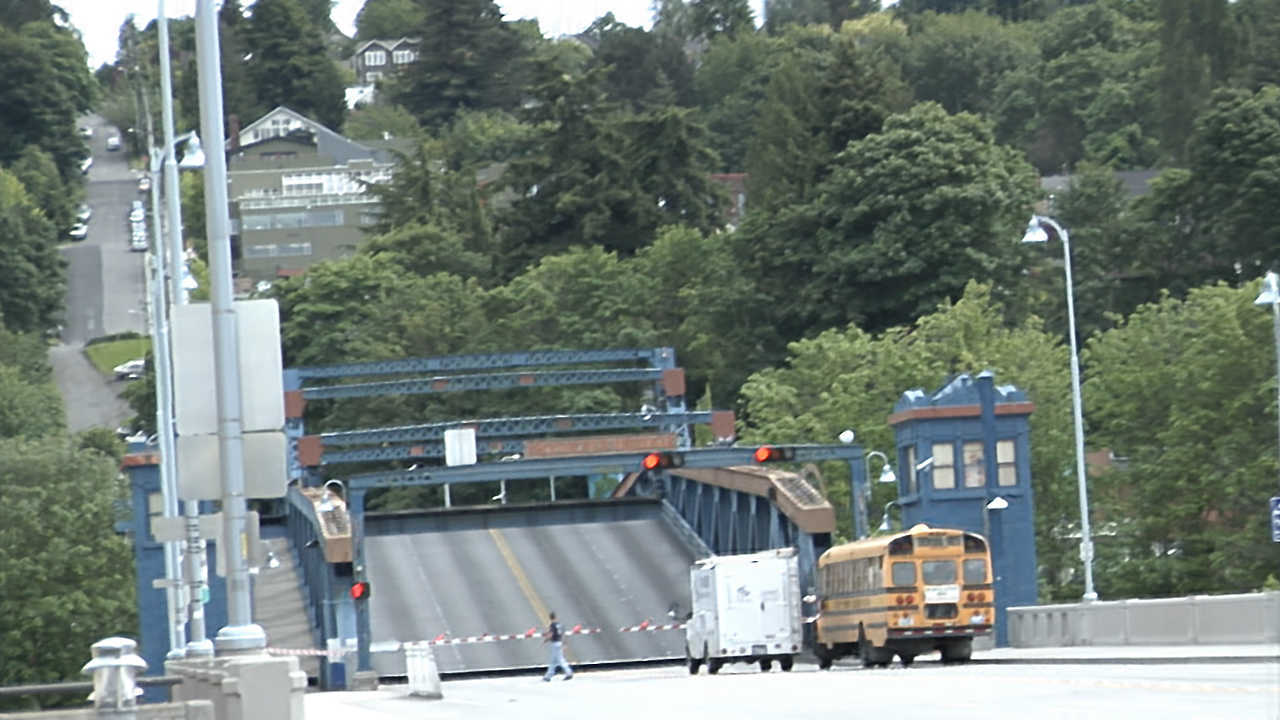} &
\bridgeincludea{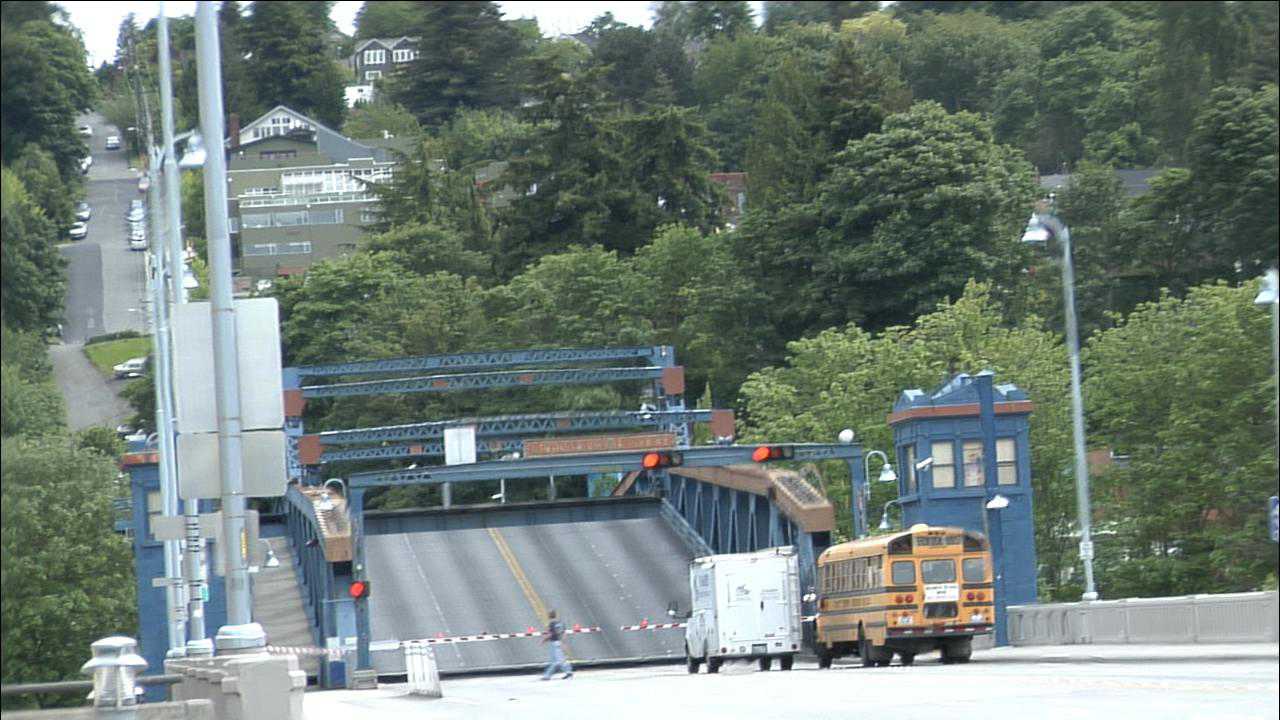} &
\bridgeincludea{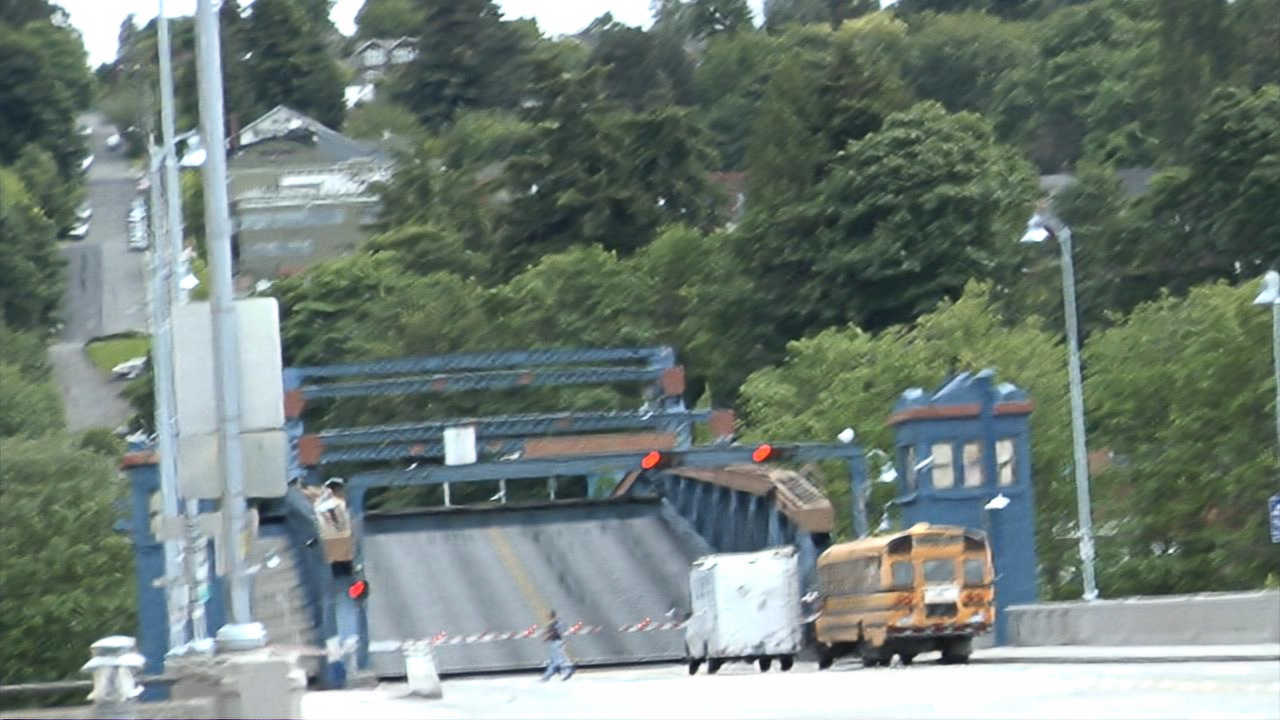} &
\bridgeincludea{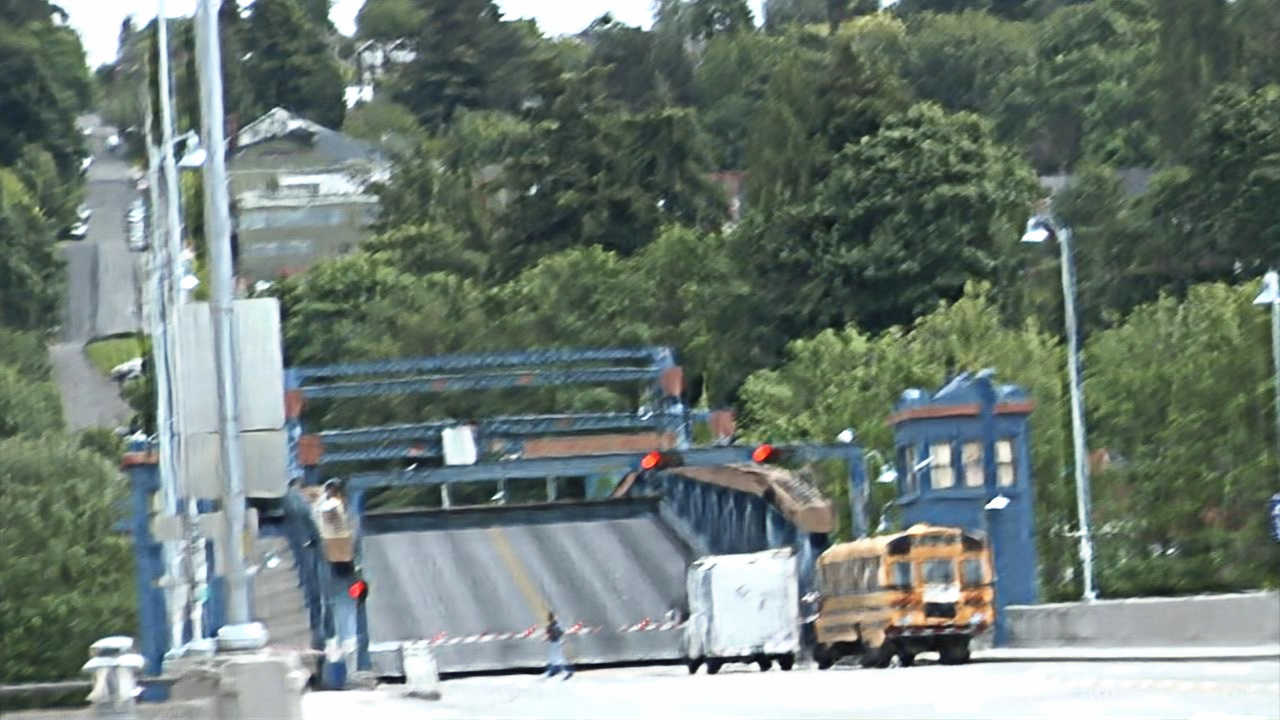} &
\bridgeincludea{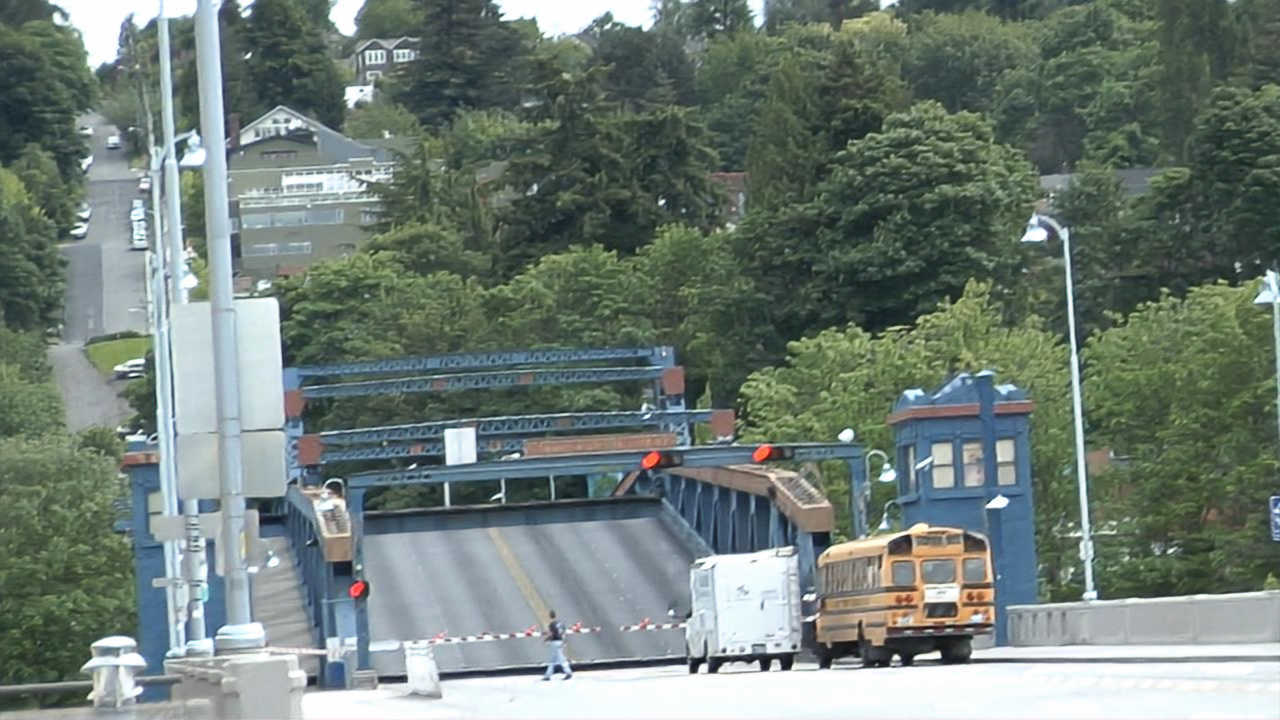} &
\bridgeincludea{figures/qualitative_comparisons/16_OF_baseline_deeper2} \\

\includegraphics[height=.08\textheight,clip,trim=50 0 50 0]{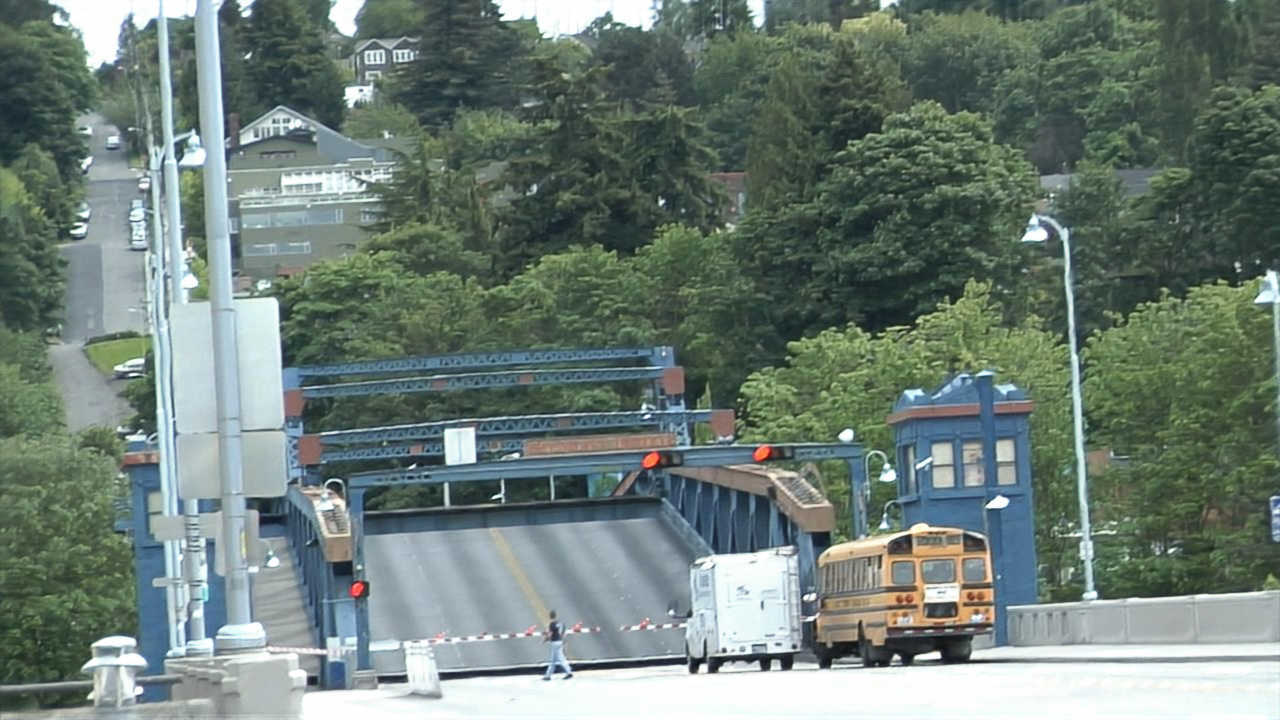} &
\bridgeincludeb{figures/qualitative_comparisons/16_input} &
\bridgeincludeb{figures/qualitative_comparisons/16_ps_deblur} &
\bridgeincludeb{figures/qualitative_comparisons/16_cho} &
\bridgeincludeb{figures/qualitative_comparisons/16_kim} &
\bridgeincludeb{figures/qualitative_comparisons/16_wfa_oliver} &
\bridgeincludeb{figures/qualitative_comparisons/16_single_ms} &
\bridgeincludeb{figures/qualitative_comparisons/16_none_baseline_deeper2} &
\bridgeincludeb{figures/qualitative_comparisons/16_homography_baseline_deeper} &
\bridgeincludeb{figures/qualitative_comparisons/16_OF_baseline_deeper2} \\

\includegraphics[height=.08\textheight,clip,trim=50 0 50 0]{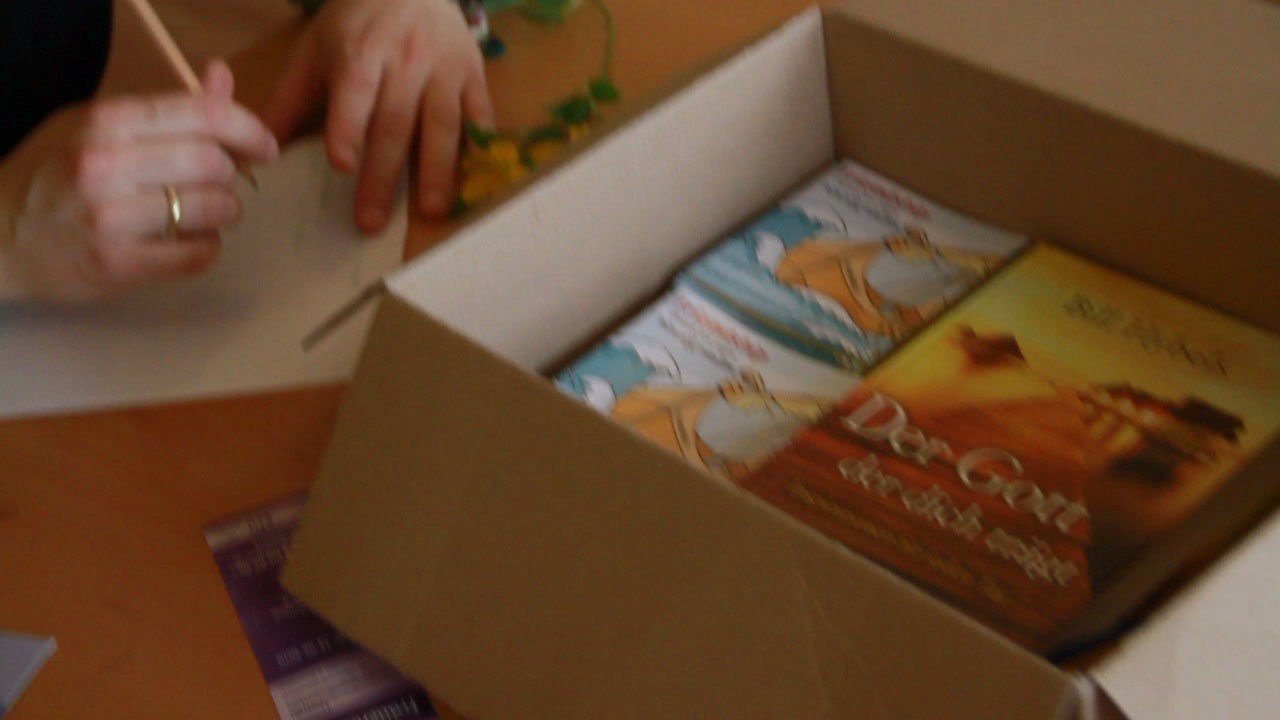} &
\bookincludea{figures/qualitative_comparisons/17_input} &
\bookincludea{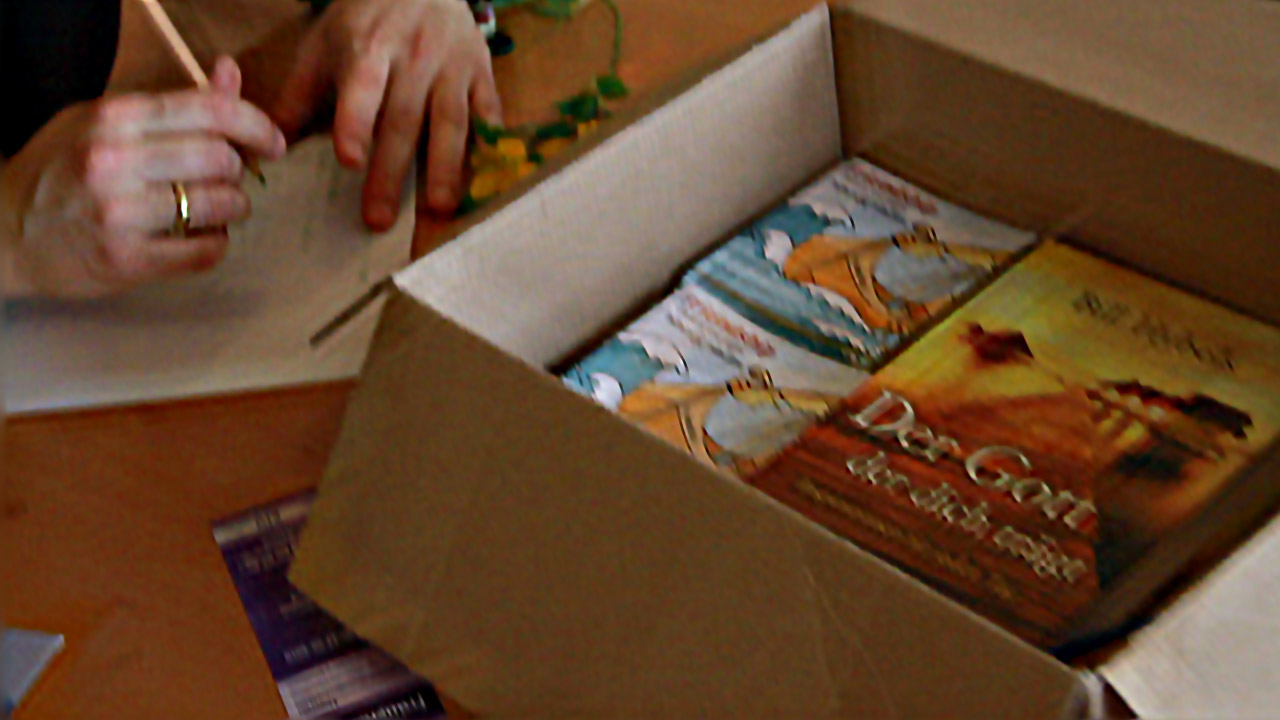} &
\bookincludea{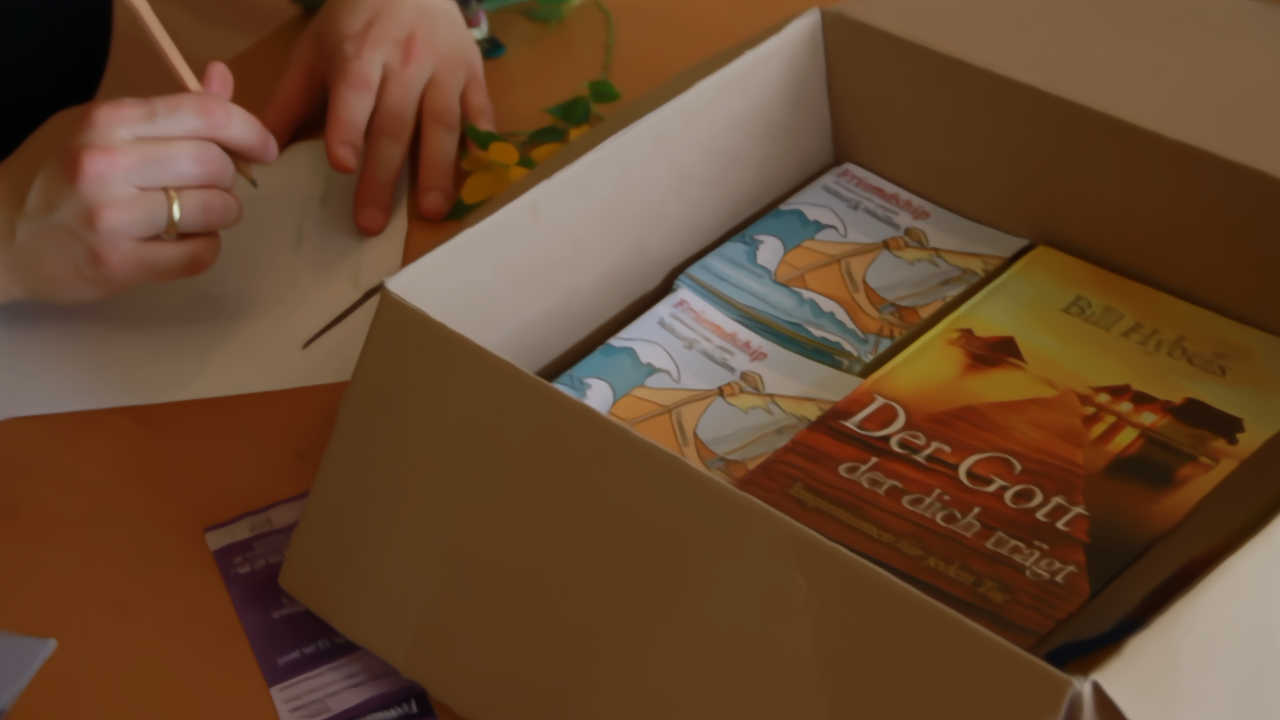} &
\bookincludea{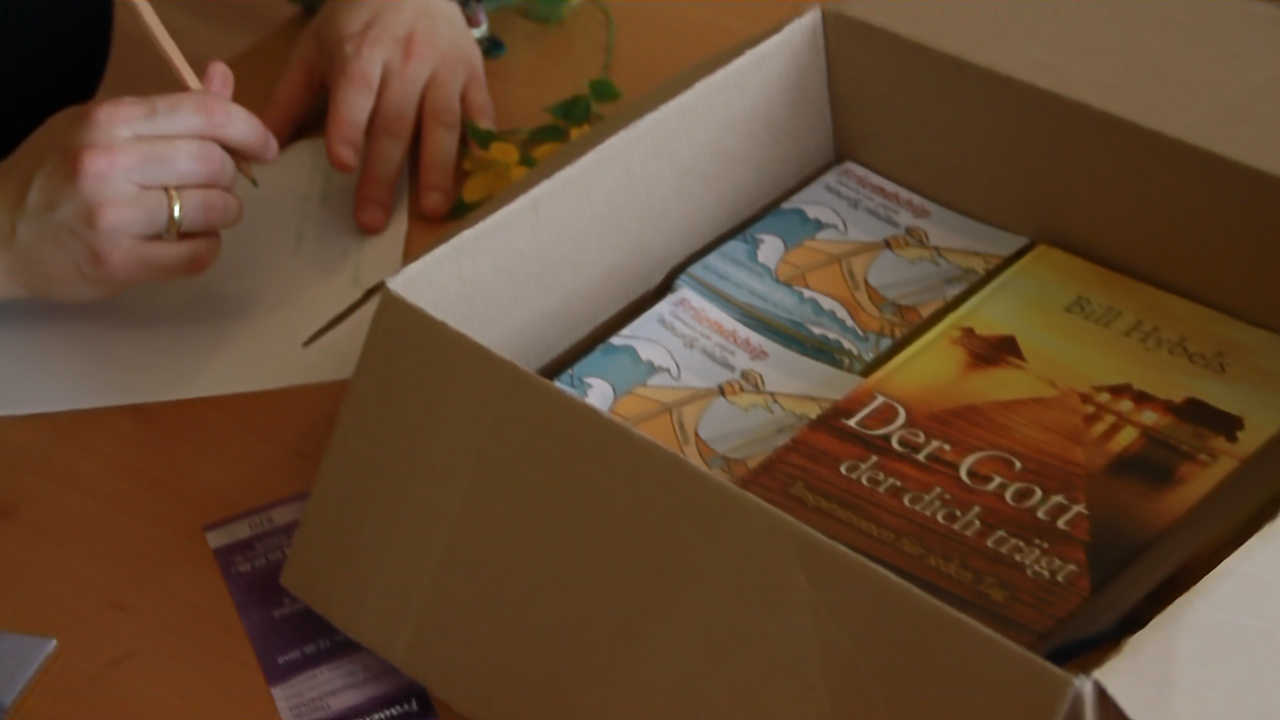} &
\bookincludea{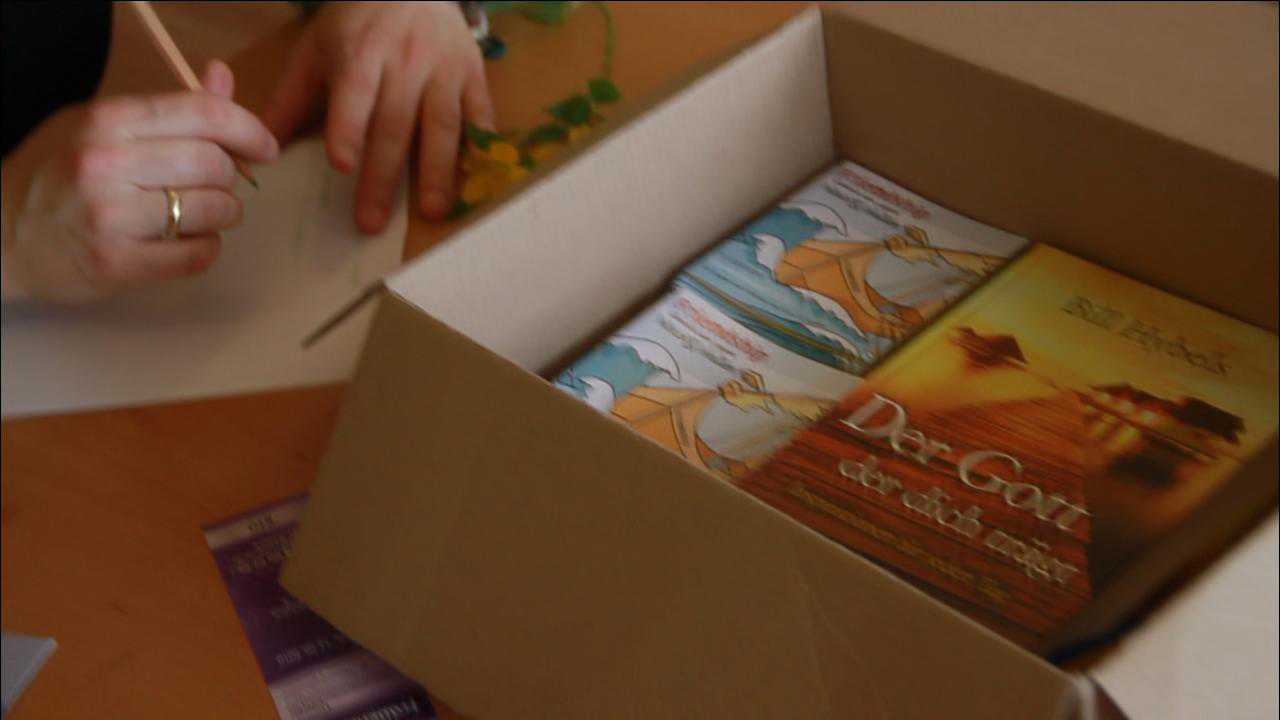} &
\bookincludea{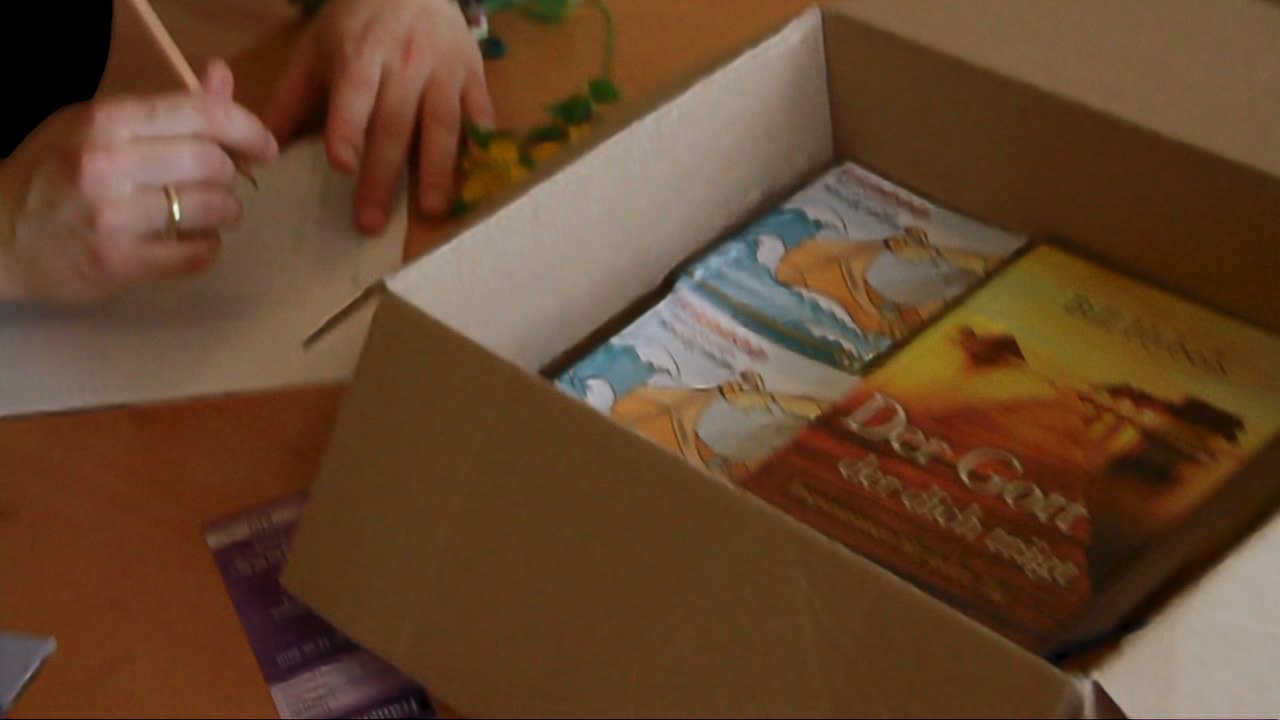} &
\bookincludea{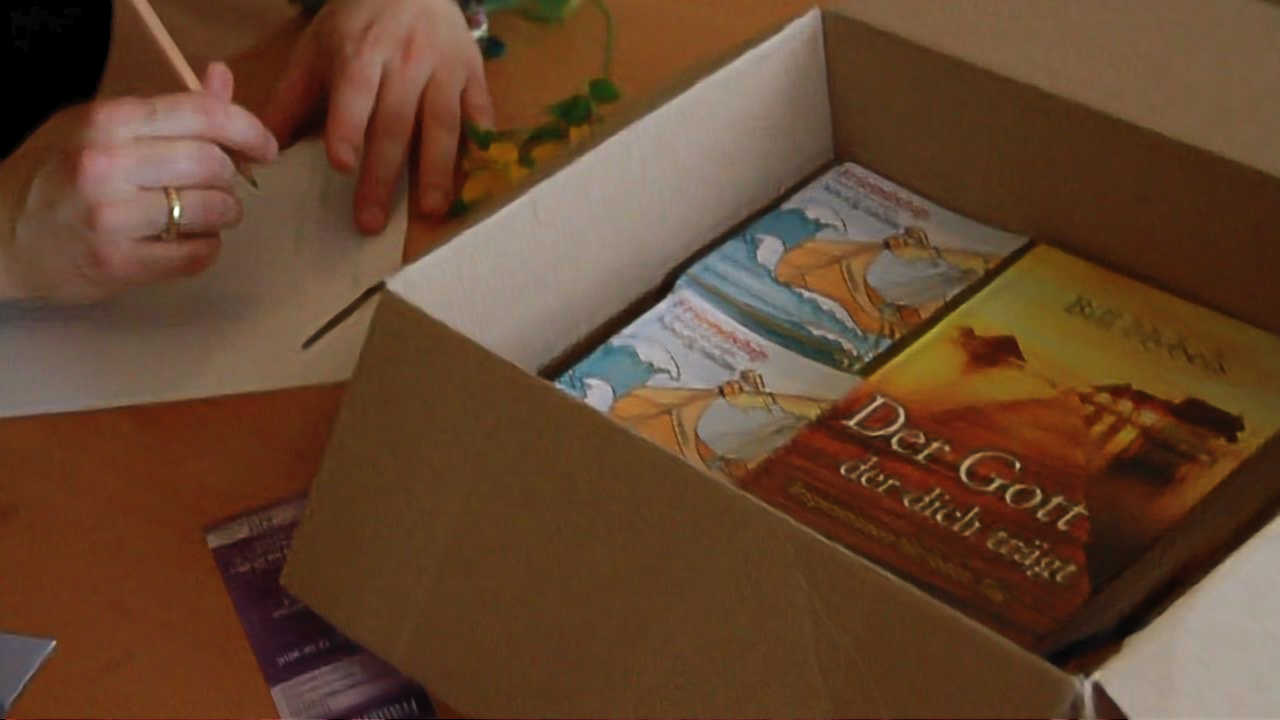} &
\bookincludea{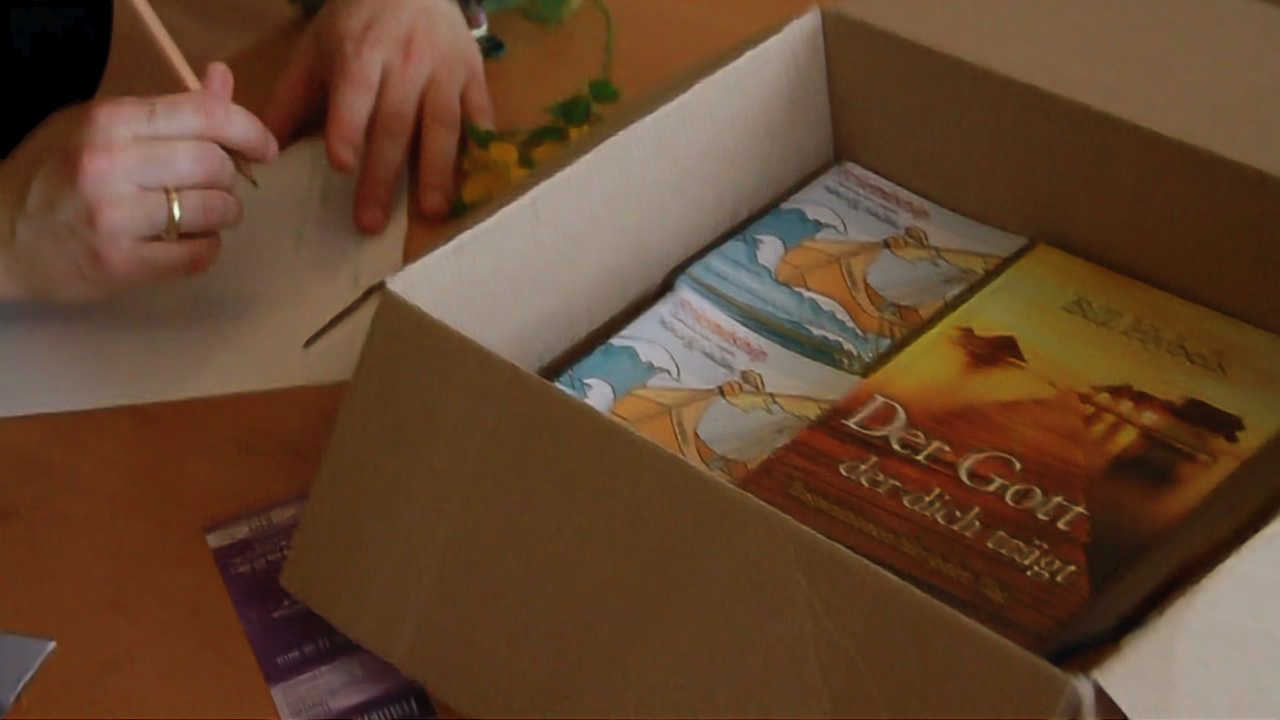} &
\bookincludea{figures/qualitative_comparisons/17_OF_baseline_deeper2} \\

\includegraphics[height=.08\textheight,clip,trim=50 0 50 0]{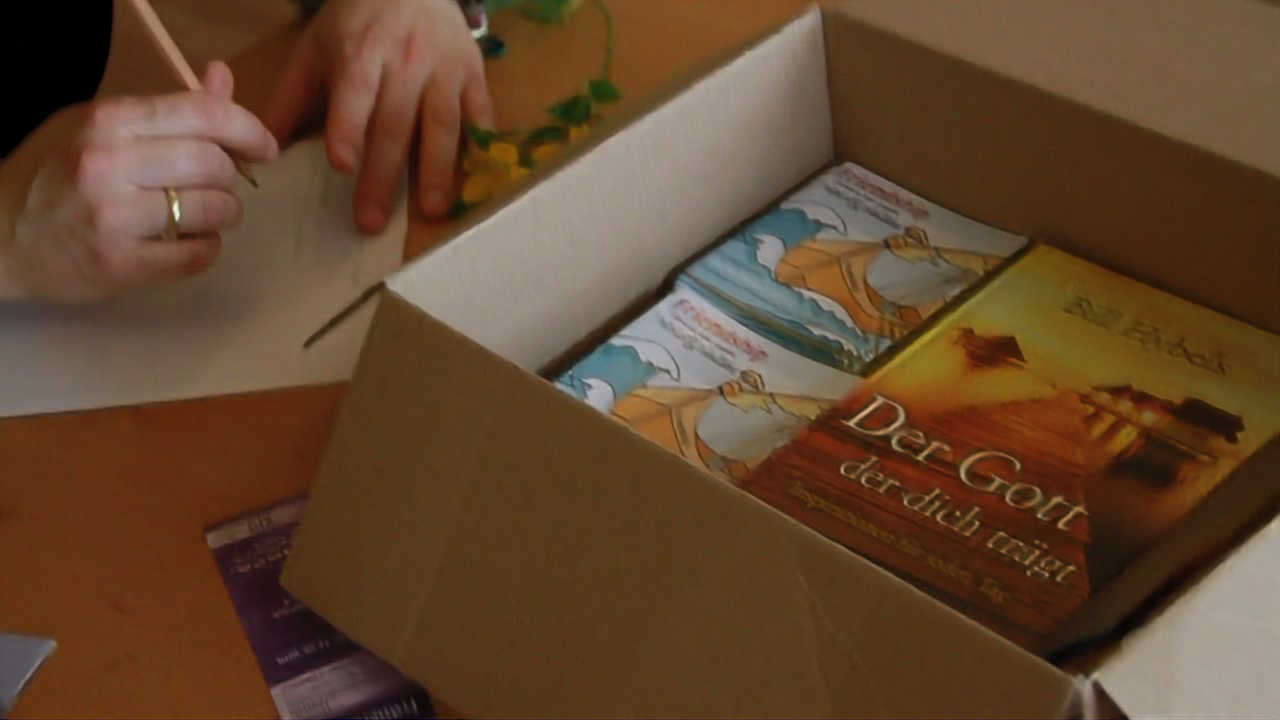} &
\bookincludeb{figures/qualitative_comparisons/17_input} &
\bookincludeb{figures/qualitative_comparisons/17_ps_deblur} &
\bookincludeb{figures/qualitative_comparisons/17_cho} &
\bookincludeb{figures/qualitative_comparisons/17_kim} &
\bookincludeb{figures/qualitative_comparisons/17_wfa_oliver} &
\bookincludeb{figures/qualitative_comparisons/17_single_ms} &
\bookincludeb{figures/qualitative_comparisons/17_none_baseline_deeper2} &
\bookincludeb{figures/qualitative_comparisons/17_homography_baseline_deeper} &
\bookincludeb{figures/qualitative_comparisons/17_OF_baseline_deeper2} \\

Input (top) / ours (bottom) & Input & \textsc{PSdeblur} & Cho et al.~\cite{cho2012video} & Kim and Lee~\cite{kim2015cvpr} & \textsc{wfa}~\cite{delbracio2015hand} & \textsc{dbn+single} & \textsc{dbn+noalign} & \textsc{dbn+homog} & \textsc{dbn+flow} \\

\end{tabular}
}
	\caption{Qualitative comparisons to existing approaches. We compare \textsc{dbn} under various alignment configurations, with prior approaches, e.g. Cho et al.~\cite{cho2012video}, Kim and Lee~\cite{kim2015cvpr}, Chakrabarti~\cite{chakrabarti2016neural}, Xu et al.~\cite{xu2013unnatural}, \textsc{wfa}~\cite{delbracio2015hand}, and Photoshop CC Shake Reduction. 
	In general \textsc{dbn} achieves decent quality without alignment, and is comparable or better when simpler frame-wise homography is applied. Note that~\cite{cho2012video} adapts homography-based motion model, while~\cite{delbracio2015hand} and~\cite{kim2015cvpr} are estimating the optical flow for alignment.}
	\label{fig:deblurnet_comparison} 
\end{figure*}

\paragraph{Effects of using multiple frames.}
We analyze the contribution of using a temporal window by keeping the same network architecture as \textsc{dbn}, but replicating the central reference frame 5 times instead of inputing a stack of neighboring frames, and retrain the network with the same hyper-parameters.
We call this approach \textsc{dbn+single}.
Qualitative comparisons are shown in Fig.~\ref{fig:deblurnet_comparison} and~\ref{fig:deblurnet_comparison_3}, and quantitative results are shown in Table~\ref{tab:quantitative}, and Fig.~\ref{fig:quantitative}.
We can see that using neighboring frames greatly improves the quality of the results.
We chose a 5 frame window as it provides a good compromise between result quality and training time~\cite{kappeler2016video}.
\emph{Single-image} methods are also provided as reference: \textsc{PSdeblur} for blind uniform deblurring with off-the-shelf shake reduction software in Photoshop, and~\cite{xu2013unnatural} for non-uniform comparisons.

\paragraph{Effects of alignment.}
In this set of experiments, we analyze the impact of input image alignment in the output restoration quality, namely we compare the results of \textsc{dbn+noalign}, \textsc{dbn+homog.}, and \textsc{dbn+flow}.
See Tab.~\ref{tab:quantitative}, and Fig.~\ref{fig:quantitative}, for quantitative comparisons, and the qualitative comparison in Fig.~\ref{fig:deblurnet_comparison}.
Our main conclusions are that \textsl{DeBlurNet} with optical flow and homography are often qualitatively equivalent, and \textsc{dbn+flow} often has higher PSNR.
On the other hand, \textsc{dbn+noalign} performs even better than \textsc{dbn+flow} and \textsc{dbn+homog} in terms of PSNR, especially when the input frames are not too blurry, e.g. $>$29dB.
However, we observe that \textsc{dbn+flow} fails gracefully when inputs frame are much blurrier, which leads to a drop in PSNR and MSSIM (see Tab.~\ref{tab:quantitative} and Fig.~\ref{fig:quantitative}). 
In this case, \textsc{dbn+flow} and \textsc{dbn+homog.} perform better. 
One possible explanation for this is that when the input quality is good, optical flow errors will dominate the final performance of the deblurring procedure. 
Indeed, sequences with high input PSNR have small relative motion (consequence of how the dataset is created) so there is not too much displacement from one frame to the next, and \textsc{dbn+noalign} is able to directly handle the input frames without any alignment.

\paragraph{Comparisons to existing approaches.}
We compare our method to existing approaches in Fig.~\ref{fig:deblurnet_comparison}. 
Specifically, we show a quantitative comparison to WFA~\cite{delbracio2015hand}, and qualitative comparisons to Cho et al.~\cite{cho2012video}, Kim et al.~\cite{kim2016dynamic}, and WFA~\cite{delbracio2015hand}.
We also compare to \emph{single image} deblurring methods, Chakrabarti~\cite{chakrabarti2016neural}, Xu et al.~\cite{xu2013unnatural}, and the Shake Reduction feature in Photoshop CC 2015 (\textsc{PSdeblur}).
We note that \textsc{PSdeblur} can cause ringing artifacts when used in an automatic setting on sharp images, resulting in a sharp degradation in quality (Fig.~\ref{fig:quantitative}). 
The results of \cite{cho2012video} and \cite{kim2016dynamic} are the ones provided by the authors, WFA~\cite{delbracio2015hand} was applied a single iteration with the same temporal window, and for \cite{xu2013unnatural,chakrabarti2016neural} we use the implementations provided by the authors.
Due to the large number of frames, we are only able to compare quantitatively to approaches which operate sufficiently fast, which excludes many non-uniform deconvolution based methods.
The complete sequences are given in the supplementary material.
It is important to note that the test images have not been seen during the training procedure, and many of them have been shot by other cameras. 
Our conclusion is that \textsc{dbn} often produces superior quality deblurred frames, even when the input frames are aligned with a global homography, which requires substantially less computation than prior methods. 

\begin{figure}[ht]
		\centering
	\footnotesize
	\newcommand{\darka}[1]{\includegraphics[height=2cm,width=.23\linewidth,clip,trim=800 100 820 700]{#1}}
	\newcommand{\darkb}[1]{\includegraphics[height=2cm,width=.23\linewidth,clip,trim=1590 190 180 720]{#1}}
	\newcommand{\cara}[1]{\includegraphics[height=2cm,width=.23\linewidth,clip,trim=160 540 950 20]{#1}}
	\newcommand{\carb}[1]{\includegraphics[height=2cm,width=.23\linewidth,clip,trim=530 350 470 50]{#1}}
	\begin{tabular}{*{2}{c@{\hspace{3px}}}}
	\includegraphics[width=0.465\linewidth]{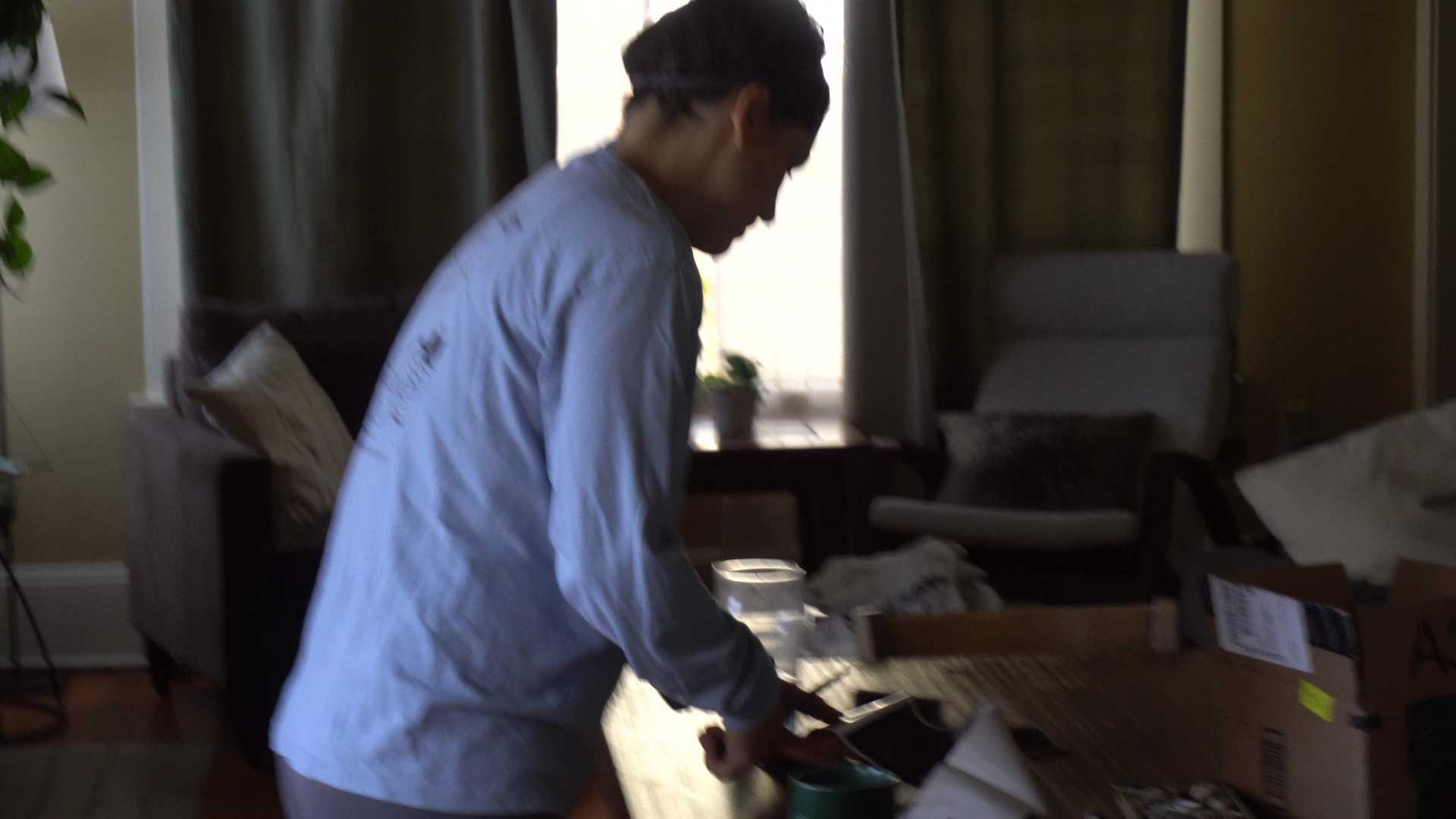} &
	\includegraphics[width=0.465\linewidth]{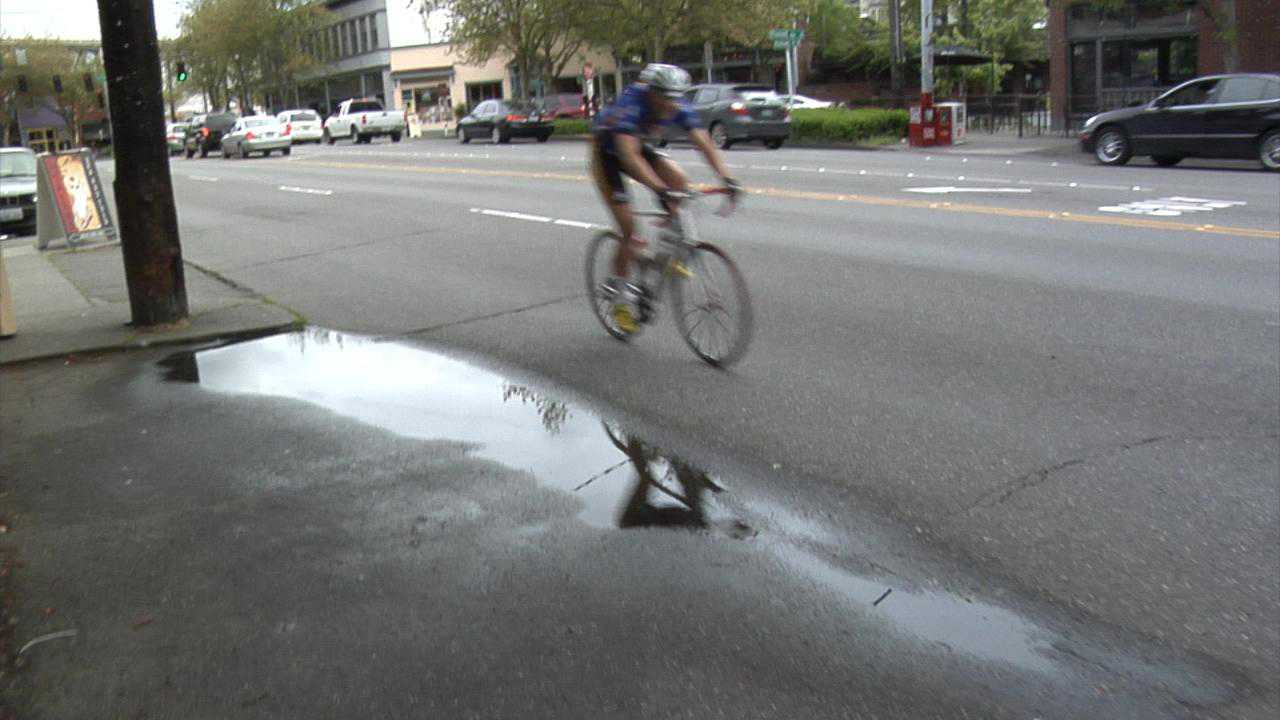} \\
	\end{tabular}
	\begin{tabular}{*{4}{c@{\hspace{2px}}}}
		\darka{figures/generation/dark_in} & 
		\darka{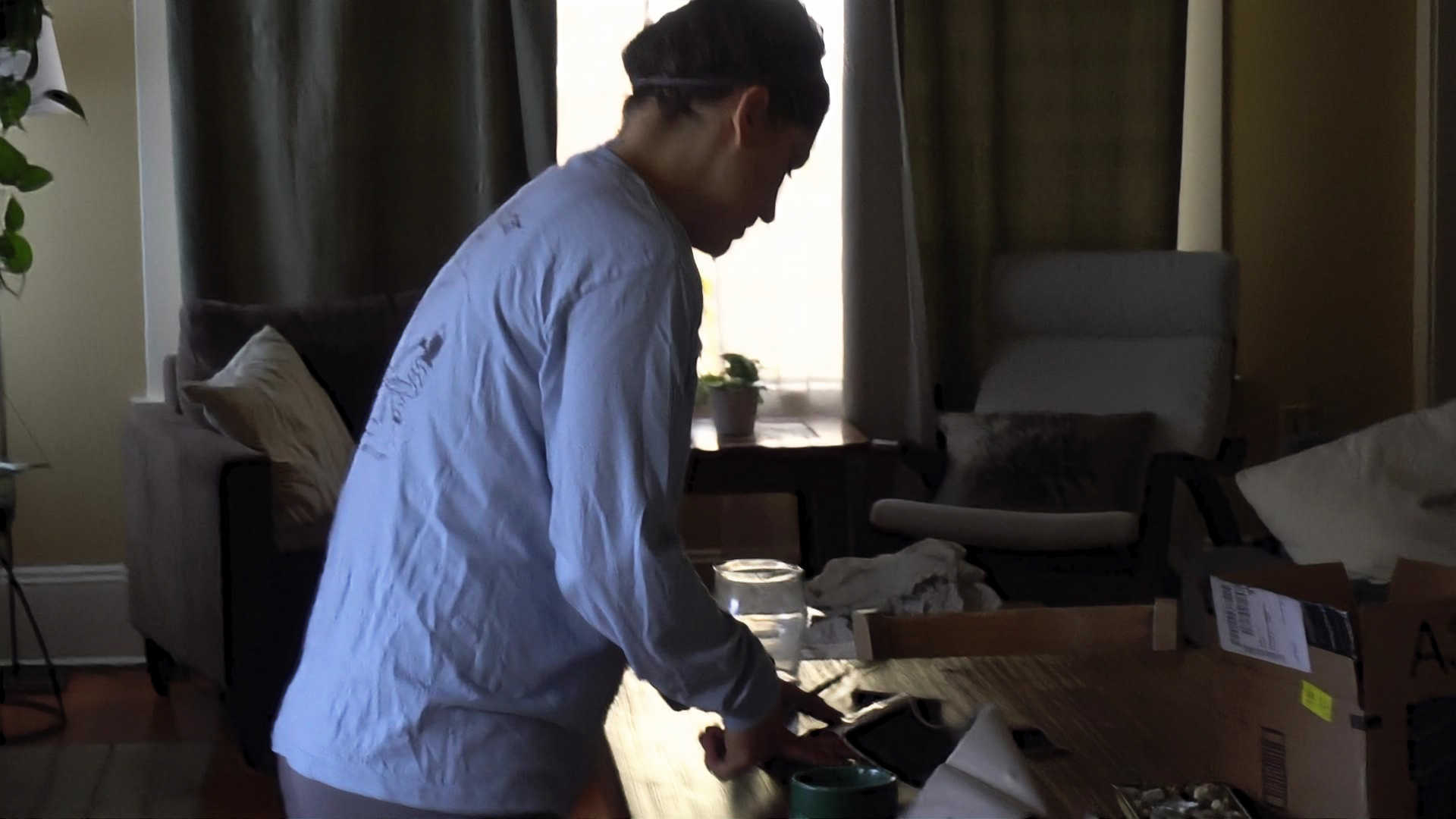} & 
		\cara{figures/generation/moving_obj_in} & 
		\cara{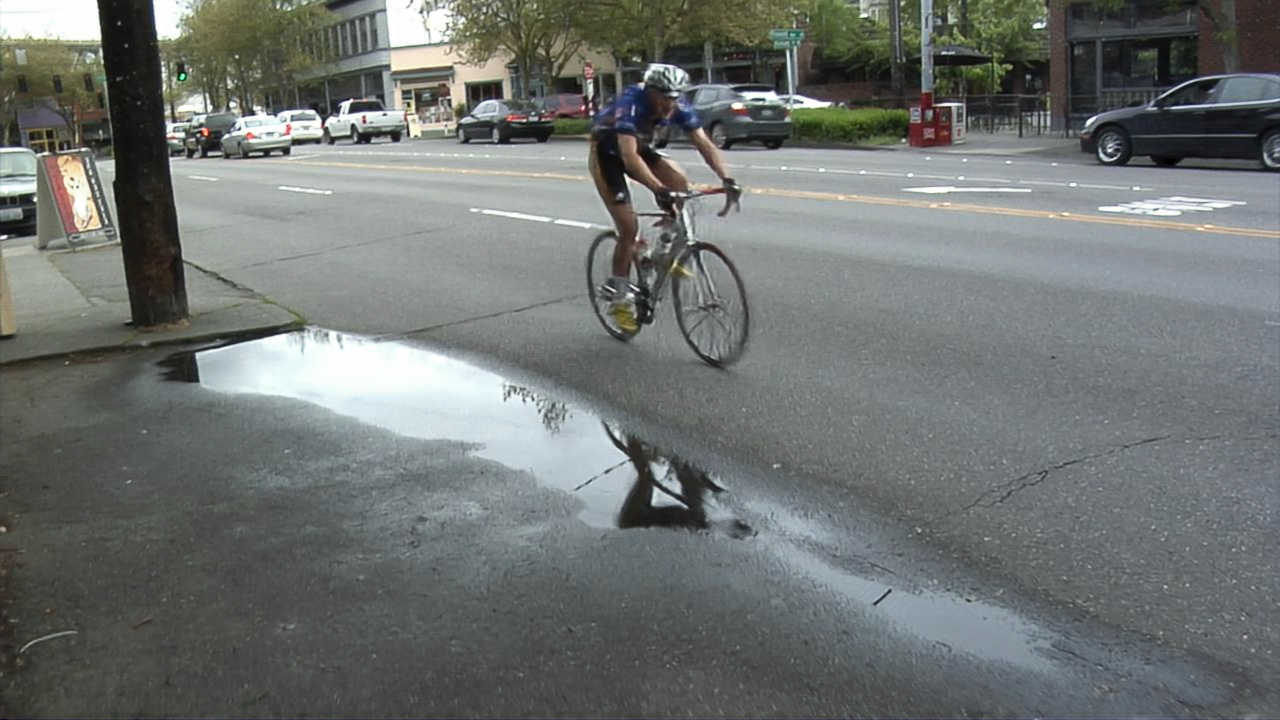} \\		
		\darkb{figures/generation/dark_in} & 
		\darkb{figures/generation/dark_out} & 
		\carb{figures/generation/moving_obj_in} & 
		\carb{figures/generation/moving_obj_out} \\
		Input & \textsc{dbn+homog} & Input & \textsc{dbn+homog} \\
	\end{tabular}	
	\caption{\label{fig:general}This figure shows that our proposed method can generalize to types of data not seen in the training set. The first example shows a low-light, noisy video, and the second shows an example with motion blur, rather than camera shake. The biker is in motion, and is blurred in all frames in the stack, but the network can still perform some moderate deblurring.}
\end{figure}

\paragraph{Generalization to other types of videos.}
As discussed in Sec.~\ref{sec:dataset}, our training set has some limitations. 
Despite these, Fig.~\ref{fig:general} shows that our method can generalize well to other types of scenes not seen during training.
This includes videos captured in indoor, low-light scenarios and motion blur originating from an object moving, rather than the temporally uncorrelated blur from camera shake.
While our dataset has instances of motion blur in it, it is dominated by camera-shake blur.
Nonetheless, the network is able to produce a moderate amount of object motion deblurring as well, which is not handled by other lucky imaging approaches. 

\paragraph{Other experiments.}
We tested with different fusion strategies, for example late fusion, i.e. aggregating features from deeper layers after high-level image content has been extracted from each frame, with both shared and non-shared weights. 
Experimental results show that this produced slightly worse PSNR and training and validation loss, but it occasionally helped in challenging cases where \textsc{dbn+noalign} fails.
However this improvement is not consistent, so we left it out of our proposed approach. 

Multi-scale phase-based methods have proven to be able to generate sharp images using purely Eulerian representations~\cite{meyer2015phase}, so we experimented with multiscale-supervised, Laplacian reconstructions, but found similarly inconclusive results. 
While the added supervision helps in some cases, it likely restricts the network from learning useful feature maps that help in other frames.

We also tried directly predicting the sharp Fourier coefficients, as in~\cite{delbracio2015cvpr}, however this approach did not work as well as directly predicting output pixels. 
One possible reason is that the image quality is more prone to reconstruction errors of Fourier coefficients, and we have not found a robust way to normalize the scale of Fourier coefficients during training, compared with the straightforward way of applying Sigmoid layers when inputs are in the spatial domain. 

\paragraph{Limitations.}
One limitation of this work is that we address only a subset of the types of blur present in video, in particular we focus on motion blur that arises due to camera-shake from hand-held camera motion.
In practice, our dataset contains all types of blur that can be reduced by a shorter exposure time, including object motion, but this type of motion occurs much less frequently. 
Explicitly investigating other sources of blur, for example focus and object motion, which would require different input and training data, is an interesting area for future work.

Although no temporal coherence is explicitly imposed and no post-processing is done,
the processed sequences are in general temporally smooth. 
We refer the reader to the video provided in the supplementary material.
However, when images are severely blurred, our proposed model, especially \textsc{dbn+noalign}, can introduce temporal artifacts that becomes more visible after stabilization.
In the future, we plan to investigate better strategies to handle unaligned cases, for example through the multi-scale reconstruction~\cite{ghiasi2016laplacian,cai2016unified}. 

We would like also to augment our training set with a wider range of videos, as this should increase general applicability of the proposed approach.

\section{Conclusion}
We have presented a learning-based approach to multi-image video deblurring. Despite the above limitations, our method generates results that are often as good as or superior to the state-of-the-art approaches, with no parameter tuning and without the explicit need for challenging image alignment. It is also highly efficient due to the relaxation of the quality of alignment required -- using a simplified alignment method, our approach can generate high quality results within a second, which is substantially faster than existing approaches many of which take minutes per frame. 

In addition, we conducted a number of experiments showing the quality of results varying the input requirements. 
We believe that similar strategies could be applied to other aggregation based applications.

\section*{Acknowledgement}
This work was supported in part by Adobe and the Baseline Funding of KAUST. 

\appendix
\section*{Appendices}
\begin{figure*}[t]
	\centering
	\includegraphics[width=0.32\linewidth]{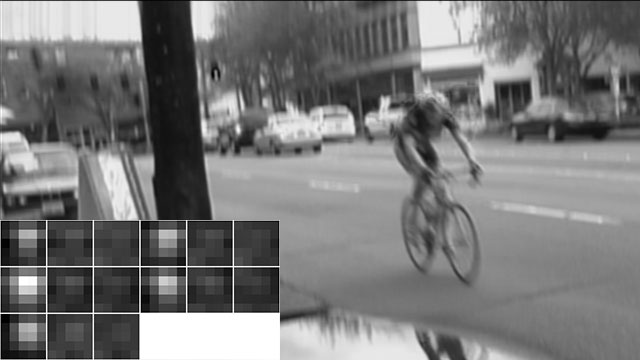}\;
	\includegraphics[width=0.32\linewidth]{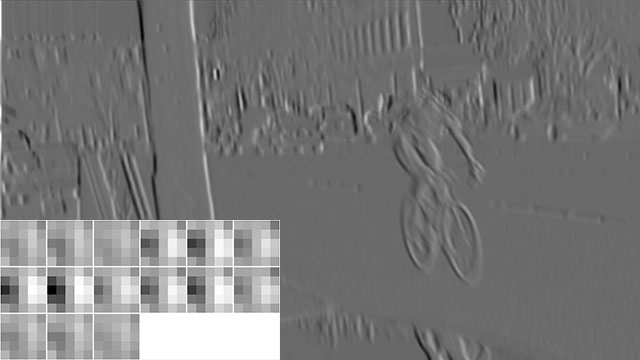}\;
	\includegraphics[width=0.32\linewidth]{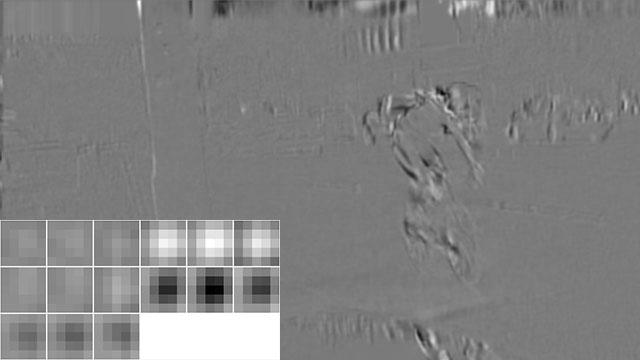}
	\caption{Visualization of 3 out of 64 learned filters and feature maps at F0 from \textsc{dbn+flow}. }
	\label{fig:learned_filters} 
\end{figure*}
\paragraph{Visualization of learned filters. }
Here we visualize some filters learned from \textsc{dbn+flow}, specifically at F0, to gain some insights of how it deblurs an input stack. It can be observed that \textsc{dbn} not only learns to locate the corresponding color channels to generate the correct tone (Fig.~\ref{fig:learned_filters}, left), but is also able to extract edges of different orientations (Fig.~\ref{fig:learned_filters}, middle), and to locate the warping artifacts (Fig.~\ref{fig:learned_filters}, right). 

\begin{figure}[ht]
	\centering
	\includegraphics[width=0.92\linewidth]{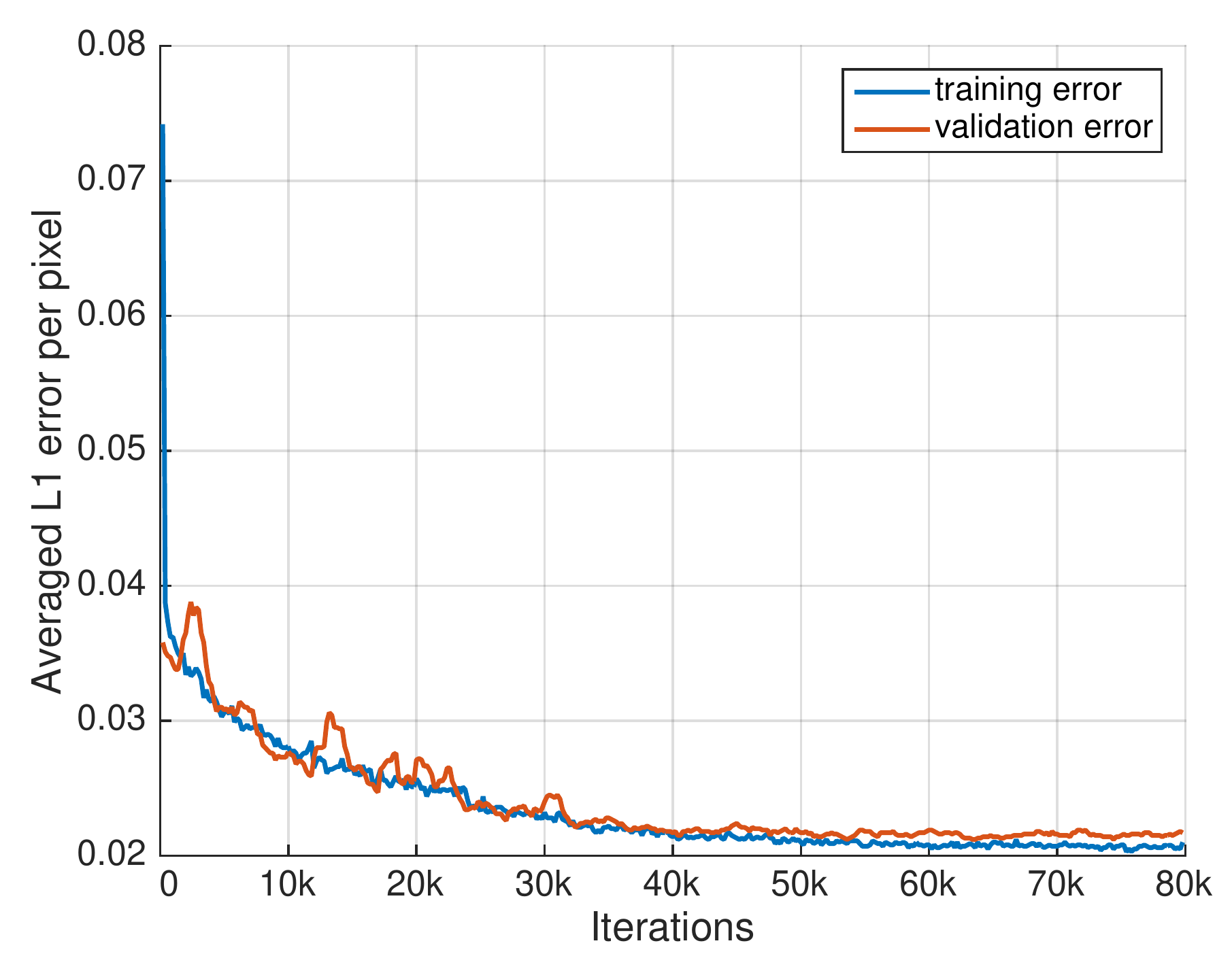}
	\caption{Training and validation errors of \textsc{dbn+flow}. }
	\label{fig:convergence} 
\end{figure}

\paragraph{Convergence. }
The convergence plot of \textsc{dbn+flow} is given in Fig.~\ref{fig:convergence}.

{\small
\bibliographystyle{ieee}
\bibliography{videodeblurring}
}

\end{document}